\documentclass[manuscript]{acmart}
\pdfoutput=1

\AtBeginDocument{
  \providecommand\BibTeX{{
    \normalfont B\kern-0.5em{\scshape i\kern-0.25em b}\kern-0.8em\TeX}}}

\setcopyright{acmcopyright}
\copyrightyear{2018}
\acmYear{2018}
\acmDOI{10.1145/1122445.1122456}

\copyrightyear{2022} 
\acmYear{2022} 
\setcopyright{rightsretained} 
\acmConference[FAccT '22]{2022 ACM Conference on Fairness, Accountability, and Transparency}{June 21--24, 2022}{Seoul, Republic of Korea}
\acmBooktitle{2022 ACM Conference on Fairness, Accountability, and Transparency (FAccT '22), June 21--24, 2022, Seoul, Republic of Korea}\acmDOI{10.1145/3531146.3533118}
\acmISBN{978-1-4503-9352-2/22/06}

\usepackage{array} 
\usepackage{varwidth} 
\usepackage{caption}
\usepackage{longtable}
\usepackage{paralist}
\usepackage{colortbl}
\usepackage{multirow}
\newcolumntype{P}[1]{>{\centering\arraybackslash}p{#1}}
\usepackage{enumitem}
\definecolor{darkpastelgreen}{rgb}{0.01, 0.75, 0.24}
\definecolor{bleudefrance}{rgb}{0.19, 0.55, 0.91}
\definecolor{lust}{rgb}{1.0, 0.63, 0.48}
\definecolor{lust}{rgb}{0.9, 0.13, 0.13}
\usepackage{lscape}
\usepackage[subtle]{savetrees}

\begin{document}

.
\title{Towards a multi-stakeholder value-based assessment framework for algorithmic systems}

\author{Mireia Yurrita}
\email{m.yurritasemperena@tudelft.nl}
\orcid{0000-0002-9685-4873}
\affiliation{
  \institution{Delft University of Technology}
  \city{Delft}
  \country{The Netherlands}
  \postcode{2628 CE}
}

\author{Dave Murray-Rust}
\orcid{0000-0001-6098-7861}
\affiliation{
  \institution{Delft University of Technology}
  \city{Delft}
  \country{The Netherlands}}
\email{D.S.Murray-Rust@tudelft.nl}

\author{Agathe Balayn}
\orcid{0000-0003-2725-5305}
\affiliation{
  \institution{Delft University of Technology}
  \city{Delft}
  \country{The Netherlands}}
\email{a.m.a.balayn@tudelft.nl}

\author{Alessandro Bozzon}
\orcid{0000-0002-3300-2913}
\affiliation{
  \institution{Delft University of Technology}
  \city{Delft}
  \country{The Netherlands}
}
\email{A.Bozzon@tudelft.nl}

\renewcommand{\shortauthors}{Yurrita et al.}

\begin{abstract}

In an effort to regulate Machine Learning-driven (ML) systems, current auditing processes mostly focus on detecting harmful algorithmic biases. While these strategies have  proven to be impactful, some values outlined in documents dealing with ethics in ML-driven systems are still underrepresented in auditing processes. Such \textit{unaddressed} values mainly deal with contextual factors  that cannot be easily quantified. In this paper, we develop a value-based assessment framework that is not limited to bias auditing and that covers prominent ethical principles for algorithmic systems. Our framework presents a circular arrangement of values with two bipolar dimensions that make common motivations and potential tensions explicit. In order to operationalize these high-level principles, values are then broken down into specific criteria and their manifestations. However, some of these value-specific criteria are mutually exclusive and require negotiation. As opposed to some other auditing frameworks that merely rely on ML researchers' and practitioners' input, we argue that it is necessary to include stakeholders that present diverse standpoints to systematically negotiate and consolidate value and criteria tensions. To that end, we map stakeholders with different insight needs, and assign tailored means for communicating value manifestations to them. We, therefore, contribute to current ML auditing practices with an assessment framework that visualizes closeness and tensions between values and we give guidelines on how to operationalize them, while opening up the evaluation and deliberation process to a wide range of stakeholders.

\end{abstract}




\maketitle

\section{Introduction}

In recent years, it has become clear that algorithmic systems might encode harmful biases and might lead to unfair outcomes \cite{morley2020, shen2021}. The dangers of using Machine Learning (ML) in Computer Vision (CV) \cite{buolamwini2018} or Natural Language Processing (NLP) \cite{davidson2019, sap2019, bender2021, larsonnneur2021, alshaabi2021}, for assessing recidivism \cite{tolan2019}, for candidate screening \cite{raghavan2020} and for recommending content on social media platforms \cite{zubiaga2019, parisi2020, kaiser2020, hortaribeiro2020} have been pinpointed. The origins of harmful algorithmic bias\footnote{Following the approach adopted by Shen et al. \cite{shen2021}, we will distinguish between harmful algorithmic biases and harmful algorithmic behaviors, since not all harmful algorithmic behaviors originate from biases and not all algorithmic biases are necessarily harmful \cite{blodgett2020}.} might be diverse \cite{shen2021, suresh2021_2}. Just to mention a few, representativeness issues, play a key role in disparate algorithmic performance \cite{chasalow2021, krafft2020}. The way in which data is collected \cite{bender2021, paullada2020} and labelled \cite{crawford2019, paullada2020, draws2021_5} is a major menace to data soundness. Beyond the data generation process, aggregation, learning, evaluation and deployment biases have been identified throughout the ML pipeline \cite{suresh2021_2}. In response to harmful algorithmic bias, current auditing processes \footnote{We will use the term \textit{auditing} processes to refer to external audits, where third parties only have access to model outputs \cite{sandvig2014}. We will use the term \textit{assessment} processes to refer to an evaluation process that is applied ``throughout the development process and that enables proactive ethical intervention methods'' \cite{raji2020}. We will not use the term \textit{Internal Audit} defined by Raji and Smart \cite{raji2020} to avoid erroneous inferences that would limit the stakeholders of our framework to the employees of an organization.} \cite{saleiro2018, wilson2021, adler2018} have provided numerous useful bias detection techniques \cite{hube2019, ghai2020, xu2018, amini2019, edizel2020, ballburack2021, bellamy2018, zhang2018, yan2020}.

However, harmful algorithmic behavior is not limited to biases encoded in the ML life cycle \cite{shen2021}. The lack of social and cultural context in the mathematical representation of socio-technical systems \cite{martinJr2020, shen2021} or the omission of changing practices and long-term effects of the deployed systems \cite{kuijer2018, damour2020, bountouridis2019, ionescu2021} are also some problematic aspects that are hardly considered in current auditing processes. Such processes mostly consist of quantitative analysis for assessing the conformance of those systems to applicable standards \cite{IEEEaudit}, rather than additionally gaining insights into their contextual implications \cite{shen2021, raji2020}. Furthermore, these auditing approaches solely rely on ML researchers, and practitioners, who can fail to detect issues that arise from context-dependent unanticipated circumstances during usage time \cite{shen2021}. 

In this paper, we argue that:
firstly, assessment processes for algorithmic systems should go beyond bias auditing and take into account additional high-level values \footnote{We will adopt the definition of \textit{values} used in philosophy of science, following Birhane et al. \cite{bihrane2021}. Values of an entity are, thus, defined as properties that are desirable for that kind of entity. } that are outlined in Artificial Intelligence (AI) ethics documents \cite{europeancommissionEthicsAI2019, OECD2019, aichina2018, japanai2019, usai2016,amnestyai2018, royalsociety2019, googleai2018, IBMai2019, microsoftai2018, fjeld2020}. Contestability, for example, has been identified as a key value of algorithmic systems, but there is still little guidance on what contestability requires \cite{lyons2021}. In order to provide a good coverage of values that deal with principled algorithmic behavior, we develop a value-based assessment framework, where contextual conditions are considered along with quantifiable measurements. We organize such values in a circular layout with two bipolar dimensions. As claimed by Friedman et al. \cite{friedman2017}, values do not exist in isolation. They are situated in a delicate balance and touching one value might have implications in another value \cite{friedman2017}. This means that value interactions need to be taken into account when making choices about value prioritization and situating algorithmic systems in a space of trade-offs \cite{bakalar2021}. The circularity of our framework makes such interactions explicit and facilitates the identification of common motivations and tensions among values.

Secondly, an assessment process should give tangible guidelines for the operationalization \footnote{Our strategy follows the definition  by Shahin et al. \cite{shahin2021}, where ``operationalizing values'' is defined as the process of identifying values and translating them into concrete system specifications that can be implemented.} of values, so as to eventually put ethics into practice following a context-aware approach \cite{Shklovski2022}. To this end, each  value in our framework is broken down into criteria manifested through quantifiable indicators, process-oriented practices or signifiers\footnote{We adopt the definition given by Don Norman in his 2013 edition of ``The Design of Everyday Things''. Signifiers are perceivable cues of an affordance, affordances being ``the relationship between the properties of an object and the capabilities of the agent that determine how the object could be possibly used''. In this paper, the ``object'' in question is the ML-driven system.}. These value-specific criteria and their manifestations can be used either as a checklist if our framework is applied for evaluating a system that is already developed, or for promoting such values if it is being used during design time. 

Thirdly, assessment processes should allow critical reflection on algorithmic systems and engage in conflictual plurality\footnote{We understand \textit{conflictuality} as a solution for dealing with the ``figure of alterity''. Unlike \textit{conflict}, it represents a method for linking opposing views and opening out onto the exercise of thinking \cite{gaillard2016}}. Inevitable value tensions inherent in the nature of socio-technical systems \cite{groves2015} require spaces for ethical discussions \cite{Shklovski2022}, that can benefit from the insights of multiple stakeholders beyond ML practitioners \cite{shen2021, bakalar2021}. To enable fruitful multi-stakeholder discussions \cite{lee2021}, we map and match value-specific communication means with different stakeholders. We, therefore, contribute with:
\begin{itemize}[leftmargin=*]
\item A review of prominent high-level values in AI ethics and translation into specific criteria through the:
    \begin{itemize}
    \item Design of an assessment framework that facilitates the identification of common motivations and tensions among values encoded in ML-driven systems.
    \item Definition of guidelines to deal with the complex middle ground between abstract values and concrete system specifications.
    \end{itemize}
\item Translation of value-specific criteria into manifestations that are understandable for diverse stakeholders through the:
    \begin{itemize}
    \item Review of available means to communicate value manifestations to different stakeholders based on their insight needs and nature of knowledge.
    \item Definition of steps to introduce those communication means into multi-stakeholder deliberation dynamics.
    \end{itemize}
\end{itemize}

The remainder of the paper is organized as follows: in section \ref{background}, we analyze related work for documenting and auditing ML systems. We also introduce the theoretical basis of our framework. Section \ref{framework} describes and justifies the selected values, criteria and manifestations and their arrangement in our framework. Section \ref{stakeholders} maps the stakeholders involved in the algorithm evaluation process and reviews the available means for communicating system-specific information to them. In sections \ref{framework} and \ref{stakeholders}, we illustrate the necessary steps for navigating our framework through an example in the area of life insurance application. We  discuss our approach, its implications, and future lines of work in section \ref{limitations}, and we conclude this paper in section \ref{conclusions}. 

\section{Background and related work} \label{background}

In this section, we survey current practices for documenting and auditing technical specifications of algorithmic systems. We also provide the theoretical basis of our framework. 

\subsection{Background}

\subsubsection{Standardized documentation.}
In order to facilitate the audit of ML-driven systems, it is important that technical specifications are collected and documented in a standardized way. So far, ML system documentation practices are limited to datasets and models.

\textit{Documenting datasets.} Recent studies in documentation practices for ML datasets claim the need for greater data transparency \cite{hutchinson2021}. Since the quality of the prediction made by the ML system highly depends on the way the data has been collected, the need for setting rigorous practices (as it is the case in other areas of knowledge, such as social sciences or humanities \cite{stuartgeiger2020}) has been highlighted \cite{paullada2020}. Likewise, the choice of what data to collect and how to collect this data is in itself a value-laden decision \cite{denton2020, scheuerman2021}. To standardize documentation for ML datasets and make data-related decisions more transparent for other practitioners, various methodologies have been suggested in the last years, ``Datasheets'' \cite{gebru2020} and ``Dataset Nutrition Labels'' \cite{holland2018}, for instance. For
NLP techniques, ``Data Statements'' are regarded as a dataset characterization approach that helps developers anticipate biases in language technology and understand how these can be better deployed \cite{bender2018}.

\textit{Documenting models.} In addition to documenting datasets, the importance of disclosing the technical characteristics of ML models has also been emphasized. A good example of model documentation practices are the ``Model Cards'' \cite{mitchell2019}.

\subsubsection{Auditing techniques.}
Various methodologies and tools for incorporating auditing tasks into the Machine Learning workflow have been suggested. Aequitas \cite{saleiro2018} is an open source toolkit to detect traces of bias in models. The toolkit designed by Saleiro et al. \cite{saleiro2018} facilitates the creation of equitable algorithmic decision-making systems where data scientists and policymakers can easily use Aequitas for model selection, evaluation and approval. Wilson et al. \cite{wilson2021} described a framework that helps ensure fairness in socio-technical systems, and used it for auditing the model of the startup \textit{pymetrics}. Adler et al. \cite{adler2018} studied auditing techniques for black-box models to discover whether proxy variables linked to sensitive attributes  indirectly influence the predictions of the system. The end-to-end ``Internal Audit Framework'' suggested by Raji and Smart \cite{raji2020} is of special interest for justifying the need of setting specific guidelines to enable multi-stakeholder deliberation in assessment processes. It  consists of five main stages where the need for stakeholder diversity  is highlighted, e.g. the scoping stage calls for covering a ``critical range of viewpoints'' to review the ethical implications of the system use case.

\subsubsection{Motivation.} While standardized documentation practices \cite{gebru2020, mitchell2019, holland2018, bender2018} and
audits \cite{saleiro2018, adler2018, wilson2021} have been influential methodologies for dealing with harmful algorithmic bias, their scope is limited to performing quantitative analysis over data and model outputs so as to ensure compliance with applicable standards \cite{IEEEaudit}. Such an approach does not deal with additional ethical values which cannot be easily quantified \cite{lee2021} and that are essential for ensuring desirable algorithmic behavior. One could argue that ``Datasheets'' \cite{gebru2020} and ``ModelCards'' \cite{mitchell2019} already devote a section to the description of ethical considerations of datasets and models. Yet, there are no specific guidelines on how to identify ethical issues. As Shklovski et al. \cite{Shklovski2022} discovered, technical people both in industry and academia struggle to identify what an ethical issue entails. To address this caveat, as part of our value-based framework, we give tangible guidelines for putting ethics into practice \cite{Shklovski2022, morley2020}.  We operationalize each high-level value into actionable value criteria and their manifestations. One could also argue that Raji and Smart \cite{raji2020} already included an Ethics Review as part of their end-to-end internal audit framework. Indeed, they exemplified such a review by describing ethical considerations and potential mitigation strategies against bias and privacy threats for a smile detection system. However, this review does not address most of the values that are referred in AI ethics documents. We fill in this gap by offering a good coverage of values to examine, including those that normally go unnoticed in current documenting and auditing practices.

\subsection{Accounting for human values in the assessment of algorithmic systems}

Our ML assessment framework identifies and arranges values encoded in algorithmic systems by covering prominent principles in AI ethics and organizing them in a circular structure.

\subsubsection{Addressing human values in technology.}

For the definition of our value-based framework, we followed other theoretically grounded approaches, such as Value Sensitive Design (VSD) \cite{friedman2017}. VSD represents a pioneering endeavour where human values are proactively considered throughout the process of technology design \cite{davis2015}. Just as VSD does with interactive systems, we address the need to account for human values during the design, implementation, use, and evaluation \cite{davis2015} of algorithmic systems. To this end, we select and define values involved in ML-driven systems, and we identify stakeholders that will be in contact with such systems and whose standpoints need to be considered. Our approach resonates with conceptual investigations described in VSD literature \cite{davis2015}.

The circular nature of our framework is inspired by Schwartz's Theory of Basic Human Values \cite{schwartz2012}. This theory identifies individual value priorities based on ten basic personal values. Values are arranged in a circular form and categorized in four quadrants. These quadrants are located in two bipolar dimensions, which visualize ``oppositions between competing values''. In addition, adjacency between values denotes a common motivation, which results in these values forming a circular continuum. The advantage of adopting a circular arrangement, like the one suggested by Schwartz, for ML-driven systems is that value commonalities and trade-offs can be easily identified thanks to their positioning. Considering the struggles of technical people when addressing ethical issues \cite{Shklovski2022}, an explicit representation of value interactions will facilitate the analysis of trade-offs and decision-making about value prioritization.

\subsubsection{Ethical principles for ML-driven systems}
The values considered in our assessment framework cover prominent principles outlined in AI ethics. In the last five years, many institutions have studied and defined high-level principles that AI systems should follow \cite{fjeld2020}. As a matter of fact, documents that aim at guiding the ``ethical development, deployment and governance of AI'' are converging into a common set of principles \cite{mittelstadt2019, morley2020}. However, high-level principles are far from being actionable \cite{morley2020} and it is necessary to provide answers on how to proceed \cite{aizenberg2020}. Efforts for going from  ``what'' to ``how'' \footnote{Expression used by Morley et al. \cite{morley2020} to refer to the operationalization of ethical principles in AI. The 'what' refers to the ethical principles themselves, whereas the 'how' refers to the act of putting such principles into practice.} include the review carried out by Morley et al. \cite{morley2020}, where available tools for operationalizing ethical principles were examined. Similarly, the AI Ethics Impact (AIEI) Group designed a framework for rating the presence of ethical principles in AI systems, getting inspiration from energy efficiency labels \cite{krafft2020}. 

Our value-based framework differs from previous applied ethics frameworks \cite{morley2020, krafft2020} in various ways. Firstly, we arrange values in a circular form, which makes it easier to navigate common motivations and  trade-offs between values. Although such common motivations and trade-offs can be inferred from current AI ethics documents, we make them explicit by arranging values in a geometrically meaningful way. This is especially useful for identifying overlaps between values that are adjacent to each other and for detecting potential value tensions that need to be negotiated and consolidated. Secondly, we do not limit our ethics framework to a mere checklist. We follow Shklovski et al. \cite{Shklovski2022} and combine the enumeration of tangible and actionable value manifestations with the generation of an open space for ethical debate. As opposed to the deterministic approach adopted by the AIEI group \cite{krafft2020}, we map communication means for facilitating ethical reflections of algorithmic systems and for addressing ethical issues in practice \cite{Shklovski2022}. Thirdly, as opposed to previous applied ethics frameworks \cite{morley2020, krafft2020}, we embrace diversity in ethical reflections and deal with the complexities that arise from plurality. In order to facilitate multi-stakeholder discussions, we match available communication means for addressing different value manifestations with stakeholders that present different insight needs.

\section{Design of our value-based framework} \label{framework}

In this section, we describe the composition of our value-based framework and justify its arrangement. We provide the definition of each of the selected values and the derived criteria and manifestations.

\subsection{Methodology for reviewing values, criteria and manifestations in ML-driven systems}
To design our framework, we analysed documents outlining high-level ethical principles that ML systems should follow. Our starting point was the review performed by Fjeld et al. \cite{fjeld2020}, where principles coming from governments, inter-governmental organizations, multiple stakeholders, the private sector, and the civil society were examined. In their review, Fjeld et al. identify nine key themes, some of which overlap with the values outlined in our framework.
The identification of prominent high-level values was also complemented with other reviews \cite{krafft2020,bihrane2021,morley2020,hidalgo2021,simons2019}. 
To identify the criteria that define the fulfilment of prominent high-level values, we navigated the visual representation provided by Fjeld et al. \cite{fjeld2020} and accessed the documents that offer a higher coverage of the value in question. For instance, for the value of \textit{privacy}, one of our main references has been the GDPR \cite{GDPR2018}. 

We went from criteria to value manifestations through an extensive exploration of available value-specific reviews that identify such manifestations. For instance, for the value of \textit{security} Xiong et al. \cite{xiong2021} presented a thorough study of  mechanisms  used for securing the ML pipeline against external threats. For \textit{explainability}, Barredo-Arrieta et al. \cite{barredoarrieta2020} put together more than four hundred references and mapped strategies in the field of Explainable Artificial Intelligence \cite{barredoarrieta2020}. We partly rely on such reviews for identifying value manifestations because our contribution lies in covering and putting together a set of values and their manifestations in ML-driven systems to end up with a ``health-check'' for assessing algorithmic systems, rather than rediscovering such value manifestations ourselves. Similarly, for the values of \textit{performance} and \textit{fairness}, we only included the main value manifestations that represent the basis for any other derived metrics. That is to say, just as Verma et al. \cite{verma2018} did, we outline the main quantifiable indicators (false positives, false negatives etc) used for measuring \textit{performance} and \textit{fairness}, but we are aware that  many other metrics that  derive from these ones can be insightful for specific contexts. Dealing with such compound metrics is out of the scope of this work.

\subsection{Assessment of algorithmic systems through a circular value-based framework} \label{frameworkdetails}

\begin{figure}
  \centering
  \includegraphics[width = 0.5\textwidth]{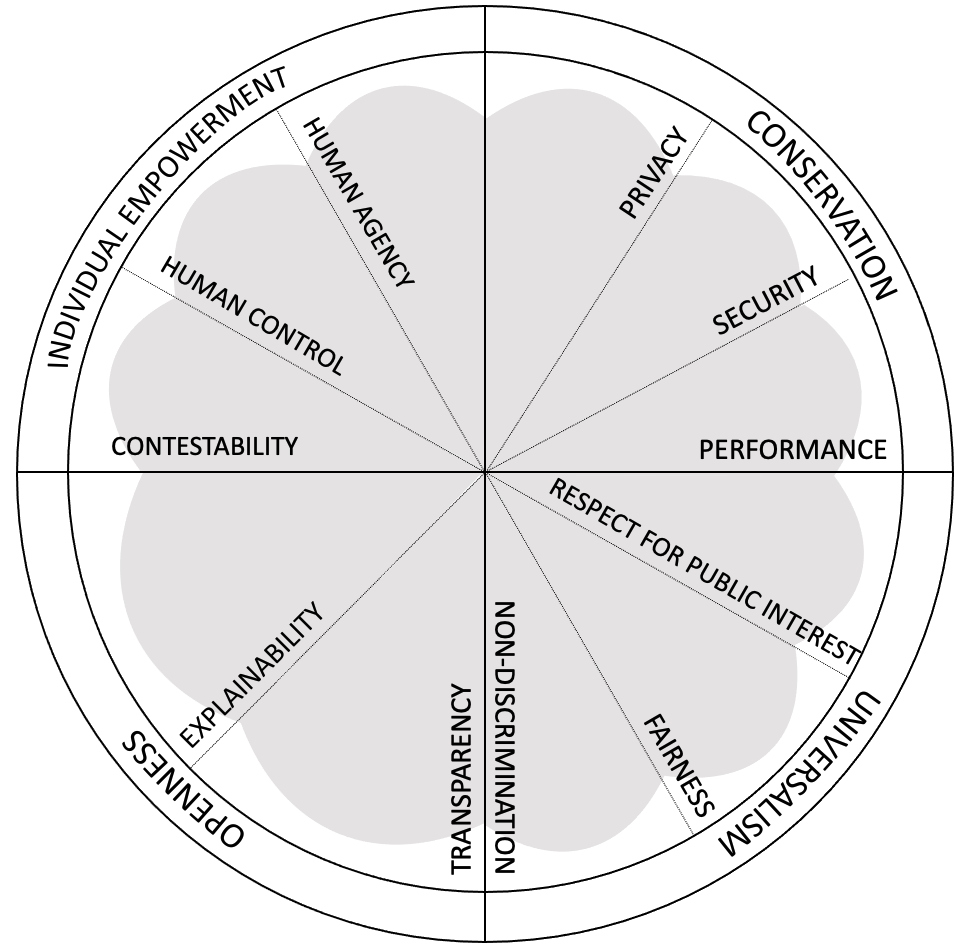}
  \caption{Graphic representation of our circular value-based assessment framework. Oppositions between competing values are illustrated through the arrangement of those values in bipolar dimensions and common motivations through adjacency between values, which form a circular continuum.}
  \label{fig:framework}
\end{figure}

\begin{figure}
  \centering
  \includegraphics[scale = 0.26]{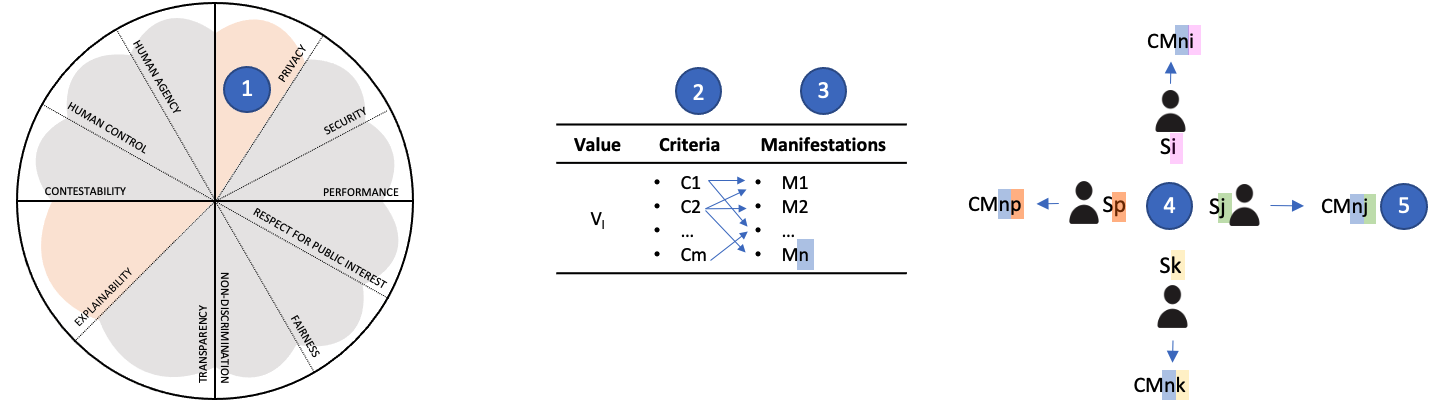}
  \caption{Workflow for operationalizing high-level values and for enabling multi-stakeholder assessment of algorithmic systems. This workflow represents the methodology that we followed for structuring our framework and the steps that researchers and practitioners should take to make use of it.
  (1) Select and discuss project-specific values (V), (2) Decide on criteria (C) for embodying those values, (3) Select the manifestations (M) that enact value-specific criteria, (4) Map relevant stakeholders (S) to enable ethical reflection of value and criteria tensions, (5) Match adequate communication means (CM) to stakeholders.}
  \label{fig:framework_diagram}
\end{figure}

Our resulting ML assessment framework arranges values in a circular form (figure \ref{fig:framework}). Adjacency between values denotes a common motivation and oppositions between competing values are represented through two bipolar dimensions. For instance, adjacency between \textit{privacy} and \textit{security} denotes a common objective towards the protection of sensitive information  \cite{fjeld2020, russel2015} and resilience to external threats \cite{microsoftai2018}. The trade-off between \textit{privacy} and \textit{explainability}, on the other hand, is made explicit by their opposing positioning in our circular framework. High-level values are then broken down into specific criteria and their manifestations, as indicated in figure \ref{fig:framework_diagram}. Criteria defining a specific value ultimately represent a set of questions to be asked as part of the assessment process to ensure the fulfillment of the value in question ---if the framework is being applied before deployment--- or the promotion of a specific value ---if the framework is being applied during design time---. These sets of criteria are not unique and exclusive to one value.
For instance, when defining the criteria for \textit{privacy} we refer to ``data protection'', which is also involved in \textit{security} in the form of ``resilience to attacks''. These overlaps are precisely what we want to highlight and make explicit thanks to the circularity of our framework and adjacency between values.

Manifestations are classified in three groups depending on their nature: (1) \textit{\textcolor{magenta}{Quantifiable indicators}} are specific measurable parameters that numerically manifest the (lack of) adequacy in the standards set for a criterion (magenta). (2) \textit{\textcolor{olive}{Process-oriented practices}} are actions and mechanisms implemented during the ML development or deployment process that advocate for a certain value (olive). (3) \textit{\textcolor{orange}{Signifiers} \footnote{Check footnote 5}} are files and reports that 
describe the relationship between the properties of the algorithmic system and humans that determine how that system can be used (orange). There is a many-to-many relationship between criteria and manifestations.

In the next subsections, we present opposing value categories in pairs. We explain in detail values, the criteria that define them and their manifestations.

\paragraph{Motivating example.}
For illustrative purposes, we guide the reader through each stage of our framework with a hypothetical yet plausible use case. Consider a team of researchers is developing an ML-driven system for automating life insurance application processes. The system shall accept or refuse the request of a life insurance based on the following data: physiological information of the candidate, details about their employment, insurance history, and individual and family medical history. As part of a new wave of ethical finance companies, the team would like to ensure that their work is ethically grounded. However, it is not clear what that means in practice. From looking at prominent literature, they can develop a sense that the model should be fair and unbiased, and potentially that there should be some level of human control or intervention possible. Yet, they cannot be sure if they have a good set of representative values covered, and they do not know how to go about communicating the way that their model embodies those values. They are now reading through our multi-stakeholder value-based assessment framework.

\subsection{Conservation vs Openness}

The first dimension of our value-based framework captures the conflict between \textit{conservation} and \textit{openness}. Values included within the \textit{conservation} category emphasize the necessity of ML-driven systems to preserve confidentiality with regards to information, as well as, the need for the system to preserve adequate robustness when it comes to performance. On the contrary, the category of \textit{openness} encompasses values that advocate for making system components and specifications more accessible to the public.

\subsubsection{Conservation.} Privacy, security and performance uphold confidentiality and robustness within ML systems \cite{fjeld2020}. 

\paragraph{Privacy}
The defining goal of \textit{privacy} is the need for ML-driven systems to respect individual's informational confidentiality \cite{fjeld2020, krafft2020} as part of their user rights \cite{bihrane2021}. When applying this value to the ML development pipeline, data processing itself should integrate privacy standards \cite{krafft2020, fjeld2020, morley2020}, so that there is no possibility of identifying sensitive information about individuals \cite{hidalgo2021, wachter2019}. Furthermore, the need to provide humans with agency over their data is emphasized \cite{fjeld2020}. Based on these definitions, we  identified six main criteria for the fulfillment of \textit{privacy} within ML systems (table \ref{tab:exampleprivacy}). (1) Consent for data usage \cite{GDPR2018, fjeld2020, krafft2020}: data subjects should be appropriately informed when their data is being used and their explicit approval is needed. (2) Implementation of data protection mechanisms \cite{floridi2019, fjeld2020, krafft2020}: during the development of algorithmic systems, resources should be devoted to making user data management secure and confidential. (3) Users having control over their data and ability to restrict its processing \cite{GDPR2018, fjeld2020}: users should be able to limit the way their personal data is being used. (4) Users having the right to rectify \cite{GDPR2018, fjeld2020, krafft2020}: users should be able to modify their data at any time. (5) Users having the right to erase their data \cite{GDPR2018, fjeld2020, krafft2020}: this criterion refers to the right that users have to be forgotten. (6) Users having right to access their data \cite{GDPR2018, IEEE2019}: this right empowers users to have agency over their data.  These criteria   manifest in various  ways. Signifiers include: a written declaration of consent \cite{GDPR2018}, detailed descriptions of the collected data, how data is handled, how long it will be kept and the purpose of collecting that data \cite{mehldau2007}. These signifiers are necessary for users to fully understand what sharing their data entails. Process-oriented practices include the obfuscation of data \cite{krafft2020} and forms and submission mechanisms to object data collection and  make complaints \cite{blazevic2021}.

\setlist[itemize]{wide=0pt, noitemsep, label=\textbullet, topsep=2pt, leftmargin=*, after =\vspace*{-\baselineskip}}
\setlist[enumerate]{wide=0pt, noitemsep , topsep=2pt, leftmargin=*, after =\vspace*{-\baselineskip}}
\begin{singlespace}
\begin{longtable}{p{.025\textwidth} m{.115\textwidth} p{.35\textwidth} p{.41\textwidth}} \hline
\footnotesize
 & Value & Criteria    & Manifestations \\ \hline \endhead
 \multirow{8}{*}{\rotatebox[origin=c]{90}{Conservation}} & \multirow{8}{*}{Privacy} & \mbox{}\par\vspace{-\baselineskip} \begin{enumerate} \item \textbf{Consent for data usage} \cite{GDPR2018, fjeld2020, krafft2020} \item \textbf{Data protection} \cite{floridi2019, fjeld2020, krafft2020} \item \textbf{Control over data / ability to restrict processing} \cite{GDPR2018, fjeld2020} \item \textbf{Right to rectification} \cite{GDPR2018, fjeld2020, krafft2020} \item \textbf{Right to erase the data} \cite{GDPR2018, fjeld2020, krafft2020} \item \textbf{Right of access by data subject, data agency} \cite{GDPR2018, IEEE2019}\end{enumerate} & \mbox{}\par\vspace{-\baselineskip} \begin{itemize} \item \textcolor{orange}{Written declaration of consent} \cite{GDPR2018} \item \textcolor{orange}{Description of what data is collected} \cite{mehldau2007} \item \textcolor{orange}{Description of how data is handled} \cite{mehldau2007} \item \textcolor{orange}{Purpose statement of data collection} \cite{mehldau2007} \item \textcolor{orange}{Statement of how long the data is kept} \cite{mehldau2007} \item \textcolor{olive}{Form and submission mechanisms to object data collection and to make complaints} \cite{blazevic2021} \item \textcolor{olive}{Obfuscation of data} \cite{krafft2020} \end{itemize} \\ \hline
\caption{Illustration of how to move from values, to criteria and their manifestations for \textit{privacy}.}\label{tab:exampleprivacy}
\end{longtable}
\end{singlespace}

\paragraph{Security}
Definitions characterizing \textit{security} ---table \ref{tab:examplesecurity}--- highlight the need for ML systems to be (1) resilient to potential maleficent attacks \cite{morley2020, fjeld2020} and to present a (2) predictable  \cite{fjeld2020, krafft2020, europeancommissionEthicsAI2019} and (3) robust \cite{krafft2020} behavior at any time. This includes implementing mechanisms to protect user privacy, such as strategies that ensure that inferences about an individual cannot be made by interrogating the model \cite{microsoftai2018, hidalgo2021, wachter2019}. Following the survey performed by Xiong et al. \cite{xiong2021}, different methodologies that aim at protecting algorithmic systems against external threats (process-oriented practices) have been classified into two main groups. The first group consists of defence methods against integrity threats at two different stages of the ML pipeline: during training time \cite{biggio2018, cretu2008, globerson2006} and during prediction time \cite{biggio2018, lyu2015, goodfellow2014, papernot2016}. The second group aim at defending the ML system against privacy threats, namely membership inference attacks \cite{shokri2017, nasr2018, ye2021, dwork2006, jia2019}. 

\setlist[itemize]{wide=0pt, noitemsep, label=\textbullet, topsep=2pt, leftmargin=*, after =\vspace*{-\baselineskip}}
\setlist[enumerate]{wide=0pt, noitemsep , topsep=2pt, leftmargin=*, after =\vspace*{-\baselineskip}}
\begin{singlespace}
\begin{longtable}{p{.025\textwidth} m{.115\textwidth} p{.35\textwidth} p{.41\textwidth}} \hline
\footnotesize
 & Value & Criteria    & Manifestations \\ \hline \endhead
\multirow{25}{*}{\rotatebox[origin=c]{90}{Conservation}}& \multirow{25}{*}{Security} & \begin{itemize} \item[\textcolor{white}{\textbullet}] \item[\textcolor{white}{\textbullet}] \item[\textcolor{white}{\textbullet}] \item[\textcolor{white}{\textbullet}] \item[\textcolor{white}{\textbullet}] \item[\textcolor{white}{\textbullet}] \item[\textcolor{white}{\textbullet}] \item[\textcolor{white}{\textbullet}] \item[\textcolor{white}{\textbullet}]  \end{itemize}\mbox{}\par\vspace{-\baselineskip} \begin{enumerate}  \item \textbf{Resilience to attacks}: protection of privacy \cite{microsoftai2018, hidalgo2021, wachter2019}, vulnerabilities, fallback plans \cite{morley2020, fjeld2020, krafft2020, googleai2018} \item \textbf{Predictability} \cite{fjeld2020, krafft2020, europeancommissionEthicsAI2019} \item \textbf{Robustness / reliability}: prevent manipulation \cite{krafft2020}\end{enumerate} &  AGAINST INTEGRITY THREATS \cite{xiong2021}: \begin{itemize} \item Training time \cite{xiong2021} Ex.: \begin{itemize} \item \textcolor{olive}{Data sanitization} \footnote{It ensures data soundness by identifying abnormal input samples and by removing them \cite{xiong2021}.} \cite{biggio2018, cretu2008} \item \textcolor{olive}{Robust learning} \footnote{It ensures that algorithms are trained on statistically robust datasets, with little sensitivity to outliers \cite{xiong2021}.} \cite{biggio2018, globerson2006}\end{itemize} \vspace{0.5cm}\item Prediction time \cite{xiong2021}\begin{itemize} \item \textcolor{olive}{Model enhancement} \cite{biggio2018, lyu2015, goodfellow2014, papernot2016} Ex.: \begin{itemize} \item \textcolor{olive}{Adversarial Learning} \footnote{Adversarial samples are introduced to the training set \cite{xiong2021}.} \item \textcolor{olive}{Gradient masking} \footnote{Input gradients are modified to enhance model robustness \cite{xiong2021}.} \item \textcolor{olive}{Defensive Distillation} \footnote{The dimensionality of the network is reduced \cite{xiong2021}.} \vspace{0.5cm} \end{itemize} \end{itemize} \end{itemize} \vspace{0.8cm}AGAINST PRIVACY THREATS \cite{xiong2021}:\begin{itemize} \item Mitigation techniques \cite{nasr2018}: \begin {itemize}\item \textcolor{olive}{Restrict prediction vector to top k classes} \footnote{Applicable when the number of classes is very large. Even if the model only outputs the most likely k classes, it will still be useful \cite{shokri2017}.} \cite{shokri2017} \item \textcolor{olive}{Coarsen the precision of the prediction vector} \footnote{It consists in rounding the classification probabilities down \cite{shokri2017}.} \cite{shokri2017} \item \textcolor{olive}{Increase entropy of the prediction vector} \footnote{Modification of  the softmax layer (in neural networks) to increase its normalizing temperature \cite{shokri2017}.} \cite{shokri2017} \item \textcolor{olive}{Use regularization} \footnote{Technique to avoid overfitting in ML that penalizes large parameters by adding a regularization factor $\lambda$ to the loss function \cite{shokri2017}.} \cite{shokri2017, kaya2020} \end{itemize}\vspace{0.3cm} \item Differential privacy mechanisms \cite{nasr2018}: \begin {itemize}\item \textcolor{olive}{Differential privacy} \footnote{It prevents any adversary from distinguishing the predictions of a model when its training dataset is used compared to when other dataset is used \cite{ye2021}} \cite{ye2021, dwork2006}. \textcolor{black} Ex.: \begin{itemize} \item \textcolor{olive}{Adversarial regularization} \footnote{Membership privacy is modeled as a min-max optimization problem, where a model is trained to achieve minimum loss of accuracy and maximum robustness against the strongest inference attack \cite{nasr2018}.} \cite{nasr2018} \item \textcolor{olive}{MemGuard} \footnote{Noise is added to the confidence vector of the attacker so as to mislead the attacker's classifier \cite{jia2019}}\vspace{0.5cm} \cite{jia2019}\end{itemize} \vspace{0.5cm} \end{itemize} \end{itemize} \\\hline

\caption{Criteria and their manifestations for \textit{security}.}\label{tab:examplesecurity}

\end{longtable}
\end{singlespace}

\paragraph{Performance}
The value of \textit{performance} ---table \ref{tab:exampleperformance}--- is defined by the (1) correctness of predictions \cite{fjeld2020, europeancommissionEthicsAI2019}, along with the (2-5) resources necessary to reach such predictions \cite{bihrane2021, krafft2020, Kulynych2020}. The conditions under which systems are evaluated will have a direct impact on the ``appropriateness score'' that these systems will obtain in the form of a quantifiable indicator \cite{dotan2019}. In other words, if the level of performance is solely measured in terms of accuracy, regardless of the needed data, prerequisites will be inherently favoring big ``data-hungry'' \cite{lee1973} models.  As far as the measurement of performance is concerned, this is mainly done through quantifiable indicators, either referring to the preciseness of the results \cite{mitchell2019, wexler2019} or to the estimated consumption of environmental resources \cite{garciamartin2019, gao2020, mahendran2021, assran2020, dalton2020}. 

\setlist[itemize]{wide=0pt, noitemsep, label=\textbullet, topsep=2pt, leftmargin=*, after =\vspace*{-\baselineskip}}
\setlist[enumerate]{wide=0pt, noitemsep , topsep=2pt, leftmargin=*, after =\vspace*{-\baselineskip}}
\begin{singlespace}
\begin{longtable}{p{.025\textwidth} m{.115\textwidth} p{.35\textwidth} p{.41\textwidth}} \hline
\footnotesize
 & Value & Criteria    & Manifestations \\ \hline \endhead

\multirow{15}{*}{\rotatebox[origin=c]{90}{Conservation}}& \multirow{15}{*}{Performance}                 & 
\multirow{15}{=}{ \begin{enumerate} \item \textbf{Correctness of predictions} \cite{fjeld2020, europeancommissionEthicsAI2019, bihrane2021} \item \textbf{Memory efficiency} \cite{bihrane2021, krafft2020} \item \textbf{Training efficiency} \cite{bihrane2021} \item \textbf{Energy efficiency} \cite{bihrane2021, krafft2020} \item \textbf{Data efficiency} \cite{bihrane2021}\end{enumerate}   }        
& \mbox{}\par\vspace{-\baselineskip}  \begin{itemize} \item \textcolor{magenta}{Accuracy (for classification, sum of true positive and true negative rates)} \cite{mitchell2019, wexler2019} \item \textcolor{magenta}{False Positive and False Negative rates} \cite{mitchell2019, wexler2019} \item \textcolor{magenta}{False Discovery and Omission Rate} \cite{mitchell2019}
\item \textcolor{magenta}{Mean and median error} \cite{wexler2019}\item \textcolor{magenta}{R2 score} \cite{bird2020} \item \textcolor{magenta}{Precision and recall rates} \cite{wexler2019} \item \textcolor{magenta}{Area under ROC curve (AUC)} \cite{bird2020}  \item \textcolor{magenta}{Estimation of energy consumption through} \cite{garciamartin2019}: \begin{itemize} \item performance counters \item simulation \item instruction- or architecture-level estimations \item real-time estimation \end{itemize} \item \vspace{0.3cm} \textcolor{magenta}{Estimation of GPU memory consumption}  \cite{gao2020, mahendran2021} \item \textcolor{magenta}{Wall-clock training time} \cite{assran2020, dalton2020}\end{itemize}                    \\ \hline

\caption{Criteria and their manifestations for \textit{performance}.}\label{tab:exampleperformance}

\end{longtable}
\end{singlespace}

\subsubsection{Openness.} Transparency and explainability advocate for making system components and specifications accessible.

\paragraph{Transparency}
Documents providing high-level principles for AI  define \textit{transparency} as the property that enables traceability and monitoring of algorithmic systems \cite{morley2020, fjeld2020} ---table \ref{tab:exampletransparency}---. \textit{Transparency} relates to the right to information \cite{fjeld2020} and requires that data or algorithms present some level of accessibility \cite{royalsociety2019}. That is to say, data and models should present some level of (1) interpretability \cite{royalsociety2019, bihrane2021}, so as to (2) enable human oversight \cite{fjeld2020, morley2020}. Those data and models should also be (3) accessible \cite{krafft2020, royalsociety2019, fjeld2020}, as a step towards achieving (4) traceability \cite{morley2020}  and (5) reproducibility \cite{bihrane2021}. Manifestations of such criteria emerge mostly in the form of documentation detailing technical aspects of the algorithmic system (considered signifiers in our framework) \cite{gebru2020, bender2018, chasalow2021, stuartgeiger2020, krafft2020, morley2020, mitchell2019, royalsociety2019}. Process-oriented practices mostly focus on giving open access to data and algorithms \cite{fjeld2020, bihrane2021, krafft2020, royalsociety2019}, regularly reporting key information about the system \cite{fjeld2020} and notifying users whenever they are being subject to or interacting with an algorithmic system \cite{fjeld2020}.

\setlist[itemize]{wide=0pt, noitemsep, label=\textbullet, topsep=2pt, leftmargin=*, after =\vspace*{-\baselineskip}}
\setlist[enumerate]{wide=0pt, noitemsep , topsep=2pt, leftmargin=*, after =\vspace*{-\baselineskip}}
\begin{singlespace}
\begin{longtable}{p{.025\textwidth} m{.115\textwidth} p{.35\textwidth} p{.41\textwidth}} \hline
\footnotesize
& Value & Criteria    & Manifestations \\ \hline \endhead

\multirow{8}{*}{\rotatebox[origin=c]{90}{Openness}  }& \multirow{8}{*}{Transparency} & \mbox{}\par\vspace{-\baselineskip} \begin{enumerate} \item \textbf{Interpretability of data and models} \cite{royalsociety2019, bihrane2021} \item \textbf{Enabling human oversight of operations} \cite{fjeld2020, morley2020} \item \textbf{Accessibility of data and algorithm} \cite{krafft2020, royalsociety2019, fjeld2020} \item \textbf{Traceability} \cite{morley2020} \item \textbf{Reproducibility} \cite{bihrane2021} \end{enumerate}  & \mbox{}\par\vspace{-\baselineskip} \begin{itemize} \item \textcolor{orange}{Description of data generation process} \cite{gebru2020, bender2018, chasalow2021, stuartgeiger2020, krafft2020, morley2020} \item \textcolor{orange}{Disclosure of origin and properties of models and data} \cite{mitchell2019, krafft2020, royalsociety2019} \item \textcolor{olive}{Open access to data and algorithm} \cite{fjeld2020, bihrane2021, krafft2020, royalsociety2019}  \item \textcolor{orange}{Notification of usage/interaction} \cite{fjeld2020} \item \textcolor{orange}{Regular reporting} \cite{fjeld2020}\end{itemize} \\ \hline

\caption{Criteria and their manifestations for \textit{transparency}.}\label{tab:exampletransparency}
\end{longtable}
\end{singlespace}

\paragraph{Explainability}
Explainable Artificial Intelligence (XAI) is formed by a set of techniques that allow a wide range of stakeholders to understand why or how a decision was reached by an algorithmic system \cite{royalsociety2019, floridi2019}. \textit{Explainability} is, thus, conceived as ---table \ref{tab:exampleexplainability}--- an interface  that translates reasoning mechanisms of the system into formats that are (1) comprehensible \cite{barredoarrieta2020, fjeld2020, floridi2018, floridi2019, europeancommissionEthicsAI2019, OECD2019, bihrane2021, royalsociety2019}. In addition, strategies for making black-box algorithms more interpretable facilitate their (2) monitoring \cite{morley2020} and, therefore, make them (3) suitable for evaluation \cite{fjeld2020, morley2020}. XAI techniques (process-oriented practices) are very diverse in nature. As claimed by Vera Liao et al. \cite{veraliao2020} and Barredo-Arrieta et al. \cite{barredoarrieta2020}, \textit{explainability} methodologies are usually classified by the scope of the explanation, complexity of the model, model specificity and the stage of the ML pipeline where such a strategy is to be used. For our framework, we will consider that explainable models can be either (a) interpretable by design or they can be (b) explained by additional \textit{post-hoc} explanations \cite{barredoarrieta2020}. 

\setlist[itemize]{wide=0pt, noitemsep, label=\textbullet, topsep=2pt, leftmargin=*, after =\vspace*{-\baselineskip}}
\setlist[enumerate]{wide=0pt, noitemsep , topsep=2pt, leftmargin=*, after =\vspace*{-\baselineskip}}
\begin{singlespace}
\begin{longtable}{p{.025\textwidth} m{.115\textwidth} p{.35\textwidth} p{.41\textwidth}} \hline
\footnotesize
& Value & Criteria    & Manifestations \\ \hline \endhead

\multirow{5}{*}{\rotatebox[origin=c]{90}{Openness}}& \multirow{5}{*}{Explainability}              & \mbox{}\par\vspace{-\baselineskip} \begin{enumerate} \item \textbf{Ability to understand AI systems and the decision reached} \cite{floridi2018, floridi2019, europeancommissionEthicsAI2019, OECD2019, bihrane2021, royalsociety2019}  \item \textbf{Traceability} \cite{morley2020} \item \textbf{Enable evaluation} \cite{fjeld2020, morley2020}\end{enumerate}                   & \mbox{}\par\vspace{-\baselineskip} \begin{itemize} \item \textcolor{olive}{Interpretability by design} \cite{barredoarrieta2020} \item \textcolor{olive}{Post-hoc explanations} \cite{barredoarrieta2020}\end{itemize}   \\ \hline

\caption{Criteria and their manifestations for \textit{explainability}.}\label{tab:exampleexplainability}
\end{longtable}
\end{singlespace}

\subsection{Universalism vs Individual Empowerment}
The second dimension captures the conflict between \textit{universalism} and \textit{individual empowerment}. Values included within the \textit{individual empowerment} category emphasize the defense of the decision subjects' interests. These principles  advocate for giving decision subjects the means to oppose to the conclusion reached and uphold the need for putting humans in the loop. Values within the \textit{universalism} category emphasize the need to equalize system behavior to \textit{all} and to ensure that such a system adheres to the interests of society as a whole, beyond the interests of a few individuals.

\subsubsection{Universalism.} Respect for public interest, fairness and non-discrimination uphold the need to ensure equitable and socially acceptable system behavior for \textit{all}.

\setlist[itemize]{wide=0pt, noitemsep, label=\textbullet, topsep=2pt, leftmargin=*, after =\vspace*{-\baselineskip}}
\setlist[enumerate]{wide=0pt, noitemsep , topsep=2pt, leftmargin=*, after =\vspace*{-\baselineskip}}
\begin{singlespace}
\begin{longtable}{p{.025\textwidth} m{.115\textwidth} p{.35\textwidth} p{.41\textwidth}} \hline
\footnotesize
& Value & Criteria    & Manifestations \\  \hline

\multirow{5}{*}{\rotatebox[origin=c]{90}{Universalism}} & \multirow{5}{=}{Respect for public interest} &  \mbox{}\par\vspace{-\baselineskip}  \begin{enumerate} \item \textbf{Desirability of technology} \cite{chasalow2021, abebe2020, keyes2019} \item \textbf{Benefit to society} \cite{fjeld2020, floridi2018, floridi2019, morley2020} \item \textbf{Environmental impact} \cite{krafft2020, bender2021}\end{enumerate}  & \mbox{}\par\vspace{-\baselineskip}  \begin{itemize} \item \textcolor{olive}{Diverse and inclusive forum for discussion} \cite{frenchminister2019, fjeld2020}\item \textcolor{orange}{Measure of social and environmental impact} \cite{morley2020, raji2020, bender2021} \vspace{0.3cm}\end{itemize} \\ \hline

\caption{Criteria and their manifestations for \textit{respect for public interest}.}\label{tab:examplerespect}
\end{longtable}
\end{singlespace}

\paragraph{Respect for public interest}
The value of \textit{respect for public interest} ---table \ref{tab:examplerespect}--- deals with the (1) appropriateness of developing algorithmic systems for a certain purpose within a specific context. As Keyes et al. \cite{keyes2019} claimed, making ML-driven systems fairer, more transparent and more accountable is insufficient if we ignore the purpose of developing and implementing these systems in a certain context in the very first place \cite{leins2020, tsarapatsanis2021}. Algorithmic systems should, therefore, (2) be beneficial to society and humanity as a whole \cite{floridi2018, floridi2019, morley2020, fjeld2020}, respect law \cite{bihrane2021} and be aligned with human norms \cite{fjeld2020}. This involves giving a clear justification of the purpose and benefits of building such a system \cite{morley2020, chasalow2021, abebe2020, keyes2019}, so that the deployment of the system in question upholds public-spirited goals \cite{fjeld2020}. Universalism aims at protecting the welfare of \textit{all}, both people and nature \cite{Kulynych2020}. AI systems' (3) negative impacts on environment should, therefore, be considered and valued \cite{krafft2020, bender2021}. To this end, process-oriented practices include the creation of diverse and inclusive forums for discussion \cite{frenchminister2019, fjeld2020}, whereas signifiers include the qualitative measurement of social and environmental impact \cite{raji2020, bender2021, morley2020}.

\paragraph{Fairness}
The value of \textit{fairness} represents a complex concept that accepts multiple definitions \cite{kearns2019, bakalar2021}, some of which cannot be satisfied simultaneously \cite{kearns2019, hidalgo2021, harrison2020}. Overall, we will understand \textit{fairness} in terms of parity in output \cite{dodge2019} and equal treatment \cite{krafft2020} among individuals. When addressing more specific definitions of \textit{fairness} (1-8) ---table \ref{tab:examplefairness}---, we will adopt the approach followed by Verma et al. \cite{verma2018}, which was also echoed by Mehrabi et al. \cite{mehrabi2021}. ML techniques generally conceive fairness in terms of statistical metrics \cite{GreenHu2018} and observe whether specific quantifiable indicators are above or below the thresholds set for a certain application. Even if error rates were equal across groups for a certain application, if those rates are too high, the system could still be considered unfair \cite{harrison2020}. This means that for our value-based framework we outline the quantifiable indicators that are normally used for manifesting fairness-related criteria, but we do not determine the threshold for these indicators to be considered good enough for a specific application. Similarly, the quantifiable indicators relate to the output of the system, rather than the outcome that these outputs lead to. \\

\setlist[itemize]{wide=0pt, noitemsep, label=\textbullet, topsep=2pt, leftmargin=*, after =\vspace*{-\baselineskip}}
\setlist[enumerate]{wide=0pt, noitemsep , topsep=2pt, leftmargin=*, after =\vspace*{-\baselineskip}}
\begin{singlespace}
\begin{longtable}{p{.025\textwidth} m{.115\textwidth} p{.35\textwidth} p{.41\textwidth}} \hline
\footnotesize
& Value & Criteria    & Manifestations \\ \hline \endfirsthead
\multirow{12}{*}{\rotatebox[origin=c]{90}{Universalism}} & \multirow{12}{*}{Fairness}                    & \mbox{}\par\vspace{-\baselineskip}  \begin{enumerate} \item \textbf{Individual fairness} \footnote{Similar individuals should be treated in a similar way. Diverging definitions state that: two individuals that are similar with respect to a common metric should receive the same outcome (\textit{fairness through awareness}); or any protected attribute should not be used when making a decision (\textit{fairness through unawareness}); or the outcome obtained by an individual should be the same if this individual belonged to a counterfactual world or group (\textit{counterfactual fairness}) \cite{mehrabi2021}. }\cite{mehrabi2021, dwork2011, barredoarrieta2020, kusner2017} \item \textbf{Demographic parity} \footnote{The probability of getting a positive outcome should be the same whether the individual belongs to a protected group or not \cite{mehrabi2021}.} \cite{harrison2020, hidalgo2021, srivastava2019, mehrabi2021, dwork2011, kusner2017, barredoarrieta2020, kearns2018, verma2018} \item \textbf{Conditional Statistical parity} \footnote{Given a set of factors L, individuals belonging to the protected or unprotected group should have the same probability of getting a positive outcome \cite{mehrabi2021}.} \cite{mehrabi2021, verma2018} \item \textbf{Equality of opportunity} \footnote{The probability for a person from class A (positive class) of getting a positive outcome, which should be the same regardless of the group (protected group or not) that the individual belongs to \cite{mehrabi2021}.} \cite{mehrabi2021, hardt2016, vanBerkel2021} \item \textbf{Equalized odds} \footnote{The probability for a person from class A (positive class) of getting a positive outcome and the probability for a person from class B (negative class) of getting a negative outcome should be the same \cite{mehrabi2021}.} \cite{mehrabi2021} \item \textbf{Treatment equality} \footnote{The ratio of false positives and negatives has to be the same for both groups \cite{mehrabi2021}.} \cite{mehrabi2021, berk2017} \item \textbf{Test fairness} \footnote{For any probability score S, the probability of correctly belonging to the positive class should be the same for both the protected and unprotected group \cite{mehrabi2021}.}\cite{mehrabi2021, verma2018, chouldechova2016} \item \textbf{Procedural fairness} \footnote{It deals with the fairness of the decision-making process that leads to the outcome in question \cite{grgic-hlaca2018}.} \cite{mehrabi2021, grgic-hlaca2018, kusner2017}  \end{enumerate} & \mbox{}\par\vspace{-\baselineskip}  \begin{itemize} \item \textcolor{magenta}{Accuracy across groups (for classification, sum of true positive and true negative rates)} \cite{kleinberg2016, chouldechova2016, harrison2020, morley2020} \item \textcolor{magenta}{False positive and negative rates across groups} \cite{kleinberg2016, chouldechova2016, mehrabi2021, wang2020, saleiro2018} \item \textcolor{magenta}{False discovery and omission rates across groups} \cite{mitchell2019, saleiro2018} \item \textcolor{magenta}{Pinned AUC} \cite{mitchell2019, dixon2018} \item \textcolor{olive}{Debiasing algorithms} \cite{bellamy2018} \item \textcolor{olive}{Election of protected classes based on user considerations} \cite{grgic-hlaca2018}\end{itemize} \\\hline

\caption{Criteria and their manifestations for \textit{fairness}.}\label{tab:examplefairness}
\end{longtable}
\end{singlespace}

\paragraph{Non-discrimination}
The value of \textit{non-discrimination}, as defined in our framework ---table \ref{tab:examplenondiscrim}---, deals with algorithmic systems not being socially biased \cite{bihrane2021} and ensuring that equal accessibility is provided to all individuals \cite{morley2020}. This means that (1) quality and integrity of data should be evaluated and ensured \cite{hidalgo2021, fjeld2020, stuartgeiger2020, paullada2020, morley2020} in order to prevent ``socially constructed biases, inaccuracies, errors, and mistakes'' \cite{morley2020} from being present in the data. Processes that safeguard inclusive data generation \cite{chasalow2021, stuartgeiger2020, krafft2020, morley2020} and analysis procedures for identifying potential biases in data and for assessing its quality \cite{hidalgo2021, fjeld2020, stuartgeiger2020, paullada2020, morley2020} are strategies that avoid social stereotypes being codified, maintained and amplified \cite{hidalgo2021}. Furthermore, non-discriminatory systems should (2) ensure diversity and inclusiveness in the design process \cite{fjeld2020, europeancommissionEthicsAI2019, morley2020}. From a process-oriented perspective, participants involved in the development process should, thus, present diverse profiles \cite{fjeld2020, zhouWeb, lee2019, krafft2020}. Finally, giving (3) equal access to the technology \cite{bihrane2021, krafft2020, morley2020, fjeld2020} avoids the growth of inequalities as a consequence of deploying AI systems \cite{fjeld2020}.   

\setlist[itemize]{wide=0pt, noitemsep, label=\textbullet, topsep=2pt, leftmargin=*, after =\vspace*{-\baselineskip}}
\setlist[enumerate]{wide=0pt, noitemsep , topsep=2pt, leftmargin=*, after =\vspace*{-\baselineskip}}
\begin{singlespace}
\begin{longtable}{p{.025\textwidth} m{.115\textwidth} p{.35\textwidth} p{.41\textwidth}} \hline
\footnotesize
& Value & Criteria    & Manifestations \\ \hline \endfirsthead

\multirow{6}{*}{\rotatebox[origin=c]{90}{Universalism}}& \multirow{6}{=}{Non-discrimination}          & \multirow{6}{=}{\begin{enumerate} \item \textbf{Quality and integrity of data} \cite{hidalgo2021, fjeld2020, stuartgeiger2020, paullada2020, morley2020} \item \textbf{Inclusiveness in design} \cite{fjeld2020, europeancommissionEthicsAI2019, morley2020} \item \textbf{Accessibility} \cite{bihrane2021, krafft2020, morley2020, fjeld2020}\end{enumerate}}   & 
\mbox{}\par\vspace{-\baselineskip} \begin{itemize} \item \textcolor{olive}{Inclusive data generation process} \cite{chasalow2021, stuartgeiger2020, krafft2020, morley2020} \item \textcolor{magenta}{Analysis of data for potential biases, data quality assessment} \cite{krafft2020, mehrabi2021, fjeld2020, hidalgo2021, gebru2020} \item \textcolor{olive}{Diversity of participant in development process} \cite{fjeld2020, zhouWeb, lee2019, krafft2020} \item \textcolor{olive}{Access to code and technology to all} \cite{fjeld2020, bihrane2021, krafft2020, morley2020}\end{itemize}  \\ \hline

\caption{Criteria and their manifestations for \textit{non-discrimination}.}\label{tab:examplenondiscrim}
\end{longtable}
\end{singlespace}

\subsubsection{Individual empowerment.} Contestability, human control and human agency address the politics behind algorithmic systems \cite{arnstein2019, winner1980} and deal with the issues caused by power imbalances \cite{chasalow2021, kalluri2020, bihrane2021, lyons2021, vaccaro2020}.

\paragraph{Contestability}
The value of \textit{contestability} ---table \ref{tab:examplecontestab}--- is defined as the value that ensures that users have the necessary information to (1) enable argumentation against conclusions reached by algorithmic systems \cite{balayn2021Edri, kyunglee2017, alfrink2020, fjeld2020, kalluri2020, royalsociety2019, lyons2021, europeancommissionEthicsAI2019}. This involves (2) empowering citizens \cite{balayn2021Edri, kalluri2020, europeancommissionEthicsAI2019} to investigate and influence AI \cite{kalluri2020}, as part of a broader regulatory approach \cite{lyons2021}. As a matter of fact, \textit{contestability} has been identified as a ``critical aspect of future public decision-making systems'' \cite{alfrink2020}. This implies that, from a documentation perspective (signifiers), users should be made aware of who determines what constitutes a contestable decision, who is accountable for it and who can contest a decision. This last point is particularly necessary to determine whether (legal) representatives of decision subjects can act on their behalf. The review mechanism in place and the workflow of contestations \cite{lyons2021} are policy-related details that users should also be informed about. From a process-oriented standpoint, mechanisms for users to ask questions and to record disagreements should also be put in place \cite{hirsch2017, mitra2021}.

\setlist[itemize]{wide=0pt, noitemsep, label=\textbullet, topsep=2pt, leftmargin=*, after =\vspace*{-\baselineskip}}
\setlist[enumerate]{wide=0pt, noitemsep , topsep=2pt, leftmargin=*, after =\vspace*{-\baselineskip}}
\begin{singlespace}
\begin{longtable}{p{.025\textwidth} m{.115\textwidth} p{.35\textwidth} p{.41\textwidth}} \hline
\footnotesize
& Value & Criteria    & Manifestations \\ \hline \endfirsthead

\multirow{11}{*}{\rotatebox[origin=c]{90}{Individual empowerment}}& \multirow{11}{*}{Contestability} & \mbox{}\par\vspace{-\baselineskip} \begin{itemize} \item[\textcolor{white}{\textbullet}] \item[\textcolor{white}{\textbullet}] \item[\textcolor{white}{\textbullet}] \item[\textcolor{white}{\textbullet}] \end{itemize} \begin{enumerate} \item \textbf{Enable argumentation / negotiation against a decision} \cite{balayn2021Edri, kyunglee2017, alfrink2020, fjeld2020, kalluri2020, royalsociety2019, lyons2021, europeancommissionEthicsAI2019} \item \textbf{Citizen empowerment} \cite{balayn2021Edri, kalluri2020, europeancommissionEthicsAI2019}\end{enumerate}      & \mbox{}\par\vspace{-\baselineskip} \begin{itemize} \item \textcolor{orange}{Information of who determines and what constitutes a contestable decision and who is accountable} \cite{lyons2021} \item \textcolor{orange}{Determination of who can contest the decision (subject or representative)} \cite{lyons2021} \item \textcolor{orange}{Indication of type of review in place} \cite{lyons2021} \item \textcolor{orange}{Information regarding the contestability workflow} \cite{lyons2021} \item \textcolor{olive}{Mechanisms for users to ask questions and record disagreements with system behavior} \cite{hirsch2017, mitra2021}\end{itemize}               \\\hline

\caption{Criteria and their manifestations for \textit{contestability}.}\label{tab:examplecontestab}
\end{longtable}
\end{singlespace}

\paragraph{Human Control}
The value of \textit{human control} ---table \ref{tab:examplecontrol}--- addresses the influence that data-driven technologies have over humans and that leads to a reduction of human agency, power and control \cite{patel2021}. Algorithmic systems should be controllable \cite{bihrane2021} and (1) subject to user and collective influence \cite{bihrane2021, kyunglee2017}. They should also be (2) subject to human review \cite{fjeld2020}. Governance mechanisms that ensure human oversight of automated decisions are, thus, necessary to maintain control and influence over such systems \cite{morley2020}. It should be possible to (3) choose how and even whether (in the very first place) to delegate a decision to an automated system \cite{fjeld2020}. From a development perspective, levels of human discretion should be established \cite{europeancommissionEthicsAI2019, microsoftai2018} and the ability to override decisions made by a system \cite{europeancommissionEthicsAI2019} ought to be set up by design. Once the system is deployed, it should be continuously monitored to enable adequate intervention when necessary \cite{fjeld2020, teliaai2019, europeancommissionEthicsAI2019}. 

\setlist[itemize]{wide=0pt, noitemsep, label=\textbullet, topsep=2pt, leftmargin=*, after =\vspace*{-\baselineskip}}
\setlist[enumerate]{wide=0pt, noitemsep , topsep=2pt, leftmargin=*, after =\vspace*{-\baselineskip}}
\begin{singlespace}
\begin{longtable}{p{.005\textwidth} p{.02\textwidth} m{.115\textwidth} p{.35\textwidth} p{.4\textwidth}} \hline
\footnotesize
& & Value & Criteria    & Manifestations \\ \hline \endfirsthead

\multirow{6}{*}{\rotatebox[origin=c]{90}{Individual}}& \multirow{6}{*}{\rotatebox[origin=c]{90}{ empowerment}} & \multirow{6}{=}{Human Control}               &  \mbox{}\par\vspace{-\baselineskip} \begin{enumerate} \item \textbf{User/collective influence} \cite{bihrane2021, kyunglee2017} \item \textbf{Human review of automated decision} \cite{fjeld2020} \item \textbf{Choice of how and whether to delegate} \cite{fjeld2020}\end{enumerate}  & \mbox{}\par\vspace{-\baselineskip} \begin{itemize} \item \textcolor{olive}{Continuous monitoring of system to intervene} \cite{fjeld2020, teliaai2019, europeancommissionEthicsAI2019} \item \textcolor{olive}{Establishment levels of human discretion during the use of the system} \cite{europeancommissionEthicsAI2019, microsoftai2018} \item \textcolor{olive}{Ability to override the decision made by a system} \cite{europeancommissionEthicsAI2019}\end{itemize}\\ \hline

\caption{Criteria and their manifestations for \textit{human control}.}\label{tab:examplecontrol}
\end{longtable}
\end{singlespace}

\paragraph{Human Agency}
The value of \textit{human agency} ---table \ref{tab:exampleagency}--- deals with the risks of algorithmic systems displacing human autonomy \cite{europeancommissionEthicsAI2019, fjeld2020}. As claimed by Cila et al. \cite{cila2020}, algorithmic systems may displace human agency in governance processes and may undermine human autonomy. ML-driven systems advocating for human agency should, therefore, (1) respect human autonomy \cite{europeancommissionEthicsAI2019, fjeld2020, morley2020} and (2) citizens' power to decide \cite{bihrane2021, europeancommissionEthicsAI2019}. In addition, (3) decision subjects should be able to opt out of an automated decision \cite{fjeld2020, europeancommissionEthicsAI2019}. The manifestations of such criteria involve giving knowledge and tools to users to comprehend and interact with AI systems \cite{europeancommissionEthicsAI2019} (signifier) and, from a process-oriented perspective, providing strategies for users to self-assess the systems \cite{europeancommissionEthicsAI2019}. 

\setlist[itemize]{wide=0pt, noitemsep, label=\textbullet, topsep=2pt, leftmargin=*, after =\vspace*{-\baselineskip}}
\setlist[enumerate]{wide=0pt, noitemsep , topsep=2pt, leftmargin=*, after =\vspace*{-\baselineskip}}
\begin{singlespace}
\begin{longtable}{p{.005\textwidth} p{.02\textwidth} m{.115\textwidth} p{.35\textwidth} p{.4\textwidth}} \hline
\footnotesize
& & Value & Criteria    & Manifestations \\ \hline \endfirsthead
\multirow{7}{*}{\rotatebox[origin=c]{90}{Individual}} & \multirow{7}{*}{\rotatebox[origin=c]{90}{empowerment}} & \multirow{7}{=}{Human agency}                & \mbox{}\par\vspace{-\baselineskip}  \begin{enumerate}\item \textbf{Respect for human autonomy} \cite{europeancommissionEthicsAI2019, fjeld2020,  morley2020} \item \textbf{Power to decide. Ability to make informed autonomous decision} \cite{bihrane2021, europeancommissionEthicsAI2019} \item \textbf{Ability to opt out of an automated decision} \cite{fjeld2020, europeancommissionEthicsAI2019}\end{enumerate}               & \mbox{}\par\vspace{-\baselineskip}  \begin{itemize} \item \textcolor{orange}{Give knowledge and tools to comprehend and interact with AI system} \cite{europeancommissionEthicsAI2019} \item \textcolor{olive}{Opportunity to self-assess the system} \cite{europeancommissionEthicsAI2019}\end{itemize}\\ \hline 

\caption{Criteria and their manifestations for \textit{human agency}.}\label{tab:exampleagency}
\end{longtable}
\end{singlespace}

\paragraph{Selecting values, criteria and manifestations for our example use case.} Returning  to the hypothetical insurance modelling team from our motivating example (section \ref{frameworkdetails}), they decided to apply our value-based framework before launching their system. They quickly realised that they need to consider more values than those outlined in current auditing processes. For example transparency, non-discrimination, supporting human agency and the public good. They also discovered a range of methods for enacting those values: from data handling processes that ensure anonymity and meaningful consent around the model, to models of fairness appropriate to their case. 

Although we cover prominent ethical principles in AI and the assessment of the algorithmic system might include all of them, here we focus on a subset of those values for illustrative purposes. We imagine that the researchers developing the algorithmic life insurance application system want to focus on  \textit{explainability} and \textit{privacy} (fig \ref{fig:framework_diagram}). We  assume that they are dealing with a blackbox algorithm that is not interpretable by design. The team  needs to examine whether the algorithmic system and the decision reached are understandable. Additionally, the deployed XAI methods should enable traceability and evaluation of the system. As far as the \textit{explainability} manifestations are concerned, since they are dealing with a blackbox algorithm, they need to deploy adequate post-hoc explanations. When it comes to \textit{privacy}, the data used for training and testing the algorithmic model should have been obtained through the explicit approval of the decision subjects. These subjects should have been informed about the nature and purpose of the data that is collected, the way this data is handled and stored. Decision subjects should also have agency and control over their data. Additionally, data protection mechanisms should have been implemented to make sure that there is no possibility of identifying sensitive (in this case medical) data about the subjects. These two values that the team needs to advocate for, represent some trade-offs: XAI methods uphold interpretability of algorithmic systems and some of them even rely on comparing data instances at inference time with those used for training the system. This would directly violate the subjects' right to have their data protected and confidentiality ensured. 

\section{Towards a multi-stakeholder critical reflection of algorithmic systems} \label{stakeholders}

Since there are value trade-offs, like the one outlined in our example use case, and certain value-specific criteria are mutually exclusive, we follow the claim made by Raji and Smart \cite{raji2020}, and advocate for standpoint diversity. This implies involving a wide range of stakeholders in the negotiation process \cite{Shklovski2022} to discuss and critically reflect on the degree to which each of the values should be promoted in detriment of the other one and how the prioritization process should take place. These stakeholders will possess different types of knowledge and will present different insight needs. In this section we map those stakeholders and match them with the most suitable communication means.

\subsection{Methodology for identifying relevant stakeholders and communication means} \label{methodmapping}

To identify relevant stakeholders, we follow the stakeholder characterization of Suresh et al. \cite{suresh2021}. They classified stakeholders in a two dimensional matrix, where one dimension captured the nature of the knowledge of the stakeholders (formal, instrumental or personal) and the second one identified the context in which that knowledge manifests (Machine Learning, data domain, and the general milieu). Formal knowledge entails a deep understanding of the theories of a certain domain. Instrumental knowledge refers to the capability of applying formal knowledge in one of the three contexts. Personal knowledge is acquired by the participation of the subject in a specific context. The two dimensional-matrix classification results in nine different stakeholder profiles. To facilitate the process of mapping the stakeholders to tailored communication means, we narrow those stakeholders down into four categories \footnote{This reduced classification is backed up by the framework employed by Barredo-Arrieta et al. \cite{barredoarrieta2020} when identifying the explainability needs of various stakeholders.}.

We then proceed to identify the means to communicate system-related information to different stakeholders. We searched  such means using arXiv and Google search, so as to cover the state of the art in terms of research papers and open source toolkits. Each search referred to specific value criteria and  manifestations, although many of the found means address more than one value. This review does not intend to be exhaustive. We expect novel research to address value manifestations that still present scarce resources in our framework. Hence our review is just a snapshot of some of the available communication means until January 2022, but we host the latest version on an online repository \footnote{\url{https://github.com/mireiayurrita/valuebasedframework}} and is open to anyone's contribution. We aim at creating a living document that will keep growing and that will address current research gaps as time goes by.

\subsection{Mapping stakeholders}

We characterize four main stakeholders in our framework ---table \ref{tab:mappingstakeholders}---: (1) The development team: they have the formal, instrumental and personal knowledge in the domain of ML \cite{suresh2021}. They want to ensure and improve product efficiency and research new functionalities \cite{barredoarrieta2020}. (2) Auditing team: they have the formal and instrumental knowledge of the general milieu, meaning that they are aware of the social theories behind AI, and are able to evaluate technical specifications of ML systems. They aim at verifying model compliance with legislation \cite{barredoarrieta2020} (3) Data domain experts: they have the theoretical (formal) and instrumental knowledge of the application context (healthcare, economics etc.). They look forward to gaining scientific or domain-specific knowledge \cite{barredoarrieta2020, suresh2021}, trust the model \cite{barredoarrieta2020, suresh2021} and act based on the model output \cite{suresh2021}. And (4) Data subjects: they have the personal knowledge of the data domain in which the AI is being applied and the general milieu. They aim at understanding their situation \cite{barredoarrieta2020}, verifying that the decision is fair \cite{barredoarrieta2020}, contesting the decision (if needed) \cite{suresh2021} and understanding how their data is being used \cite{suresh2021}. 

\setlist[itemize]{wide=0pt, noitemsep, label=\textbullet, topsep=2pt, leftmargin=*, after =\vspace*{-\baselineskip}}
\begin{singlespace}
\begin{longtable}{p{.15 \textwidth} p{.15 \textwidth} p{.3\textwidth} p{.3\textwidth} } \hline
    \textbf{Stakeholder} & \textbf{Mapping} \cite{suresh2021} & \textbf{Nature of knowledge} & \textbf{Purpose of insight}\\ \hline
    \multirow{5}{=}{Development team} & \multirow{5}{=}{ML, Formal + Instrumental + Personal} & \mbox{}\par\vspace{-\baselineskip} \begin{itemize} \item ``Knowledge of the math behind the architecture'' \cite{suresh2021} \item ``Stakeholder involved in an ex-ante impact assessment of the automatic decision system''\cite{henin2021}\end{itemize} & \mbox{}\par\vspace{-\baselineskip} \begin{itemize} \item Ensure/improve product efficiency and debug \cite{barredoarrieta2020} \item Research new functionalities \cite{barredoarrieta2020} \end{itemize}  \\ \hline
    \multirow{4}{=}{Auditing team} & \multirow{4}{=}{Milieu, Formal + Instrumental} & \mbox{}\par\vspace{-\baselineskip} \begin{itemize} \item ``Familiarity with broader ML-enabled systems'' \cite{suresh2021} \item ``Experts who intervene wither upstream or downstream'' \cite{henin2021} \end{itemize}  & \mbox{}\par\vspace{-\baselineskip} \begin{itemize} \item Verify model compliance with legislation \cite{barredoarrieta2020}\end{itemize} \\ \hline
    \multirow{5}{=}{Data domain experts} & \multirow{5}{=}{Data domain,  Formal + Instrumental} & \mbox{}\par\vspace{-\baselineskip}\begin{itemize} \item ``Theories relevant to the data domain'' \cite{suresh2021} \item ``Professional involved in the operational phase of the automatic decision system'' \cite{henin2021} \end{itemize} & \mbox{}\par\vspace{-\baselineskip} \begin{itemize} \item Gain scientific or domain-specific knowledge \cite{barredoarrieta2020, suresh2021} \item Trust the model \cite{barredoarrieta2020, suresh2021} \item Act based on the output \cite{suresh2021} \end{itemize} \\ \hline
    \multirow{5}{=}{Decision subjects} & \multirow{5}{=}{Data domain + Milieu, Personal} & \mbox{}\par\vspace{-\baselineskip} \begin{itemize} \item ``Lived experience and cultural knowledge'' \cite{suresh2021} \item ``Layperson affected by the outcomes of the automatic decision system'' \cite{henin2021} \end{itemize} & \mbox{}\par\vspace{-\baselineskip} \begin{itemize} \item Understand their situation \cite{barredoarrieta2020} \item Verify fair decision \cite{barredoarrieta2020} \item Contest decision \cite{suresh2021} \item Understand how one's data is being used \cite{suresh2021} \end{itemize} \\\hline
    
    \caption{Description of potential stakeholders that can be brought together as part of our value-based framework. These stakeholders have been mapped following the two dimensional criteria (type of knowledge ---formal, instrumental or personal--- and contexts in which this knowledge manifests ---ML, data domain, milieu---) outlined by Suresh et al. \cite{suresh2021}. The nature of their knowledge and the purpose of gaining insight for each of them have also been defined.} \label{tab:mappingstakeholders}   
\end{longtable}
\end{singlespace}

\paragraph{Mapping stakeholders in our example use case}

Going back to our example, once \textit{explainability} and \textit{privacy} have been broken down into specific criteria manifestations, the team needs to map the stakeholders who will take part in the assessment process (fig \ref{fig:framework_diagram}). The development team represents the stakeholders who have the knowledge of the math behind the system. An external auditing team will join the discussion to make sure that the model is aligned with current legislation. Since the algorithmic life insurance application system deals with medical data, the data domain experts will be represented by a medical team and a life insurance expert. Decision subjects will be laypeople who seek to understand and verify their situation with regards to data usage and the decision reached by the system.

\subsection{Mapping tailored communication means}
We then examine each of the reviewed means and identify their typology, the value manifestations that they cover and the stakeholders that can make use of it, as illustrated in table \ref{tab:exampleprivacymeans} for privacy dashboards. The objective of mapping value manifestations, stakeholder profiles and communication means is that of enabling a fruitful and informed discussion among stakeholders. We classify these means in three categories: (1) \textit{ \textcolor{red}{Descriptive documents}} (red), (2) \textit{\textcolor{bleudefrance}{Design strategies}} (blue), and (3) \textit{\textcolor{darkpastelgreen}{Ready-to-use tools}} (green). \textbf{Appendix \ref{appC} summarizes the rest of the communication means we identified and maps them to value manifestations and stakeholders for whom such methods are suitable.}

The stakeholders assigned to a specific communication means are based on the audience addressed by the original authors of such methodologies. In some cases, the characterization of the intended audience was not as granular as our stakeholder mapping and the authors merely differed experts in ML from non-experts. Based on the nature of knowledge that we assigned to each of the mentioned stakeholders in section \ref{methodmapping}, we considered that the development and auditing teams are able to understand technically formulated system details (experts) whereas data domain experts and decision subjects would require more accessible communication means (non-experts). Similarly, some of the communication means identified for \textit{explainability} are suitable for any stakeholder, but the original authors formulated the post-hoc explanations with varying degrees of complexity, which should be taken into account when trying to deploy such strategies. If the target audience are data subjects, we echo van Berkel et al. \cite{vanBerkel2021} and Cheng et al. \cite{cheng2019} and recommend to limit presentation complexity and to instruct participants throughout the session. 
\begin{singlespace}
\begin{longtable}{p{.02\textwidth} p{.09\textwidth} p{.07\textwidth}  p{.18\textwidth}  p{.02\textwidth} p{.02\textwidth} p{.02\textwidth} p{.02\textwidth} p{.1\textwidth} p{.1\textwidth} p{.1\textwidth}} \hline
&\multirow{2}{=}{\textbf{Means}} & \multirow{2}{=}{\textbf{Value}} & \multirow{2}{=}{\textbf{Manifestation(s)}} & \multicolumn{4}{c}{\textbf{Stakeholder}} & \multirow{2}{=}{\textbf{Application} (model) }& \multirow{2}{=}{\textbf{Approach}} & \multirow{2}{=}{\textbf{Visual elements}} \\ \cline{5-8}
& & & & DT & AT & DE & DS & & &  \\ \hline 
\multirow{9}{=}{\textcolor{darkpastelgreen}{[B]}} & \small \multirow{9}{=}{Privacy dashboards \cite{zimmerman2014, earp2016, fischerhubner2016, herder2020, farke2021}} & \small \multirow{5}{=}{Privacy} & \small \mbox{}\par\vspace{-\baselineskip} \begin{itemize}  \item Description of what and why data is collected \item Description of how data is handled \end{itemize} & & & & \small \multirow{8}{=}{\checkmark} & 
\small \multirow{8}{=}{Agnostic} & & \multirow{8}={ \small \begin{itemize}  \item Timelines \item  Bar charts \item Maps \item Network graphs \end{itemize}}  \\\cline{3-4}
& &\small   Human agency & \small \mbox{}\par\vspace{-\baselineskip} \begin{itemize}  \item Self-assessment of the system \end{itemize} & & & & & & &   \\ \cline{3-4}
& &\small   Trans-parency & \small \mbox{}\par\vspace{-\baselineskip} \begin{itemize}  \item Disclosure of properties of data  \end{itemize} & & & & & & &   \\\hline

\caption{Illustration of how we mapped communication means with values, manifestations and stakeholders (DT = Development Team; AT = Auditing Team; DE = Data Domain Experts; DS = Decision Subjects). Privacy dashboards are tools (green) that allow users to interactively assess the collection and usage of their data. The rest of the reviewed communication means are characterized in appendix \ref{appC}.}\label{tab:exampleprivacymeans}

\end{longtable}
\end{singlespace}

It should be noted that this mapping process represents a first step to making a wide range of stakeholders with different backgrounds understand each other. We are aware that communicating system-related information in a tailored way does not directly lead to the resolution of value trade-offs, and that design strategies are necessary for facilitating such conversations \cite{heijne2019}. In any case, the exercise of resolving value tensions should be a communicative process, rather than a simple explanation \cite{ohara2020}. However, the means used for communicating specifications of the system will play a key role in the dynamics that will take place in those sessions. 

\paragraph{Assigning communication means to each stakeholder in our example use case.}

The life insurance researchers are now looking into appropriate methods for communicating values to different stakeholders (fig \ref{fig:framework_diagram}), so that they can develop a comprehensive plan that ensures both compliance and communication of values. 

Based on the mapping presented in appendix \ref{appC}, tables \ref{tab:tailoredmeanscodes} and \ref{tab:detailedmeans}, for the value of \textit{explainability} and its manifestations in the form of post-hoc explanations, the team can use various design strategies and tools as part of their assessment process. To facilitate the navigation of table \ref{tab:detailedmeans}, they first examine table \ref{tab:tailoredmeanscodes} to locate the type of means (tool, strategy, or documentation), values and stakeholders they are interested in. Once they select the codes associated to each communication means, they check table \ref{tab:detailedmeans} to see whether the value manifestations in question are addressed and to  explore the selected communication means. If the team working on the life insurance case prefers a ready-to-use tool over the description of design strategies for assessing \textit{explainability}, they can use InterpretML \cite{nori2019} and especially the DiCE  \cite{mothilal2019} functionality, (code \textcolor{darkpastelgreen}{[AC]}) with the development and auditing teams to evaluate counterfactual examples. These counterfactual examples tell  how input features should change in order for the output of the system to be different. That is to say, how the individual applying for life insurance should be different, physically, or when it comes to insurance or medical history, for them to accept the application (if the original output was a refusal). However, this tool might not be suitable for non-experts who are not familiar with ML-related concepts. In the life insurance use case, the medical and insurance team and the decision subjects should receive a description of how the output changes if a feature is perturbed, absent or present adapted to their insight needs. This can be done by describing the answers to the questions ``Why, Why not and How to be that'' for a certain output \cite{veraliao2020} (code \textcolor{bleudefrance}{[P]}). As for \textit{privacy} manifestations, the development and auditing teams can examine data collection and storage specifications through the Datasheet \cite{gebru2020} associated to the dataset in question (code \textcolor{red}{[K]}). Special attention should be paid to the ``Collection'' and ``Preprocessing/cleaning/labelling'' sections. For decision subjects, iconsets \cite{GDPR2018, rossi2017, holtz2011, mehldau2007} (code \textcolor{bleudefrance}{[A]}) and privacy dashboards \cite{zimmerman2014, earp2016, fischerhubner2016, herder2020, farke2021} (code \textcolor{darkpastelgreen}{[B]}) are means for them to explore how their data is being used. It should be noted that the cell that intersects between data domain experts and \textit{privacy} in table \ref{tab:tailoredmeanscodes} is blank. Based on the characterization of stakeholders that we provided, data privacy-related matters are not directly linked to the purpose that data domain experts show when willing to explore algorithmic systems. This is translated into scarcity of methodologies related to \textit{privacy} manifestations that directly  address data domain experts.

\section{Discussion and future work} \label{limitations}
We discuss important aspects of our framework below.

\textit{Design choices for creating a value-based framework.}
We aim at examining values that characterize ML-driven systems rather than the organizations responsible for these systems. Hence, we did not integrate accountability or responsibility as a value \textit{per se} in our framework. We are aware that algorithms cannot be held responsible for the potential harm that they might cause \cite{bryson2017, hidalgo2021}, and that in order to effectively deploy such systems, there is an urgent call for accountability \cite{yazdanpanah2021, alfrink2020}. Likewise, we are aware of the need for rigorous frameworks that support accountability \cite{hutchinson2021} and we consider that the act of conceiving an assessment framework itself answers to the need to evaluate and audit algorithmic systems \cite{fjeld2020}. Nevertheless, we did not explicitly highlight the profiles of the people accountable for the system. We decided to follow Zhu et al. \cite{zhu2021} and considered accountability as a governance issue. We do, however, believe that entities up the chain of command should be held accountable for the potential harm caused by algorithmic systems \cite{lyons2021, hidalgo2021, schaerer2009, krafft2020, fjeld2020}. 
It should also be noted that values and criteria presented in this paper might not be unique \cite{Shklovski2022}. 
We acknowledge current discussions in VSD about the shortcomings of pre-selecting values \cite{davis2015} and, hence, do not claim  universality. Extension and modification of values is possible in our framework, but are subject to respecting continuity and opposition between values. Similarly, criteria and manifestations can be extended and subsets could be included to create situationally-specific versions of the framework. Since the aim of our framework is to encourage critical reflection \cite{frauenberger2016, ustek-spilda2019} and we identified some value manifestations that require additional communication means, we particularly encourage those context-specific adaptations to happen. Under no circumstances should the scarcity of communication means for certain values identified in our framework represent an excuse to justify inaction or to ignore such values.

\textit{Context dependence and consistency.} As echoed by Liscio et al. \cite{liscio2021}, in order to translate values into system requirements \cite{pommeranz2012, vandepoel2013}, to reason about conflicting values \cite{ajmeri2020,Murukannaiah2014} and to communicate them to different stakeholders \cite{raaij1994}, it is necessary to situate these values within a context. The prioritization of values  depends on the application context of such systems \cite{krafft2020}. In this paper, we showed an example of how the framework could be applied to a particular use case. However, considering the differences between value alignments and tensions that may arise due to context dependence, the validity and consistency of our framework is still to be tested. Future work
needs to validate our framework across scenarios \cite{vanBerkel2021, ghassemi2021} through user studies or synthetic experiments \cite{sokol2020}.

\textit{Need for standardization.} To systematically review and revisit value priorities and tensions among different stakeholders, our framework should be part of a broader evaluation workflow \cite{lee2021}, such as the one suggested by Raji and Smart \cite{raji2020}. Besides, 
practices from software engineering such as the Values Dashboard \cite{nurwidyantoro2021} could be adopted \cite{schneiderman2020, thew2017}. This dashboard promotes awareness of values and aims at triggering discussions among stakeholders. It claims to be beneficial in each phase of the software development process, from inception to release, and establishes strategies, such as Timelines or Issues, that are already common practice on software development platforms like Github.

\textit{Implications of our work.}  Our multi-stakeholder value-based framework facilitates the unveiling of assumptions that encode political and social values made by developers \cite{raji2020}. By bringing together a wide range of stakeholders to evaluate and discuss value manifestations, one can anticipate and remedy harmful algorithmic behaviors before deploying a system. 
Besides, we provide researchers and industry practitioners with a good coverage of values to evaluate their systems and the association of such values to actionable value manifestations. This contributes greatly to the adoption of ethical approaches by practically-minded people \cite{morley2020}. For researchers, we provide them with an easy-to-navigate mapping of value manifestations, stakeholders and communication means. Our framework also visually illustrates research gaps that need to be addressed. Blank spaces in appendix \ref{appC} or values with a scarce number of associated communication means directly refer to valuable research opportunities. For instance, for the value of \textit{fairness}, a great deal of effort has been devoted to designing ready-to-use tools for stakeholders with a deep understanding of ML (developers and auditing teams). However, means for addressing \textit{fairness} manifestations and communicating them to decision subjects have not received the same attention. For industry practitioners, we gathered ready-to-use open source toolkits (appendix \ref{appC}) that can be directly applied to their own use cases. Moreover, since we host this mapping on an online repository \footnote{Check footnote 30}  open to future contributions, we hope that the number of tools addressing each of the identified value manifestations will grow and that the benefits of designing such a framework will be even more tangible in the future.

\section{Conclusions} \label{conclusions}

In this paper, we designed a value-based framework for assessing algorithmic systems from a multi-stakeholder perspective. This provides investigators of algorithmic systems with an actionable set of criteria manifestations to operationalize high-level ethical principles. We arranged eleven prominent values of ML-driven systems in a circular composition, so that common motivations and trade-offs can be easily identified. 

We then broke down each of these values into a set of criteria and their correspondent manifestations in the form of quantifiable indicators, process-oriented practices, and signifiers. In addition, we examined available tools for communicating those value manifestations to different stakeholders based on the nature of their knowledge and their insight needs. This should enable to bring a wide range of stakeholders together to systematically assess values encoded in a system and facilitate value- and ethics-related discussions among them. This work completes previous studies that claim the need for incorporating a diverse range of stakeholders and viewpoints in the ML workflow, so that conflicting priorities and value tensions can be reviewed, negotiated and consolidated.

\begin{acks}
We thank Kars Alfrink for valuable feedback on an earlier version of this paper and our colleagues at the DCODE Network for insightful discussions.

This project has received funding from the European Union’s Horizon 2020 research and innovation programme under the Marie Skłodowska‐Curie grant agreement No 955990.
\end{acks}

\bibliographystyle{format/ACM-Reference-Format}
\bibliography{references}


\begin{thebibliography}{192}

%%% the Web address.

\ifx \showCODEN    \undefined \def \showCODEN     #1{\unskip}     \fi
\ifx \showDOI      \undefined \def \showDOI       #1{#1}\fi
\ifx \showISBNx    \undefined \def \showISBNx     #1{\unskip}     \fi
\ifx \showISBNxiii \undefined \def \showISBNxiii  #1{\unskip}     \fi
\ifx \showISSN     \undefined \def \showISSN      #1{\unskip}     \fi
\ifx \showLCCN     \undefined \def \showLCCN      #1{\unskip}     \fi
\ifx \shownote     \undefined \def \shownote      #1{#1}          \fi
\ifx \showarticletitle \undefined \def \showarticletitle #1{#1}   \fi
\ifx \showURL      \undefined \def \showURL       {\relax}        \fi
% invisible to TeX
\providecommand\bibfield[2]{#2}
\providecommand\bibinfo[2]{#2}
\providecommand\natexlab[1]{#1}
\providecommand\showeprint[2][]{arXiv:#2}

\bibitem[\protect\citeauthoryear{Abebe, Barocas, Kleinberg, Levy, Raghavan, and
  Robinson}{Abebe et~al\mbox{.}}{2020}]%
        {abebe2020}
\bibfield{author}{\bibinfo{person}{Rediet Abebe}, \bibinfo{person}{Solon
  Barocas}, \bibinfo{person}{Jon Kleinberg}, \bibinfo{person}{Karen Levy},
  \bibinfo{person}{Manish Raghavan}, {and} \bibinfo{person}{David~G.
  Robinson}.} \bibinfo{year}{2020}\natexlab{}.
\newblock \showarticletitle{{Roles for computing in social change}}. In
  \bibinfo{booktitle}{\emph{Proceedings of the 2020 Conference on Fairness,
  Accountability, and Transparency}}. \bibinfo{publisher}{ACM},
  \bibinfo{address}{New York, NY, USA}, \bibinfo{pages}{252--260}.
\newblock
\showISBNx{9781450369367}
\urldef\tempurl%
\url{https://doi.org/10.1145/3351095.3372871}
\showDOI{\tempurl}


\bibitem[\protect\citeauthoryear{Adler, Falk, Friedler, Nix, Rybeck,
  Scheidegger, Smith, and Venkatasubramanian}{Adler et~al\mbox{.}}{2018}]%
        {adler2018}
\bibfield{author}{\bibinfo{person}{Philip Adler}, \bibinfo{person}{Casey Falk},
  \bibinfo{person}{Sorelle~A Friedler}, \bibinfo{person}{Tionney Nix},
  \bibinfo{person}{Gabriel Rybeck}, \bibinfo{person}{Carlos Scheidegger},
  \bibinfo{person}{Brandon Smith}, {and} \bibinfo{person}{Suresh
  Venkatasubramanian}.} \bibinfo{year}{2018}\natexlab{}.
\newblock \showarticletitle{{Auditing black-box models for indirect
  influence}}.
\newblock \bibinfo{journal}{\emph{Knowledge and Information Systems}}
  \bibinfo{volume}{54}, \bibinfo{number}{1} (\bibinfo{year}{2018}),
  \bibinfo{pages}{95--122}.
\newblock
\showISSN{0219-3116}
\urldef\tempurl%
\url{https://doi.org/10.1007/s10115-017-1116-3}
\showDOI{\tempurl}


\bibitem[\protect\citeauthoryear{{AI Ethics Impact Group (AIEIG)}}{{AI Ethics
  Impact Group (AIEIG)}}{2020}]%
        {krafft2020}
\bibfield{author}{\bibinfo{person}{{AI Ethics Impact Group (AIEIG)}}.}
  \bibinfo{year}{2020}\natexlab{}.
\newblock \bibinfo{title}{{From Principles to Practice An interdisciplinary
  framework to operationalise AI ethics}}.
\newblock
\newblock
\urldef\tempurl%
\url{https://www.ai-ethics-impact.org/resource/blob/1961130/c6db9894ee73aefa489d6249f5ee2b9f/aieig---report---download-hb-data.pdf}
\showURL{%
\tempurl}


\bibitem[\protect\citeauthoryear{Aizenberg and van~den Hoven}{Aizenberg and
  van~den Hoven}{2020}]%
        {aizenberg2020}
\bibfield{author}{\bibinfo{person}{Evgeni Aizenberg} {and}
  \bibinfo{person}{Jeroen van~den Hoven}.} \bibinfo{year}{2020}\natexlab{}.
\newblock \showarticletitle{{Designing for human rights in AI}}.
\newblock \bibinfo{journal}{\emph{Big Data {\&} Society}} \bibinfo{volume}{7},
  \bibinfo{number}{2} (\bibinfo{date}{7} \bibinfo{year}{2020}),
  \bibinfo{pages}{2053951720949566}.
\newblock
\showISSN{2053-9517}
\urldef\tempurl%
\url{https://doi.org/10.1177/2053951720949566}
\showDOI{\tempurl}


\bibitem[\protect\citeauthoryear{Ajmeri, Guo, Murukannaiah, and Singh}{Ajmeri
  et~al\mbox{.}}{2020}]%
        {ajmeri2020}
\bibfield{author}{\bibinfo{person}{Nirav Ajmeri}, \bibinfo{person}{Hui Guo},
  \bibinfo{person}{Pradeep~K Murukannaiah}, {and} \bibinfo{person}{Munindar~P
  Singh}.} \bibinfo{year}{2020}\natexlab{}.
\newblock \showarticletitle{{Elessar: Ethics in Norm-Aware Agents}}. In
  \bibinfo{booktitle}{\emph{Proceedings of the 19th International Conference on
  Autonomous Agents and MultiAgent Systems}} \emph{(\bibinfo{series}{AAMAS
  '20})}. \bibinfo{publisher}{International Foundation for Autonomous Agents
  and Multiagent Systems}, \bibinfo{address}{Richland, SC},
  \bibinfo{pages}{16--24}.
\newblock
\showISBNx{9781450375184}


\bibitem[\protect\citeauthoryear{Alfrink, Turel, Keller, Doorn, and
  Kortuem}{Alfrink et~al\mbox{.}}{2020}]%
        {alfrink2020}
\bibfield{author}{\bibinfo{person}{Kars Alfrink}, \bibinfo{person}{T. Turel},
  \bibinfo{person}{A.~I. Keller}, \bibinfo{person}{N. Doorn}, {and}
  \bibinfo{person}{G.~W. Kortuem}.} \bibinfo{year}{2020}\natexlab{}.
\newblock \showarticletitle{{Contestable City Algorithms}}.
  \bibinfo{publisher}{International Conference on Machine Learning Workshop}.
\newblock


\bibitem[\protect\citeauthoryear{Alqaraawi, Schuessler, Wei{\ss}, Costanza, and
  Berthouze}{Alqaraawi et~al\mbox{.}}{2020}]%
        {alqaraawi2020}
\bibfield{author}{\bibinfo{person}{Ahmed Alqaraawi}, \bibinfo{person}{Martin
  Schuessler}, \bibinfo{person}{Philipp Wei{\ss}}, \bibinfo{person}{Enrico
  Costanza}, {and} \bibinfo{person}{Nadia Berthouze}.}
  \bibinfo{year}{2020}\natexlab{}.
\newblock \showarticletitle{{Evaluating saliency map explanations for
  convolutional neural networks}}. In \bibinfo{booktitle}{\emph{Proceedings of
  the 25th International Conference on Intelligent User Interfaces}}.
  \bibinfo{publisher}{ACM}, \bibinfo{address}{New York, NY, USA}.
\newblock
\showISBNx{9781450371186}
\urldef\tempurl%
\url{https://doi.org/10.1145/3377325.3377519}
\showDOI{\tempurl}


\bibitem[\protect\citeauthoryear{Alshaabi, Dewhurst, Minot, Arnold, Adams,
  Danforth, and Dodds}{Alshaabi et~al\mbox{.}}{2021}]%
        {alshaabi2021}
\bibfield{author}{\bibinfo{person}{Thayer Alshaabi},
  \bibinfo{person}{David~Rushing Dewhurst}, \bibinfo{person}{Joshua~R Minot},
  \bibinfo{person}{Michael~V Arnold}, \bibinfo{person}{Jane~L Adams},
  \bibinfo{person}{Christopher~M Danforth}, {and}
  \bibinfo{person}{Peter~Sheridan Dodds}.} \bibinfo{year}{2021}\natexlab{}.
\newblock \showarticletitle{{The growing amplification of social media:
  measuring temporal and social contagion dynamics for over 150 languages on
  Twitter for 2009–2020}}.
\newblock \bibinfo{journal}{\emph{EPJ Data Science}} \bibinfo{volume}{10},
  \bibinfo{number}{1} (\bibinfo{year}{2021}), \bibinfo{pages}{15}.
\newblock
\showISSN{2193-1127}
\urldef\tempurl%
\url{https://doi.org/10.1140/epjds/s13688-021-00271-0}
\showDOI{\tempurl}


\bibitem[\protect\citeauthoryear{Amershi, Chickering, Drucker, Lee, Simard, and
  Suh}{Amershi et~al\mbox{.}}{2015}]%
        {amershi2015}
\bibfield{author}{\bibinfo{person}{Saleema Amershi}, \bibinfo{person}{Max
  Chickering}, \bibinfo{person}{Steven~M. Drucker}, \bibinfo{person}{Bongshin
  Lee}, \bibinfo{person}{Patrice Simard}, {and} \bibinfo{person}{Jina Suh}.}
  \bibinfo{year}{2015}\natexlab{}.
\newblock \showarticletitle{{ModelTracker}}. In
  \bibinfo{booktitle}{\emph{Proceedings of the 33rd Annual ACM Conference on
  Human Factors in Computing Systems}}. \bibinfo{publisher}{ACM},
  \bibinfo{address}{New York, NY, USA}.
\newblock
\showISBNx{9781450331456}
\urldef\tempurl%
\url{https://doi.org/10.1145/2702123.2702509}
\showDOI{\tempurl}


\bibitem[\protect\citeauthoryear{Amini, Soleimany, Schwarting, Bhatia, and
  Rus}{Amini et~al\mbox{.}}{2019}]%
        {amini2019}
\bibfield{author}{\bibinfo{person}{Alexander Amini}, \bibinfo{person}{Ava~P
  Soleimany}, \bibinfo{person}{Wilko Schwarting}, \bibinfo{person}{Sangeeta~N
  Bhatia}, {and} \bibinfo{person}{Daniela Rus}.}
  \bibinfo{year}{2019}\natexlab{}.
\newblock \showarticletitle{{Uncovering and Mitigating Algorithmic Bias through
  Learned Latent Structure}}. In \bibinfo{booktitle}{\emph{Proceedings of the
  2019 AAAI/ACM Conference on AI, Ethics, and Society}}
  \emph{(\bibinfo{series}{AIES '19})}. \bibinfo{publisher}{Association for
  Computing Machinery}, \bibinfo{address}{New York, NY, USA},
  \bibinfo{pages}{289--295}.
\newblock
\showISBNx{9781450363242}
\urldef\tempurl%
\url{https://doi.org/10.1145/3306618.3314243}
\showDOI{\tempurl}


\bibitem[\protect\citeauthoryear{Amnesty~International}{Amnesty~International}{2018}]%
        {amnestyai2018}
\bibfield{author}{\bibinfo{person}{Access~Now Amnesty~International}.}
  \bibinfo{year}{2018}\natexlab{}.
\newblock \bibinfo{title}{{Toronto Declaration: Protecting the Right to
  Equality and Non-Discrimination in Machine Learning Systems}}.
\newblock
\newblock
\urldef\tempurl%
\url{https://www.accessnow.org/cms/assets/uploads/2018/08/The-Toronto-Declaration_ENG_08-2018.pdf}
\showURL{%
\tempurl}


\bibitem[\protect\citeauthoryear{Anik and Bunt}{Anik and Bunt}{2021}]%
        {anik2021}
\bibfield{author}{\bibinfo{person}{Ariful~Islam Anik} {and}
  \bibinfo{person}{Andrea Bunt}.} \bibinfo{year}{2021}\natexlab{}.
\newblock \showarticletitle{{Data-Centric Explanations: Explaining Training
  Data of Machine Learning Systems to Promote Transparency}}. In
  \bibinfo{booktitle}{\emph{Proceedings of the 2021 CHI Conference on Human
  Factors in Computing Systems}}. \bibinfo{publisher}{ACM},
  \bibinfo{address}{New York, NY, USA}.
\newblock
\showISBNx{9781450380966}
\urldef\tempurl%
\url{https://doi.org/10.1145/3411764.3445736}
\showDOI{\tempurl}


\bibitem[\protect\citeauthoryear{Arnstein}{Arnstein}{2019}]%
        {arnstein2019}
\bibfield{author}{\bibinfo{person}{Sherry~R Arnstein}.}
  \bibinfo{year}{2019}\natexlab{}.
\newblock \showarticletitle{{A Ladder of Citizen Participation}}.
\newblock \bibinfo{journal}{\emph{Journal of the American Planning
  Association}} \bibinfo{volume}{85}, \bibinfo{number}{1} (\bibinfo{date}{1}
  \bibinfo{year}{2019}), \bibinfo{pages}{24--34}.
\newblock
\showISSN{0194-4363}
\urldef\tempurl%
\url{https://doi.org/10.1080/01944363.2018.1559388}
\showDOI{\tempurl}


\bibitem[\protect\citeauthoryear{Assran, Romoff, Ballas, Pineau, and
  Rabbat}{Assran et~al\mbox{.}}{2019}]%
        {assran2020}
\bibfield{author}{\bibinfo{person}{Mahmoud Assran}, \bibinfo{person}{Joshua
  Romoff}, \bibinfo{person}{Nicolas Ballas}, \bibinfo{person}{Joelle Pineau},
  {and} \bibinfo{person}{Michael Rabbat}.} \bibinfo{year}{2019}\natexlab{}.
\newblock \showarticletitle{{Gossip-based Actor-Learner Architectures for Deep
  Reinforcement Learning}}.
\newblock  (\bibinfo{date}{6} \bibinfo{year}{2019}).
\newblock


\bibitem[\protect\citeauthoryear{Bakalar, Barreto, Bergman, Bogen, Chern,
  Corbett-Davies, Hall, Kloumann, Lam, Candela, Raghavan, Simons, Tannen, Tong,
  Vredenburgh, and Zhao}{Bakalar et~al\mbox{.}}{2021}]%
        {bakalar2021}
\bibfield{author}{\bibinfo{person}{Chloé Bakalar}, \bibinfo{person}{Renata
  Barreto}, \bibinfo{person}{Stevie Bergman}, \bibinfo{person}{Miranda Bogen},
  \bibinfo{person}{Bobbie Chern}, \bibinfo{person}{Sam Corbett-Davies},
  \bibinfo{person}{Melissa Hall}, \bibinfo{person}{Isabel Kloumann},
  \bibinfo{person}{Michelle Lam}, \bibinfo{person}{Joaquin~Quiñonero Candela},
  \bibinfo{person}{Manish Raghavan}, \bibinfo{person}{Joshua Simons},
  \bibinfo{person}{Jonathan Tannen}, \bibinfo{person}{Edmund Tong},
  \bibinfo{person}{Kate Vredenburgh}, {and} \bibinfo{person}{Jiejing Zhao}.}
  \bibinfo{year}{2021}\natexlab{}.
\newblock \showarticletitle{{Fairness On The Ground: Applying Algorithmic
  Fairness Approaches to Production Systems}}.
\newblock  (\bibinfo{date}{3} \bibinfo{year}{2021}).
\newblock


\bibitem[\protect\citeauthoryear{Balayn and G{\"{u}}rses}{Balayn and
  G{\"{u}}rses}{2021}]%
        {balayn2021Edri}
\bibfield{author}{\bibinfo{person}{Agathe Balayn} {and} \bibinfo{person}{Seda
  G{\"{u}}rses}.} \bibinfo{year}{2021}\natexlab{}.
\newblock \bibinfo{title}{{Beyond Debiasing: Regulating AI and its
  inequalities}}.
\newblock
\newblock
\urldef\tempurl%
\url{https://edri.org/our-work/if-ai-is-the-problem-is-debiasing-the-solution/}
\showURL{%
\tempurl}


\bibitem[\protect\citeauthoryear{Ball-Burack, Lee, Cobbe, and
  Singh}{Ball-Burack et~al\mbox{.}}{2021}]%
        {ballburack2021}
\bibfield{author}{\bibinfo{person}{Ari Ball-Burack}, \bibinfo{person}{Michelle
  Seng~Ah Lee}, \bibinfo{person}{Jennifer Cobbe}, {and}
  \bibinfo{person}{Jatinder Singh}.} \bibinfo{year}{2021}\natexlab{}.
\newblock \showarticletitle{{Differential Tweetment: Mitigating Racial Dialect
  Bias in Harmful Tweet Detection}}. In \bibinfo{booktitle}{\emph{Proceedings
  of the 2021 ACM Conference on Fairness, Accountability, and Transparency}}
  \emph{(\bibinfo{series}{FAccT '21})}. \bibinfo{publisher}{Association for
  Computing Machinery}, \bibinfo{address}{New York, NY, USA},
  \bibinfo{pages}{116--128}.
\newblock
\showISBNx{9781450383097}
\urldef\tempurl%
\url{https://doi.org/10.1145/3442188.3445875}
\showDOI{\tempurl}


\bibitem[\protect\citeauthoryear{Barredo~Arrieta, D{\'{i}}az-Rodr{\'{i}}guez,
  Del~Ser, Bennetot, Tabik, Barbado, Garcia, Gil-Lopez, Molina, Benjamins,
  Chatila, and Herrera}{Barredo~Arrieta et~al\mbox{.}}{2020}]%
        {barredoarrieta2020}
\bibfield{author}{\bibinfo{person}{Alejandro Barredo~Arrieta},
  \bibinfo{person}{Natalia D{\'{i}}az-Rodr{\'{i}}guez}, \bibinfo{person}{Javier
  Del~Ser}, \bibinfo{person}{Adrien Bennetot}, \bibinfo{person}{Siham Tabik},
  \bibinfo{person}{Alberto Barbado}, \bibinfo{person}{Salvador Garcia},
  \bibinfo{person}{Sergio Gil-Lopez}, \bibinfo{person}{Daniel Molina},
  \bibinfo{person}{Richard Benjamins}, \bibinfo{person}{Raja Chatila}, {and}
  \bibinfo{person}{Francisco Herrera}.} \bibinfo{year}{2020}\natexlab{}.
\newblock \showarticletitle{{Explainable Artificial Intelligence (XAI):
  Concepts, taxonomies, opportunities and challenges toward responsible AI}}.
\newblock \bibinfo{journal}{\emph{Information Fusion}}  \bibinfo{volume}{58}
  (\bibinfo{date}{6} \bibinfo{year}{2020}), \bibinfo{pages}{82--115}.
\newblock
\showISSN{1566-2535}
\urldef\tempurl%
\url{https://doi.org/10.1016/J.INFFUS.2019.12.012}
\showDOI{\tempurl}


\bibitem[\protect\citeauthoryear{Bellamy, Dey, Hind, Hoffman, Houde, Kannan,
  Lohia, Martino, Mehta, Mojsilovi{\'{c}}, Nagar, Ramamurthy, Richards, Saha,
  Sattigeri, Singh, Varshney, and Zhang}{Bellamy et~al\mbox{.}}{2019}]%
        {bellamy2018}
\bibfield{author}{\bibinfo{person}{R~K~E Bellamy}, \bibinfo{person}{K Dey},
  \bibinfo{person}{M Hind}, \bibinfo{person}{S~C Hoffman}, \bibinfo{person}{S
  Houde}, \bibinfo{person}{K Kannan}, \bibinfo{person}{P Lohia},
  \bibinfo{person}{J Martino}, \bibinfo{person}{S Mehta}, \bibinfo{person}{A
  Mojsilovi{\'{c}}}, \bibinfo{person}{S Nagar}, \bibinfo{person}{K~Natesan
  Ramamurthy}, \bibinfo{person}{J Richards}, \bibinfo{person}{D Saha},
  \bibinfo{person}{P Sattigeri}, \bibinfo{person}{M Singh},
  \bibinfo{person}{K~R Varshney}, {and} \bibinfo{person}{Y Zhang}.}
  \bibinfo{year}{2019}\natexlab{}.
\newblock \showarticletitle{{AI Fairness 360: An extensible toolkit for
  detecting and mitigating algorithmic bias}}.
\newblock \bibinfo{journal}{\emph{IBM Journal of Research and Development}}
  \bibinfo{volume}{63}, \bibinfo{number}{4/5} (\bibinfo{year}{2019}),
  \bibinfo{pages}{1--4}.
\newblock
\urldef\tempurl%
\url{https://doi.org/10.1147/JRD.2019.2942287}
\showDOI{\tempurl}


\bibitem[\protect\citeauthoryear{Bender and Friedman}{Bender and
  Friedman}{2018}]%
        {bender2018}
\bibfield{author}{\bibinfo{person}{Emily~M. Bender} {and}
  \bibinfo{person}{Batya Friedman}.} \bibinfo{year}{2018}\natexlab{}.
\newblock \showarticletitle{{Data Statements for Natural Language Processing:
  Toward Mitigating System Bias and Enabling Better Science}}.
\newblock \bibinfo{journal}{\emph{Transactions of the Association for
  Computational Linguistics}}  \bibinfo{volume}{6} (\bibinfo{date}{12}
  \bibinfo{year}{2018}).
\newblock
\showISSN{2307-387X}
\urldef\tempurl%
\url{https://doi.org/10.1162/tacl_a_00041}
\showDOI{\tempurl}


\bibitem[\protect\citeauthoryear{Bender, Gebru, McMillan-Major, and
  Shmitchell}{Bender et~al\mbox{.}}{2021}]%
        {bender2021}
\bibfield{author}{\bibinfo{person}{Emily~M Bender}, \bibinfo{person}{Timnit
  Gebru}, \bibinfo{person}{Angelina McMillan-Major}, {and}
  \bibinfo{person}{Shmargaret Shmitchell}.} \bibinfo{year}{2021}\natexlab{}.
\newblock \showarticletitle{{On the Dangers of Stochastic Parrots: Can Language
  Models Be Too Big?}}. In \bibinfo{booktitle}{\emph{Proceedings of the 2021
  ACM Conference on Fairness, Accountability, and Transparency}}
  \emph{(\bibinfo{series}{FAccT '21})}. \bibinfo{publisher}{Association for
  Computing Machinery}, \bibinfo{address}{New York, NY, USA},
  \bibinfo{pages}{610--623}.
\newblock
\showISBNx{9781450383097}
\urldef\tempurl%
\url{https://doi.org/10.1145/3442188.3445922}
\showDOI{\tempurl}


\bibitem[\protect\citeauthoryear{Berk, Heidari, Jabbari, Kearns, and Roth}{Berk
  et~al\mbox{.}}{2017}]%
        {berk2017}
\bibfield{author}{\bibinfo{person}{Richard Berk}, \bibinfo{person}{Hoda
  Heidari}, \bibinfo{person}{Shahin Jabbari}, \bibinfo{person}{Michael Kearns},
  {and} \bibinfo{person}{Aaron Roth}.} \bibinfo{year}{2017}\natexlab{}.
\newblock \showarticletitle{{Fairness in Criminal Justice Risk Assessments: The
  State of the Art}}.
\newblock  (\bibinfo{date}{3} \bibinfo{year}{2017}).
\newblock


\bibitem[\protect\citeauthoryear{Biggio and Roli}{Biggio and Roli}{2018}]%
        {biggio2018}
\bibfield{author}{\bibinfo{person}{Battista Biggio} {and}
  \bibinfo{person}{Fabio Roli}.} \bibinfo{year}{2018}\natexlab{}.
\newblock \showarticletitle{{Wild patterns: Ten years after the rise of
  adversarial machine learning}}.
\newblock \bibinfo{journal}{\emph{Pattern Recognition}}  \bibinfo{volume}{84}
  (\bibinfo{date}{12} \bibinfo{year}{2018}), \bibinfo{pages}{317--331}.
\newblock
\showISSN{00313203}
\urldef\tempurl%
\url{https://doi.org/10.1016/j.patcog.2018.07.023}
\showDOI{\tempurl}


\bibitem[\protect\citeauthoryear{Binns, Van~Kleek, Veale, Lyngs, Zhao, and
  Shadbolt}{Binns et~al\mbox{.}}{2018}]%
        {binns2018}
\bibfield{author}{\bibinfo{person}{Reuben Binns}, \bibinfo{person}{Max
  Van~Kleek}, \bibinfo{person}{Michael Veale}, \bibinfo{person}{Ulrik Lyngs},
  \bibinfo{person}{Jun Zhao}, {and} \bibinfo{person}{Nigel Shadbolt}.}
  \bibinfo{year}{2018}\natexlab{}.
\newblock \showarticletitle{{'It's Reducing a Human Being to a Percentage';
  Perceptions of Justice in Algorithmic Decisions}}.
\newblock  (\bibinfo{date}{1} \bibinfo{year}{2018}).
\newblock
\urldef\tempurl%
\url{https://doi.org/10.1145/3173574.3173951}
\showDOI{\tempurl}


\bibitem[\protect\citeauthoryear{Bird, Dud{\'{i}}k, Edgar, Horn, Lutz, Milan,
  Sameki, Wallach, and Walker}{Bird et~al\mbox{.}}{2020}]%
        {bird2020}
\bibfield{author}{\bibinfo{person}{Sarah Bird}, \bibinfo{person}{Miro
  Dud{\'{i}}k}, \bibinfo{person}{Richard Edgar}, \bibinfo{person}{Brandon
  Horn}, \bibinfo{person}{Roman Lutz}, \bibinfo{person}{Vanessa Milan},
  \bibinfo{person}{Mehrnoosh Sameki}, \bibinfo{person}{Hanna Wallach}, {and}
  \bibinfo{person}{Kathleen Walker}.} \bibinfo{year}{2020}\natexlab{}.
\newblock \bibinfo{booktitle}{\emph{{Fairlearn: A toolkit for assessing and
  improving fairness in AI}}}.
\newblock \bibinfo{type}{{T}echnical {R}eport} MSR-TR-2020-32.
  \bibinfo{institution}{Microsoft}.
\newblock
\urldef\tempurl%
\url{https://www.microsoft.com/en-us/research/publication/fairlearn-a-toolkit-for-assessing-and-improving-fairness-in-ai/}
\showURL{%
\tempurl}


\bibitem[\protect\citeauthoryear{Birhane, Kalluri, Card, Agnew, Dotan, and
  Bao}{Birhane et~al\mbox{.}}{2021}]%
        {bihrane2021}
\bibfield{author}{\bibinfo{person}{Abeba Birhane}, \bibinfo{person}{Pratyusha
  Kalluri}, \bibinfo{person}{Dallas Card}, \bibinfo{person}{William Agnew},
  \bibinfo{person}{Ravit Dotan}, {and} \bibinfo{person}{Michelle Bao}.}
  \bibinfo{year}{2021}\natexlab{}.
\newblock \showarticletitle{{The Values Encoded in Machine Learning Research}}.
\newblock  (\bibinfo{date}{6} \bibinfo{year}{2021}).
\newblock


\bibitem[\protect\citeauthoryear{Blazevic, Mugalula, and Wandera}{Blazevic
  et~al\mbox{.}}{2021}]%
        {blazevic2021}
\bibfield{author}{\bibinfo{person}{Alice~Namuli Blazevic},
  \bibinfo{person}{Patrick Mugalula}, {and} \bibinfo{person}{Andrew Wandera}.}
  \bibinfo{year}{2021}\natexlab{}.
\newblock \showarticletitle{{Towards Operationalizing the Data Protection and
  Privacy Act 2020: Understanding the Draft Data Protection and Privacy
  Regulations, 2020}}.
\newblock \bibinfo{journal}{\emph{SSRN Electronic Journal}}
  (\bibinfo{year}{2021}).
\newblock
\showISSN{1556-5068}
\urldef\tempurl%
\url{https://doi.org/10.2139/ssrn.3776353}
\showDOI{\tempurl}


\bibitem[\protect\citeauthoryear{Blodgett, Barocas, Daum{\'{e}}, and
  Wallach}{Blodgett et~al\mbox{.}}{2020}]%
        {blodgett2020}
\bibfield{author}{\bibinfo{person}{Su~Lin Blodgett}, \bibinfo{person}{Solon
  Barocas}, \bibinfo{person}{Hal Daum{\'{e}}}, {and} \bibinfo{person}{Hanna
  Wallach}.} \bibinfo{year}{2020}\natexlab{}.
\newblock \showarticletitle{{Language (Technology) is Power: A Critical Survey
  of "Bias" in NLP}}.
\newblock  (\bibinfo{date}{5} \bibinfo{year}{2020}).
\newblock


\bibitem[\protect\citeauthoryear{Bountouridis, Harambam, Makhortykh, Marrero,
  Tintarev, and Hauff}{Bountouridis et~al\mbox{.}}{2019}]%
        {bountouridis2019}
\bibfield{author}{\bibinfo{person}{Dimitrios Bountouridis},
  \bibinfo{person}{Jaron Harambam}, \bibinfo{person}{Mykola Makhortykh},
  \bibinfo{person}{Mónica Marrero}, \bibinfo{person}{Nava Tintarev}, {and}
  \bibinfo{person}{Claudia Hauff}.} \bibinfo{year}{2019}\natexlab{}.
\newblock \showarticletitle{{SIREN: A Simulation Framework for Understanding
  the Effects of Recommender Systems in Online News Environments}}. In
  \bibinfo{booktitle}{\emph{Proceedings of the Conference on Fairness,
  Accountability, and Transparency}} \emph{(\bibinfo{series}{FAT* '19})}.
  \bibinfo{publisher}{Association for Computing Machinery},
  \bibinfo{address}{New York, NY, USA}, \bibinfo{pages}{150--159}.
\newblock
\showISBNx{9781450361255}
\urldef\tempurl%
\url{https://doi.org/10.1145/3287560.3287583}
\showDOI{\tempurl}


\bibitem[\protect\citeauthoryear{Bryson, Diamantis, and Grant}{Bryson
  et~al\mbox{.}}{2017}]%
        {bryson2017}
\bibfield{author}{\bibinfo{person}{Joanna~J. Bryson},
  \bibinfo{person}{Mihailis~E. Diamantis}, {and} \bibinfo{person}{Thomas~D.
  Grant}.} \bibinfo{year}{2017}\natexlab{}.
\newblock \showarticletitle{{Of, for, and by the people: the legal lacuna of
  synthetic persons}}.
\newblock \bibinfo{journal}{\emph{Artificial Intelligence and Law}}
  \bibinfo{volume}{25}, \bibinfo{number}{3} (\bibinfo{date}{9}
  \bibinfo{year}{2017}), \bibinfo{pages}{273--291}.
\newblock
\showISSN{0924-8463}
\urldef\tempurl%
\url{https://doi.org/10.1007/s10506-017-9214-9}
\showDOI{\tempurl}


\bibitem[\protect\citeauthoryear{Buolamwini and Gebru}{Buolamwini and
  Gebru}{2018}]%
        {buolamwini2018}
\bibfield{author}{\bibinfo{person}{Joy Buolamwini} {and}
  \bibinfo{person}{Timnit Gebru}.} \bibinfo{year}{2018}\natexlab{}.
\newblock \showarticletitle{{Gender Shades: Intersectional Accuracy Disparities
  in Commercial Gender Classification}}. In
  \bibinfo{booktitle}{\emph{Proceedings of the 1st Conference on Fairness,
  Accountability and Transparency}} \emph{(\bibinfo{series}{Proceedings of
  Machine Learning Research}, Vol.~\bibinfo{volume}{81})},
  \bibfield{editor}{\bibinfo{person}{Sorelle~A Friedler} {and}
  \bibinfo{person}{Christo Wilson}} (Eds.). \bibinfo{publisher}{PMLR},
  \bibinfo{pages}{77--91}.
\newblock
\urldef\tempurl%
\url{https://proceedings.mlr.press/v81/buolamwini18a.html}
\showURL{%
\tempurl}


\bibitem[\protect\citeauthoryear{Cai, Jongejan, and Holbrook}{Cai
  et~al\mbox{.}}{2019}]%
        {cai2019}
\bibfield{author}{\bibinfo{person}{Carrie~J. Cai}, \bibinfo{person}{Jonas
  Jongejan}, {and} \bibinfo{person}{Jess Holbrook}.}
  \bibinfo{year}{2019}\natexlab{}.
\newblock \showarticletitle{{The effects of example-based explanations in a
  machine learning interface}}. In \bibinfo{booktitle}{\emph{Proceedings of the
  24th International Conference on Intelligent User Interfaces}}.
  \bibinfo{publisher}{ACM}, \bibinfo{address}{New York, NY, USA}.
\newblock
\showISBNx{9781450362726}
\urldef\tempurl%
\url{https://doi.org/10.1145/3301275.3302289}
\showDOI{\tempurl}


\bibitem[\protect\citeauthoryear{Calvert, Heikoop, Mecacci, and
  Van~Arem}{Calvert et~al\mbox{.}}{2019}]%
        {calvert2019}
\bibfield{author}{\bibinfo{person}{Simeon~C Calvert},
  \bibinfo{person}{Daniël~D Heikoop}, \bibinfo{person}{Giulio Mecacci}, {and}
  \bibinfo{person}{Bart Van~Arem}.} \bibinfo{year}{2019}\natexlab{}.
\newblock \showarticletitle{{A human centric framework for the analysis of
  automated driving systems based on meaningful human control}}.
\newblock \bibinfo{journal}{\emph{Theoretical Issues in Ergonomics Science}}
  \bibinfo{volume}{21}, \bibinfo{number}{4} (\bibinfo{year}{2019}),
  \bibinfo{pages}{478--506}.
\newblock
\urldef\tempurl%
\url{https://doi.org/10.1080/1463922X.2019.1697390}
\showDOI{\tempurl}


\bibitem[\protect\citeauthoryear{Chasalow and Levy}{Chasalow and Levy}{2021}]%
        {chasalow2021}
\bibfield{author}{\bibinfo{person}{Kyla Chasalow} {and} \bibinfo{person}{Karen
  Levy}.} \bibinfo{year}{2021}\natexlab{}.
\newblock \showarticletitle{{Representativeness in Statistics, Politics, and
  Machine Learning}}. In \bibinfo{booktitle}{\emph{Proceedings of the 2021 ACM
  Conference on Fairness, Accountability, and Transparency}}
  \emph{(\bibinfo{series}{FAccT '21})}. \bibinfo{publisher}{Association for
  Computing Machinery}, \bibinfo{address}{New York, NY, USA},
  \bibinfo{pages}{77--89}.
\newblock
\showISBNx{9781450383097}
\urldef\tempurl%
\url{https://doi.org/10.1145/3442188.3445872}
\showDOI{\tempurl}


\bibitem[\protect\citeauthoryear{Cheng, Wang, Zhang, O'Connell, Gray, Harper,
  and Zhu}{Cheng et~al\mbox{.}}{2019}]%
        {cheng2019}
\bibfield{author}{\bibinfo{person}{Hao-Fei Cheng}, \bibinfo{person}{Ruotong
  Wang}, \bibinfo{person}{Zheng Zhang}, \bibinfo{person}{Fiona O'Connell},
  \bibinfo{person}{Terrance Gray}, \bibinfo{person}{F~Maxwell Harper}, {and}
  \bibinfo{person}{Haiyi Zhu}.} \bibinfo{year}{2019}\natexlab{}.
\newblock \showarticletitle{{Explaining Decision-Making Algorithms through UI:
  Strategies to Help Non-Expert Stakeholders}}. In
  \bibinfo{booktitle}{\emph{Proceedings of the 2019 CHI Conference on Human
  Factors in Computing Systems}} \emph{(\bibinfo{series}{CHI '19})}.
  \bibinfo{publisher}{Association for Computing Machinery},
  \bibinfo{address}{New York, NY, USA}, \bibinfo{pages}{1--12}.
\newblock
\showISBNx{9781450359702}
\urldef\tempurl%
\url{https://doi.org/10.1145/3290605.3300789}
\showDOI{\tempurl}


\bibitem[\protect\citeauthoryear{{China Electronics Standardization
  Institute}}{{China Electronics Standardization Institute}}{2018}]%
        {aichina2018}
\bibfield{author}{\bibinfo{person}{{China Electronics Standardization
  Institute}}.} \bibinfo{year}{2018}\natexlab{}.
\newblock \bibinfo{title}{{Original CSET Translation of "Artificial
  Intelligence Standardization White Paper"}}.
\newblock
\newblock
\urldef\tempurl%
\url{https://cset.georgetown.edu/research/artificial-intelligence-standardization-white-paper/}
\showURL{%
\tempurl}


\bibitem[\protect\citeauthoryear{Chouldechova}{Chouldechova}{2016}]%
        {chouldechova2016}
\bibfield{author}{\bibinfo{person}{Alexandra Chouldechova}.}
  \bibinfo{year}{2016}\natexlab{}.
\newblock \showarticletitle{{Fair prediction with disparate impact: A study of
  bias in recidivism prediction instruments}}.
\newblock  (\bibinfo{date}{10} \bibinfo{year}{2016}).
\newblock


\bibitem[\protect\citeauthoryear{Cila, Ferri, de~Waal, Gloerich, and
  Karpinski}{Cila et~al\mbox{.}}{2020}]%
        {cila2020}
\bibfield{author}{\bibinfo{person}{Nazli Cila}, \bibinfo{person}{Gabriele
  Ferri}, \bibinfo{person}{Martijn de Waal}, \bibinfo{person}{Inte Gloerich},
  {and} \bibinfo{person}{Tara Karpinski}.} \bibinfo{year}{2020}\natexlab{}.
\newblock \showarticletitle{{The Blockchain and the Commons: Dilemmas in the
  Design of Local Platforms}}. In \bibinfo{booktitle}{\emph{Proceedings of the
  2020 CHI Conference on Human Factors in Computing Systems}}.
  \bibinfo{publisher}{Association for Computing Machinery},
  \bibinfo{address}{New York, NY, USA}, \bibinfo{pages}{1--14}.
\newblock
\showISBNx{9781450367080}
\urldef\tempurl%
\url{https://doi-org.tudelft.idm.oclc.org/10.1145/3313831.3376660}
\showURL{%
\tempurl}


\bibitem[\protect\citeauthoryear{Crawford and Paglen}{Crawford and
  Paglen}{2019}]%
        {crawford2019}
\bibfield{author}{\bibinfo{person}{Kate Crawford} {and} \bibinfo{person}{Trevor
  Paglen}.} \bibinfo{year}{2019}\natexlab{}.
\newblock \bibinfo{title}{{Excavating AI: The Politics of Training Sets for
  Machine Learning}}.
\newblock
\newblock


\bibitem[\protect\citeauthoryear{Cretu, Stavrou, Locasto, Stolfo, and
  Keromytis}{Cretu et~al\mbox{.}}{2008}]%
        {cretu2008}
\bibfield{author}{\bibinfo{person}{Gabriela~F. Cretu}, \bibinfo{person}{Angelos
  Stavrou}, \bibinfo{person}{Michael~E. Locasto}, \bibinfo{person}{Salvatore~J.
  Stolfo}, {and} \bibinfo{person}{Angelos~D. Keromytis}.}
  \bibinfo{year}{2008}\natexlab{}.
\newblock \showarticletitle{{Casting out Demons: Sanitizing Training Data for
  Anomaly Sensors}}. In \bibinfo{booktitle}{\emph{2008 IEEE Symposium on
  Security and Privacy (sp 2008)}}. \bibinfo{publisher}{IEEE},
  \bibinfo{pages}{81--95}.
\newblock
\showISBNx{978-0-7695-3168-7}
\showISSN{1081-6011}
\urldef\tempurl%
\url{https://doi.org/10.1109/SP.2008.11}
\showDOI{\tempurl}


\bibitem[\protect\citeauthoryear{Dalton, Frosio, and Garland}{Dalton
  et~al\mbox{.}}{2019}]%
        {dalton2020}
\bibfield{author}{\bibinfo{person}{Steven Dalton}, \bibinfo{person}{Iuri
  Frosio}, {and} \bibinfo{person}{Michael Garland}.}
  \bibinfo{year}{2019}\natexlab{}.
\newblock \showarticletitle{{Accelerating Reinforcement Learning through GPU
  Atari Emulation}}.
\newblock  (\bibinfo{date}{7} \bibinfo{year}{2019}).
\newblock


\bibitem[\protect\citeauthoryear{D'Amour, Srinivasan, Atwood, Baljekar,
  Sculley, and Halpern}{D'Amour et~al\mbox{.}}{2020}]%
        {damour2020}
\bibfield{author}{\bibinfo{person}{Alexander D'Amour}, \bibinfo{person}{Hansa
  Srinivasan}, \bibinfo{person}{James Atwood}, \bibinfo{person}{Pallavi
  Baljekar}, \bibinfo{person}{D Sculley}, {and} \bibinfo{person}{Yoni
  Halpern}.} \bibinfo{year}{2020}\natexlab{}.
\newblock \showarticletitle{{Fairness is Not Static: Deeper Understanding of
  Long Term Fairness via Simulation Studies}}. In
  \bibinfo{booktitle}{\emph{Proceedings of the 2020 Conference on Fairness,
  Accountability, and Transparency}} \emph{(\bibinfo{series}{FAT* '20})}.
  \bibinfo{publisher}{Association for Computing Machinery},
  \bibinfo{address}{New York, NY, USA}, \bibinfo{pages}{525--534}.
\newblock
\showISBNx{9781450369367}
\urldef\tempurl%
\url{https://doi.org/10.1145/3351095.3372878}
\showDOI{\tempurl}


\bibitem[\protect\citeauthoryear{{Dasha Simons}}{{Dasha Simons}}{2019}]%
        {simons2019}
\bibfield{author}{\bibinfo{person}{{Dasha Simons}}.}
  \bibinfo{year}{2019}\natexlab{}.
\newblock \bibinfo{booktitle}{\emph{{Design for fairness in AI: Cooking a fair
  AI Dish}}}.
\newblock \bibinfo{type}{{T}echnical {R}eport}. \bibinfo{institution}{Delft
  University of Technology. Graduation project. MSc in Strategic Product
  Design.}
\newblock
\urldef\tempurl%
\url{http://resolver.tudelft.nl/uuid:5a116c17-ce0a-4236-b283-da6b8545628c}
\showURL{%
\tempurl}


\bibitem[\protect\citeauthoryear{Davidson, Bhattacharya, and Weber}{Davidson
  et~al\mbox{.}}{2019}]%
        {davidson2019}
\bibfield{author}{\bibinfo{person}{Thomas Davidson}, \bibinfo{person}{Debasmita
  Bhattacharya}, {and} \bibinfo{person}{Ingmar Weber}.}
  \bibinfo{year}{2019}\natexlab{}.
\newblock \showarticletitle{{Racial Bias in Hate Speech and Abusive Language
  Detection Datasets}}. In \bibinfo{booktitle}{\emph{Proceedings of the Third
  Workshop on Abusive Language Online}}. \bibinfo{publisher}{Association for
  Computational Linguistics}, \bibinfo{address}{Stroudsburg, PA, USA}.
\newblock
\urldef\tempurl%
\url{https://doi.org/10.18653/v1/W19-3504}
\showDOI{\tempurl}


\bibitem[\protect\citeauthoryear{Davis and Nathan}{Davis and Nathan}{2015}]%
        {davis2015}
\bibfield{author}{\bibinfo{person}{Janet Davis} {and} \bibinfo{person}{Lisa~P.
  Nathan}.} \bibinfo{year}{2015}\natexlab{}.
\newblock \showarticletitle{{Value Sensitive Design: Applications, Adaptations,
  and Critiques}}.
\newblock In \bibinfo{booktitle}{\emph{Handbook of Ethics, Values, and
  Technological Design}}. \bibinfo{publisher}{Springer Netherlands},
  \bibinfo{address}{Dordrecht}, \bibinfo{pages}{11--40}.
\newblock
\urldef\tempurl%
\url{https://doi.org/10.1007/978-94-007-6970-0_3}
\showDOI{\tempurl}


\bibitem[\protect\citeauthoryear{Denton, Hanna, Amironesei, Smart, Nicole, and
  Scheuerman}{Denton et~al\mbox{.}}{2020}]%
        {denton2020}
\bibfield{author}{\bibinfo{person}{Emily Denton}, \bibinfo{person}{Alex Hanna},
  \bibinfo{person}{Razvan Amironesei}, \bibinfo{person}{Andrew Smart},
  \bibinfo{person}{Hilary Nicole}, {and} \bibinfo{person}{Morgan~Klaus
  Scheuerman}.} \bibinfo{year}{2020}\natexlab{}.
\newblock \showarticletitle{{Bringing the People Back In: Contesting Benchmark
  Machine Learning Datasets}}.
\newblock  (\bibinfo{date}{7} \bibinfo{year}{2020}).
\newblock
\urldef\tempurl%
\url{https://arxiv.org/abs/2007.07399}
\showURL{%
\tempurl}


\bibitem[\protect\citeauthoryear{Dhurandhar, Chen, Luss, Tu, Ting, Shanmugam,
  and Das}{Dhurandhar et~al\mbox{.}}{2018}]%
        {dhurandhar2018}
\bibfield{author}{\bibinfo{person}{Amit Dhurandhar}, \bibinfo{person}{Pin-Yu
  Chen}, \bibinfo{person}{Ronny Luss}, \bibinfo{person}{Chun-Chen Tu},
  \bibinfo{person}{Paishun Ting}, \bibinfo{person}{Karthikeyan Shanmugam},
  {and} \bibinfo{person}{Payel Das}.} \bibinfo{year}{2018}\natexlab{}.
\newblock \showarticletitle{{Explanations based on the Missing: Towards
  Contrastive Explanations with Pertinent Negatives}}.
\newblock  (\bibinfo{date}{2} \bibinfo{year}{2018}).
\newblock


\bibitem[\protect\citeauthoryear{Dixon, Li, Sorensen, Thain, and
  Vasserman}{Dixon et~al\mbox{.}}{2018}]%
        {dixon2018}
\bibfield{author}{\bibinfo{person}{Lucas Dixon}, \bibinfo{person}{John Li},
  \bibinfo{person}{Jeffrey Sorensen}, \bibinfo{person}{Nithum Thain}, {and}
  \bibinfo{person}{Lucy Vasserman}.} \bibinfo{year}{2018}\natexlab{}.
\newblock \showarticletitle{{Measuring and Mitigating Unintended Bias in Text
  Classification}}. In \bibinfo{booktitle}{\emph{Proceedings of the 2018
  AAAI/ACM Conference on AI, Ethics, and Society}}. \bibinfo{publisher}{ACM},
  \bibinfo{address}{New York, NY, USA}, \bibinfo{pages}{67--73}.
\newblock
\showISBNx{9781450360128}
\urldef\tempurl%
\url{https://doi.org/10.1145/3278721.3278729}
\showDOI{\tempurl}


\bibitem[\protect\citeauthoryear{Dodge, Liao, Zhang, Bellamy, and Dugan}{Dodge
  et~al\mbox{.}}{2019}]%
        {dodge2019}
\bibfield{author}{\bibinfo{person}{Jonathan Dodge}, \bibinfo{person}{Q.~Vera
  Liao}, \bibinfo{person}{Yunfeng Zhang}, \bibinfo{person}{Rachel K.~E.
  Bellamy}, {and} \bibinfo{person}{Casey Dugan}.}
  \bibinfo{year}{2019}\natexlab{}.
\newblock \showarticletitle{{Explaining Models: An Empirical Study of How
  Explanations Impact Fairness Judgment}}.
\newblock  (\bibinfo{date}{1} \bibinfo{year}{2019}).
\newblock
\urldef\tempurl%
\url{https://doi.org/10.1145/3301275.3302310}
\showDOI{\tempurl}


\bibitem[\protect\citeauthoryear{Dotan and Milli}{Dotan and Milli}{2019}]%
        {dotan2019}
\bibfield{author}{\bibinfo{person}{Ravit Dotan} {and} \bibinfo{person}{Smitha
  Milli}.} \bibinfo{year}{2019}\natexlab{}.
\newblock \showarticletitle{{Value-laden Disciplinary Shifts in Machine
  Learning}}.
\newblock  (\bibinfo{date}{12} \bibinfo{year}{2019}).
\newblock


\bibitem[\protect\citeauthoryear{Draws, Rieger, Inel, Gadiraju, and
  Tintarev}{Draws et~al\mbox{.}}{2021}]%
        {draws2021_5}
\bibfield{author}{\bibinfo{person}{Tim Draws}, \bibinfo{person}{Alisa Rieger},
  \bibinfo{person}{Oana Inel}, \bibinfo{person}{Ujwal Gadiraju}, {and}
  \bibinfo{person}{Nava Tintarev}.} \bibinfo{year}{2021}\natexlab{}.
\newblock \showarticletitle{{A Checklist to Combat Cognitive Biases in
  Crowdsourcing}}.
\newblock \bibinfo{journal}{\emph{Proceedings of the AAAI Conference on Human
  Computation and Crowdsourcing}} \bibinfo{volume}{9}, \bibinfo{number}{1}
  (\bibinfo{date}{10} \bibinfo{year}{2021}), \bibinfo{pages}{48--59}.
\newblock
\urldef\tempurl%
\url{https://ojs.aaai.org/index.php/HCOMP/article/view/18939}
\showURL{%
\tempurl}


\bibitem[\protect\citeauthoryear{Dwork, Hardt, Pitassi, Reingold, and
  Zemel}{Dwork et~al\mbox{.}}{2011}]%
        {dwork2011}
\bibfield{author}{\bibinfo{person}{Cynthia Dwork}, \bibinfo{person}{Moritz
  Hardt}, \bibinfo{person}{Toniann Pitassi}, \bibinfo{person}{Omer Reingold},
  {and} \bibinfo{person}{Rich Zemel}.} \bibinfo{year}{2011}\natexlab{}.
\newblock \showarticletitle{{Fairness Through Awareness}}.
\newblock  (\bibinfo{date}{4} \bibinfo{year}{2011}).
\newblock


\bibitem[\protect\citeauthoryear{Dwork, McSherry, Nissim, and Smith}{Dwork
  et~al\mbox{.}}{2006}]%
        {dwork2006}
\bibfield{author}{\bibinfo{person}{Cynthia Dwork}, \bibinfo{person}{Frank
  McSherry}, \bibinfo{person}{Kobbi Nissim}, {and} \bibinfo{person}{Adam
  Smith}.} \bibinfo{year}{2006}\natexlab{}.
\newblock \showarticletitle{{Calibrating Noise to Sensitivity in Private Data
  Analysis}}.
\newblock \bibinfo{pages}{265--284}.
\newblock
\urldef\tempurl%
\url{https://doi.org/10.1007/11681878_14}
\showDOI{\tempurl}


\bibitem[\protect\citeauthoryear{Earp and Staddon}{Earp and Staddon}{2016}]%
        {earp2016}
\bibfield{author}{\bibinfo{person}{Julia Earp} {and} \bibinfo{person}{Jessica
  Staddon}.} \bibinfo{year}{2016}\natexlab{}.
\newblock \showarticletitle{{"I had no idea this was a thing"}}. In
  \bibinfo{booktitle}{\emph{Proceedings of the 6th Workshop on Socio-Technical
  Aspects in Security and Trust}}. \bibinfo{publisher}{ACM},
  \bibinfo{address}{New York, NY, USA}, \bibinfo{pages}{79--86}.
\newblock
\showISBNx{9781450348263}
\urldef\tempurl%
\url{https://doi.org/10.1145/3046055.3046062}
\showDOI{\tempurl}


\bibitem[\protect\citeauthoryear{Edizel, Bonchi, Hajian, Panisson, and
  Tassa}{Edizel et~al\mbox{.}}{2020}]%
        {edizel2020}
\bibfield{author}{\bibinfo{person}{Bora Edizel}, \bibinfo{person}{Francesco
  Bonchi}, \bibinfo{person}{Sara Hajian}, \bibinfo{person}{André Panisson},
  {and} \bibinfo{person}{Tamir Tassa}.} \bibinfo{year}{2020}\natexlab{}.
\newblock \showarticletitle{{FaiRecSys: mitigating algorithmic bias in
  recommender systems}}.
\newblock \bibinfo{journal}{\emph{International Journal of Data Science and
  Analytics}} \bibinfo{volume}{9}, \bibinfo{number}{2} (\bibinfo{year}{2020}),
  \bibinfo{pages}{197--213}.
\newblock
\showISSN{2364-4168}
\urldef\tempurl%
\url{https://doi.org/10.1007/s41060-019-00181-5}
\showDOI{\tempurl}


\bibitem[\protect\citeauthoryear{{European Commission}}{{European
  Commission}}{2018}]%
        {GDPR2018}
\bibfield{author}{\bibinfo{person}{{European Commission}}.}
  \bibinfo{year}{2018}\natexlab{}.
\newblock \bibinfo{title}{{2018 reform of EU data protection rules}}.
\newblock
\newblock
\urldef\tempurl%
\url{https://ec.europa.eu/commission/sites/beta-political/files/data-protection-factsheet-changes_en.pdf}
\showURL{%
\tempurl}


\bibitem[\protect\citeauthoryear{{European Commission}}{{European
  Commission}}{2019}]%
        {europeancommissionEthicsAI2019}
\bibfield{author}{\bibinfo{person}{{European Commission}}.}
  \bibinfo{year}{2019}\natexlab{}.
\newblock \bibinfo{title}{{Ethics guidelines for trustworthy AI}}.
\newblock
\newblock
\urldef\tempurl%
\url{https://www.aepd.es/sites/default/files/2019-12/ai-ethics-guidelines.pdf}
\showURL{%
\tempurl}


\bibitem[\protect\citeauthoryear{Farke, Balash, Golla, D{\"{u}}rmuth, and
  Aviv}{Farke et~al\mbox{.}}{2021}]%
        {farke2021}
\bibfield{author}{\bibinfo{person}{Florian~M. Farke}, \bibinfo{person}{David~G.
  Balash}, \bibinfo{person}{Maximilian Golla}, \bibinfo{person}{Markus
  D{\"{u}}rmuth}, {and} \bibinfo{person}{Adam~J. Aviv}.}
  \bibinfo{year}{2021}\natexlab{}.
\newblock \showarticletitle{{Are Privacy Dashboards Good for End Users?
  Evaluating User Perceptions and Reactions to Google's My Activity (Extended
  Version)}}.
\newblock  (\bibinfo{date}{5} \bibinfo{year}{2021}).
\newblock


\bibitem[\protect\citeauthoryear{Fischer-H{\"{u}}bner, Angulo, Karegar, and
  Pulls}{Fischer-H{\"{u}}bner et~al\mbox{.}}{2016}]%
        {fischerhubner2016}
\bibfield{author}{\bibinfo{person}{Simone Fischer-H{\"{u}}bner},
  \bibinfo{person}{Julio Angulo}, \bibinfo{person}{Farzaneh Karegar}, {and}
  \bibinfo{person}{Tobias Pulls}.} \bibinfo{year}{2016}\natexlab{}.
\newblock \showarticletitle{{Transparency, Privacy and Trust – Technology for
  Tracking and Controlling My Data Disclosures: Does This Work?}}
\newblock \bibinfo{pages}{3--14}.
\newblock
\urldef\tempurl%
\url{https://doi.org/10.1007/978-3-319-41354-9_1}
\showDOI{\tempurl}


\bibitem[\protect\citeauthoryear{Fjeld, Achten, Hilligoss, Nagy, and
  Srikumar}{Fjeld et~al\mbox{.}}{2020}]%
        {fjeld2020}
\bibfield{author}{\bibinfo{person}{Jessica Fjeld}, \bibinfo{person}{Nele
  Achten}, \bibinfo{person}{Hannah Hilligoss}, \bibinfo{person}{Adam Nagy},
  {and} \bibinfo{person}{Madhulika Srikumar}.} \bibinfo{year}{2020}\natexlab{}.
\newblock \showarticletitle{{Principled Artificial Intelligence: Mapping
  Consensus in Ethical and Rights-Based Approaches to Principles for AI}}.
\newblock \bibinfo{journal}{\emph{SSRN Electronic Journal}}
  (\bibinfo{year}{2020}).
\newblock
\showISSN{1556-5068}
\urldef\tempurl%
\url{https://doi.org/10.2139/ssrn.3518482}
\showDOI{\tempurl}


\bibitem[\protect\citeauthoryear{Floridi}{Floridi}{2019}]%
        {floridi2019}
\bibfield{author}{\bibinfo{person}{Luciano Floridi}.}
  \bibinfo{year}{2019}\natexlab{}.
\newblock \showarticletitle{{Translating Principles into Practices of Digital
  Ethics: Five Risks of Being Unethical}}.
\newblock \bibinfo{journal}{\emph{Philosophy {\&} Technology}}
  \bibinfo{volume}{32}, \bibinfo{number}{2} (\bibinfo{date}{6}
  \bibinfo{year}{2019}).
\newblock
\showISSN{2210-5433}
\urldef\tempurl%
\url{https://doi.org/10.1007/s13347-019-00354-x}
\showDOI{\tempurl}


\bibitem[\protect\citeauthoryear{Floridi, Cowls, Beltrametti, Chatila,
  Chazerand, Dignum, Luetge, Madelin, Pagallo, Rossi, Schafer, Valcke, and
  Vayena}{Floridi et~al\mbox{.}}{2018}]%
        {floridi2018}
\bibfield{author}{\bibinfo{person}{Luciano Floridi}, \bibinfo{person}{Josh
  Cowls}, \bibinfo{person}{Monica Beltrametti}, \bibinfo{person}{Raja Chatila},
  \bibinfo{person}{Patrice Chazerand}, \bibinfo{person}{Virginia Dignum},
  \bibinfo{person}{Christoph Luetge}, \bibinfo{person}{Robert Madelin},
  \bibinfo{person}{Ugo Pagallo}, \bibinfo{person}{Francesca Rossi},
  \bibinfo{person}{Burkhard Schafer}, \bibinfo{person}{Peggy Valcke}, {and}
  \bibinfo{person}{Effy Vayena}.} \bibinfo{year}{2018}\natexlab{}.
\newblock \showarticletitle{{AI4People—An Ethical Framework for a Good AI
  Society: Opportunities, Risks, Principles, and Recommendations}}.
\newblock \bibinfo{journal}{\emph{Minds and Machines}} \bibinfo{volume}{28},
  \bibinfo{number}{4} (\bibinfo{date}{12} \bibinfo{year}{2018}).
\newblock
\showISSN{0924-6495}
\urldef\tempurl%
\url{https://doi.org/10.1007/s11023-018-9482-5}
\showDOI{\tempurl}


\bibitem[\protect\citeauthoryear{Frauenberger, Rauhala, and
  Fitzpatrick}{Frauenberger et~al\mbox{.}}{2016}]%
        {frauenberger2016}
\bibfield{author}{\bibinfo{person}{Christopher Frauenberger},
  \bibinfo{person}{Marjo Rauhala}, {and} \bibinfo{person}{Geraldine
  Fitzpatrick}.} \bibinfo{year}{2016}\natexlab{}.
\newblock \showarticletitle{{In-Action Ethics: Table 1.}}
\newblock \bibinfo{journal}{\emph{Interacting with Computers}}
  (\bibinfo{date}{6} \bibinfo{year}{2016}).
\newblock
\showISSN{0953-5438}
\urldef\tempurl%
\url{https://doi.org/10.1093/iwc/iww024}
\showDOI{\tempurl}


\bibitem[\protect\citeauthoryear{Fred~van Raaij and Verhallen}{Fred~van Raaij
  and Verhallen}{1994}]%
        {raaij1994}
\bibfield{author}{\bibinfo{person}{W. Fred~van Raaij} {and}
  \bibinfo{person}{Theo~M.M. Verhallen}.} \bibinfo{year}{1994}\natexlab{}.
\newblock \showarticletitle{{Domain‐specific Market Segmentation}}.
\newblock \bibinfo{journal}{\emph{European Journal of Marketing}}
  \bibinfo{volume}{28}, \bibinfo{number}{10} (\bibinfo{date}{10}
  \bibinfo{year}{1994}), \bibinfo{pages}{49--66}.
\newblock
\showISSN{0309-0566}
\urldef\tempurl%
\url{https://doi.org/10.1108/03090569410075786}
\showDOI{\tempurl}


\bibitem[\protect\citeauthoryear{Friedman, Hendry, and Borning}{Friedman
  et~al\mbox{.}}{2017}]%
        {friedman2017}
\bibfield{author}{\bibinfo{person}{Batya Friedman}, \bibinfo{person}{David~G.
  Hendry}, {and} \bibinfo{person}{Alan Borning}.}
  \bibinfo{year}{2017}\natexlab{}.
\newblock \showarticletitle{{A Survey of Value Sensitive Design Methods}}.
\newblock \bibinfo{journal}{\emph{Foundations and Trends{\textregistered} in
  Human–Computer Interaction}} \bibinfo{volume}{11}, \bibinfo{number}{2}
  (\bibinfo{year}{2017}), \bibinfo{pages}{63--125}.
\newblock
\showISSN{1551-3955}
\urldef\tempurl%
\url{https://doi.org/10.1561/1100000015}
\showDOI{\tempurl}


\bibitem[\protect\citeauthoryear{Gaillard}{Gaillard}{2016}]%
        {gaillard2016}
\bibfield{author}{\bibinfo{person}{Georges Gaillard}.}
  \bibinfo{year}{2016}\natexlab{}.
\newblock \showarticletitle{{La conflictualit{\'{e}} : une modalit{\'{e}} de
  lien o{\`{u}} s’arrime la destructivit{\'{e}} humaine}}.
\newblock \bibinfo{journal}{\emph{Connexions}} \bibinfo{volume}{106},
  \bibinfo{number}{2} (\bibinfo{year}{2016}), \bibinfo{pages}{71}.
\newblock
\showISSN{0337-3126}
\urldef\tempurl%
\url{https://doi.org/10.3917/cnx.106.0071}
\showDOI{\tempurl}


\bibitem[\protect\citeauthoryear{Gao, Liu, Zhang, Li, Zhu, Lin, and Yang}{Gao
  et~al\mbox{.}}{2020}]%
        {gao2020}
\bibfield{author}{\bibinfo{person}{Yanjie Gao}, \bibinfo{person}{Yu Liu},
  \bibinfo{person}{Hongyu Zhang}, \bibinfo{person}{Zhengxian Li},
  \bibinfo{person}{Yonghao Zhu}, \bibinfo{person}{Haoxiang Lin}, {and}
  \bibinfo{person}{Mao Yang}.} \bibinfo{year}{2020}\natexlab{}.
\newblock \showarticletitle{{Estimating GPU memory consumption of deep learning
  models}}. In \bibinfo{booktitle}{\emph{Proceedings of the 28th ACM Joint
  Meeting on European Software Engineering Conference and Symposium on the
  Foundations of Software Engineering}}. \bibinfo{publisher}{ACM},
  \bibinfo{address}{New York, NY, USA}, \bibinfo{pages}{1342--1352}.
\newblock
\showISBNx{9781450370431}
\urldef\tempurl%
\url{https://doi.org/10.1145/3368089.3417050}
\showDOI{\tempurl}


\bibitem[\protect\citeauthoryear{Garc{\'{i}}a-Mart{\'{i}}n, Rodrigues, Riley,
  and Grahn}{Garc{\'{i}}a-Mart{\'{i}}n et~al\mbox{.}}{2019}]%
        {garciamartin2019}
\bibfield{author}{\bibinfo{person}{Eva Garc{\'{i}}a-Mart{\'{i}}n},
  \bibinfo{person}{Crefeda~Faviola Rodrigues}, \bibinfo{person}{Graham Riley},
  {and} \bibinfo{person}{Håkan Grahn}.} \bibinfo{year}{2019}\natexlab{}.
\newblock \showarticletitle{{Estimation of energy consumption in machine
  learning}}.
\newblock \bibinfo{journal}{\emph{J. Parallel and Distrib. Comput.}}
  \bibinfo{volume}{134} (\bibinfo{date}{12} \bibinfo{year}{2019}),
  \bibinfo{pages}{75--88}.
\newblock
\showISSN{07437315}
\urldef\tempurl%
\url{https://doi.org/10.1016/j.jpdc.2019.07.007}
\showDOI{\tempurl}


\bibitem[\protect\citeauthoryear{Gebru, Jamie~Morgenstern, Vecchione, and
  Wortman~Vaughan}{Gebru et~al\mbox{.}}{2020}]%
        {gebru2020}
\bibfield{author}{\bibinfo{person}{Timnit Gebru}, \bibinfo{person}{Google
  Jamie~Morgenstern}, \bibinfo{person}{Briana Vecchione}, {and}
  \bibinfo{person}{Jennifer Wortman~Vaughan}.} \bibinfo{year}{2020}\natexlab{}.
\newblock \showarticletitle{{Datasheets for Datasets}}.
\newblock


\bibitem[\protect\citeauthoryear{Geiger, Yu, Yang, Dai, Qiu, Tang, and
  Huang}{Geiger et~al\mbox{.}}{2020}]%
        {stuartgeiger2020}
\bibfield{author}{\bibinfo{person}{R~Stuart Geiger}, \bibinfo{person}{Kevin
  Yu}, \bibinfo{person}{Yanlai Yang}, \bibinfo{person}{Mindy Dai},
  \bibinfo{person}{Jie Qiu}, \bibinfo{person}{Rebekah Tang}, {and}
  \bibinfo{person}{Jenny Huang}.} \bibinfo{year}{2020}\natexlab{}.
\newblock \showarticletitle{{Garbage in, Garbage out? Do Machine Learning
  Application Papers in Social Computing Report Where Human-Labeled Training
  Data Comes From?}}. In \bibinfo{booktitle}{\emph{Proceedings of the 2020
  Conference on Fairness, Accountability, and Transparency}}
  \emph{(\bibinfo{series}{FAT* '20})}. \bibinfo{publisher}{Association for
  Computing Machinery}, \bibinfo{address}{New York, NY, USA},
  \bibinfo{pages}{325--336}.
\newblock
\showISBNx{9781450369367}
\urldef\tempurl%
\url{https://doi.org/10.1145/3351095.3372862}
\showDOI{\tempurl}


\bibitem[\protect\citeauthoryear{Ghai, Liao, Zhang, and Mueller}{Ghai
  et~al\mbox{.}}{2020}]%
        {ghai2020}
\bibfield{author}{\bibinfo{person}{Bhavya Ghai}, \bibinfo{person}{Q.~Vera
  Liao}, \bibinfo{person}{Yunfeng Zhang}, {and} \bibinfo{person}{Klaus
  Mueller}.} \bibinfo{year}{2020}\natexlab{}.
\newblock \showarticletitle{{Measuring Social Biases of Crowd Workers using
  Counterfactual Queries}}.
\newblock  (\bibinfo{date}{4} \bibinfo{year}{2020}).
\newblock


\bibitem[\protect\citeauthoryear{Ghassemi, Oakden-Rayner, and Beam}{Ghassemi
  et~al\mbox{.}}{2021}]%
        {ghassemi2021}
\bibfield{author}{\bibinfo{person}{Marzyeh Ghassemi}, \bibinfo{person}{Luke
  Oakden-Rayner}, {and} \bibinfo{person}{Andrew~L Beam}.}
  \bibinfo{year}{2021}\natexlab{}.
\newblock \showarticletitle{{The false hope of current approaches to
  explainable artificial intelligence in health care}}.
\newblock \bibinfo{journal}{\emph{The Lancet Digital Health}}
  \bibinfo{volume}{3}, \bibinfo{number}{11} (\bibinfo{date}{11}
  \bibinfo{year}{2021}), \bibinfo{pages}{e745--e750}.
\newblock
\showISSN{25897500}
\urldef\tempurl%
\url{https://doi.org/10.1016/S2589-7500(21)00208-9}
\showDOI{\tempurl}


\bibitem[\protect\citeauthoryear{Globerson and Roweis}{Globerson and
  Roweis}{2006}]%
        {globerson2006}
\bibfield{author}{\bibinfo{person}{Amir Globerson} {and} \bibinfo{person}{Sam
  Roweis}.} \bibinfo{year}{2006}\natexlab{}.
\newblock \showarticletitle{{Nightmare at test time}}. In
  \bibinfo{booktitle}{\emph{Proceedings of the 23rd international conference on
  Machine learning - ICML '06}}. \bibinfo{publisher}{ACM Press},
  \bibinfo{address}{New York, New York, USA}, \bibinfo{pages}{353--360}.
\newblock
\showISBNx{1595933832}
\urldef\tempurl%
\url{https://doi.org/10.1145/1143844.1143889}
\showDOI{\tempurl}


\bibitem[\protect\citeauthoryear{Goodfellow, Shlens, and Szegedy}{Goodfellow
  et~al\mbox{.}}{2014}]%
        {goodfellow2014}
\bibfield{author}{\bibinfo{person}{Ian~J. Goodfellow},
  \bibinfo{person}{Jonathon Shlens}, {and} \bibinfo{person}{Christian
  Szegedy}.} \bibinfo{year}{2014}\natexlab{}.
\newblock \showarticletitle{{Explaining and Harnessing Adversarial Examples}}.
\newblock  (\bibinfo{date}{12} \bibinfo{year}{2014}).
\newblock


\bibitem[\protect\citeauthoryear{{Google}}{{Google}}{2018}]%
        {googleai2018}
\bibfield{author}{\bibinfo{person}{{Google}}.} \bibinfo{year}{2018}\natexlab{}.
\newblock \bibinfo{title}{{AI at Google: Our Principles}}.
\newblock
\newblock
\urldef\tempurl%
\url{https://www.blog.google/technology/ai/ai-principles/}
\showURL{%
\tempurl}


\bibitem[\protect\citeauthoryear{Green and Hu}{Green and Hu}{2018}]%
        {GreenHu2018}
\bibfield{author}{\bibinfo{person}{Ben Green} {and} \bibinfo{person}{Lily Hu}.}
  \bibinfo{year}{2018}\natexlab{}.
\newblock \showarticletitle{{The Myth in the Methodology: Towards a
  Recontextualization of Fairness in Machine Learning}}. In
  \bibinfo{booktitle}{\emph{Machine Learning: The Debates workshop at the 35th
  International Conference on Machine Learning (ICML)}}.
  \bibinfo{address}{Stockholm, Sweden}.
\newblock


\bibitem[\protect\citeauthoryear{Grgic-Hlaca, Zafar, Gummadi, and
  Weller}{Grgic-Hlaca et~al\mbox{.}}{2018}]%
        {grgic-hlaca2018}
\bibfield{author}{\bibinfo{person}{Nina Grgic-Hlaca},
  \bibinfo{person}{Muhammad~Bilal Zafar}, \bibinfo{person}{Krishna~P Gummadi},
  {and} \bibinfo{person}{Adrian Weller}.} \bibinfo{year}{2018}\natexlab{}.
\newblock \showarticletitle{{Beyond Distributive Fairness in Algorithmic
  Decision Making: Feature Selection for Procedurally Fair Learning}}.
\newblock In \bibinfo{booktitle}{\emph{Proceedings of the AAAI Conference on
  Artificial Intelligence}}. Vol.~\bibinfo{volume}{32}.
\newblock


\bibitem[\protect\citeauthoryear{Groves}{Groves}{2015}]%
        {groves2015}
\bibfield{author}{\bibinfo{person}{Christopher Groves}.}
  \bibinfo{year}{2015}\natexlab{}.
\newblock \showarticletitle{{Logic of Choice or Logic of Care? Uncertainty,
  Technological Mediation and Responsible Innovation}}.
\newblock \bibinfo{journal}{\emph{NanoEthics}} \bibinfo{volume}{9},
  \bibinfo{number}{3} (\bibinfo{date}{12} \bibinfo{year}{2015}),
  \bibinfo{pages}{321--333}.
\newblock
\showISSN{1871-4757}
\urldef\tempurl%
\url{https://doi.org/10.1007/s11569-015-0238-x}
\showDOI{\tempurl}


\bibitem[\protect\citeauthoryear{Hardt, Price, and Srebro}{Hardt
  et~al\mbox{.}}{2016}]%
        {hardt2016}
\bibfield{author}{\bibinfo{person}{Moritz Hardt}, \bibinfo{person}{Eric Price},
  {and} \bibinfo{person}{Nathan Srebro}.} \bibinfo{year}{2016}\natexlab{}.
\newblock \showarticletitle{{Equality of Opportunity in Supervised Learning}}.
  In \bibinfo{booktitle}{\emph{Proceedings of the 30th International Conference
  on Neural Information Processing Systems}}
  \emph{(\bibinfo{series}{NIPS'16})}. \bibinfo{publisher}{Curran Associates
  Inc.}, \bibinfo{address}{Red Hook, NY, USA}, \bibinfo{pages}{3323--3331}.
\newblock
\showISBNx{9781510838819}


\bibitem[\protect\citeauthoryear{Harrison, Hanson, Jacinto, Ramirez, and
  Ur}{Harrison et~al\mbox{.}}{2020}]%
        {harrison2020}
\bibfield{author}{\bibinfo{person}{Galen Harrison}, \bibinfo{person}{Julia
  Hanson}, \bibinfo{person}{Christine Jacinto}, \bibinfo{person}{Julio
  Ramirez}, {and} \bibinfo{person}{Blase Ur}.} \bibinfo{year}{2020}\natexlab{}.
\newblock \showarticletitle{{An empirical study on the perceived fairness of
  realistic, imperfect machine learning models}}. In
  \bibinfo{booktitle}{\emph{Proceedings of the 2020 Conference on Fairness,
  Accountability, and Transparency}}. \bibinfo{publisher}{ACM},
  \bibinfo{address}{New York, NY, USA}, \bibinfo{pages}{392--402}.
\newblock
\showISBNx{9781450369367}
\urldef\tempurl%
\url{https://doi.org/10.1145/3351095.3372831}
\showDOI{\tempurl}


\bibitem[\protect\citeauthoryear{Heijne and van~der Meer}{Heijne and van~der
  Meer}{2019}]%
        {heijne2019}
\bibfield{author}{\bibinfo{person}{Katrina Heijne} {and} \bibinfo{person}{Han
  van~der Meer}.} \bibinfo{year}{2019}\natexlab{}.
\newblock \bibinfo{booktitle}{\emph{{Road Map for Creative Problem Solving
  Techniques Organizing and facilitating group sessions}}}.
\newblock \bibinfo{publisher}{Boom Uitgevers Amsterdam}.
\newblock


\bibitem[\protect\citeauthoryear{Hemment, Aylett, Belle, Murray-Rust, Luger,
  Hillston, Rovatsos, and Broz}{Hemment et~al\mbox{.}}{2019}]%
        {hemment2019}
\bibfield{author}{\bibinfo{person}{Drew Hemment}, \bibinfo{person}{Ruth
  Aylett}, \bibinfo{person}{Vaishak Belle}, \bibinfo{person}{Dave Murray-Rust},
  \bibinfo{person}{Ewa Luger}, \bibinfo{person}{Jane Hillston},
  \bibinfo{person}{Michael Rovatsos}, {and} \bibinfo{person}{Frank Broz}.}
  \bibinfo{year}{2019}\natexlab{}.
\newblock \showarticletitle{{Experiential AI}}.
\newblock \bibinfo{journal}{\emph{AI Matters}} \bibinfo{volume}{5},
  \bibinfo{number}{1} (\bibinfo{date}{4} \bibinfo{year}{2019}),
  \bibinfo{pages}{25--31}.
\newblock
\urldef\tempurl%
\url{https://doi.org/10.1145/3320254.3320264}
\showDOI{\tempurl}


\bibitem[\protect\citeauthoryear{Henderson, Hu, Romoff, Brunskill, Jurafsky,
  and Pineau}{Henderson et~al\mbox{.}}{2020}]%
        {henderson2020}
\bibfield{author}{\bibinfo{person}{Peter Henderson}, \bibinfo{person}{Jieru
  Hu}, \bibinfo{person}{Joshua Romoff}, \bibinfo{person}{Emma Brunskill},
  \bibinfo{person}{Dan Jurafsky}, {and} \bibinfo{person}{Joelle Pineau}.}
  \bibinfo{year}{2020}\natexlab{}.
\newblock \showarticletitle{{Towards the Systematic Reporting of the Energy and
  Carbon Footprints of Machine Learning}}.
\newblock  (\bibinfo{date}{1} \bibinfo{year}{2020}).
\newblock


\bibitem[\protect\citeauthoryear{Henin and Le~M{\'{e}}tayer}{Henin and
  Le~M{\'{e}}tayer}{2021}]%
        {henin2021}
\bibfield{author}{\bibinfo{person}{Clément Henin} {and}
  \bibinfo{person}{Daniel Le~M{\'{e}}tayer}.} \bibinfo{year}{2021}\natexlab{}.
\newblock \showarticletitle{{Beyond explainability: justifiability and
  contestability of algorithmic decision systems}}.
\newblock \bibinfo{journal}{\emph{AI {\&} SOCIETY}} (\bibinfo{date}{7}
  \bibinfo{year}{2021}).
\newblock
\showISSN{0951-5666}
\urldef\tempurl%
\url{https://doi.org/10.1007/s00146-021-01251-8}
\showDOI{\tempurl}


\bibitem[\protect\citeauthoryear{Herder and van Maaren}{Herder and van
  Maaren}{2020}]%
        {herder2020}
\bibfield{author}{\bibinfo{person}{Eelco Herder} {and} \bibinfo{person}{Olaf
  van Maaren}.} \bibinfo{year}{2020}\natexlab{}.
\newblock \showarticletitle{{Privacy Dashboards: The Impact of the Type of
  Personal Data and User Control on Trust and Perceived Risk}}. In
  \bibinfo{booktitle}{\emph{Adjunct Publication of the 28th ACM Conference on
  User Modeling, Adaptation and Personalization}}. \bibinfo{publisher}{ACM},
  \bibinfo{address}{New York, NY, USA}, \bibinfo{pages}{169--174}.
\newblock
\showISBNx{9781450379502}
\urldef\tempurl%
\url{https://doi.org/10.1145/3386392.3399557}
\showDOI{\tempurl}


\bibitem[\protect\citeauthoryear{Hidalgo, Orghian, Albo-Canals, de~Almeida, and
  Martin}{Hidalgo et~al\mbox{.}}{2021}]%
        {hidalgo2021}
\bibfield{author}{\bibinfo{person}{C{\'e}sar Hidalgo}, \bibinfo{person}{Diana
  Orghian}, \bibinfo{person}{Jordi Albo-Canals}, \bibinfo{person}{Filipa de
  Almeida}, {and} \bibinfo{person}{Natalia Martin}.}
  \bibinfo{year}{2021}\natexlab{}.
\newblock \bibinfo{booktitle}{\emph{{How Humans Judge Machines}}}.
\newblock \bibinfo{publisher}{MIT Press}.
\newblock
\urldef\tempurl%
\url{https://hal.archives-ouvertes.fr/hal-03058652}
\showURL{%
\tempurl}


\bibitem[\protect\citeauthoryear{Hirsch, Merced, Narayanan, Imel, and
  Atkins}{Hirsch et~al\mbox{.}}{2017}]%
        {hirsch2017}
\bibfield{author}{\bibinfo{person}{Tad Hirsch}, \bibinfo{person}{Kritzia
  Merced}, \bibinfo{person}{Shrikanth Narayanan}, \bibinfo{person}{Zac~E.
  Imel}, {and} \bibinfo{person}{David~C. Atkins}.}
  \bibinfo{year}{2017}\natexlab{}.
\newblock \showarticletitle{{Designing Contestability}}. In
  \bibinfo{booktitle}{\emph{Proceedings of the 2017 Conference on Designing
  Interactive Systems}}. \bibinfo{publisher}{ACM}, \bibinfo{address}{New York,
  NY, USA}.
\newblock
\showISBNx{9781450349222}
\urldef\tempurl%
\url{https://doi.org/10.1145/3064663.3064703}
\showDOI{\tempurl}


\bibitem[\protect\citeauthoryear{Holland, Hosny, Newman, Joseph, and
  Chmielinski}{Holland et~al\mbox{.}}{2018}]%
        {holland2018}
\bibfield{author}{\bibinfo{person}{Sarah Holland}, \bibinfo{person}{Ahmed
  Hosny}, \bibinfo{person}{Sarah Newman}, \bibinfo{person}{Joshua Joseph},
  {and} \bibinfo{person}{Kasia Chmielinski}.} \bibinfo{year}{2018}\natexlab{}.
\newblock \showarticletitle{{The Dataset Nutrition Label: A Framework To Drive
  Higher Data Quality Standards}}.
\newblock  (\bibinfo{date}{5} \bibinfo{year}{2018}).
\newblock


\bibitem[\protect\citeauthoryear{Holtz, Nocun, and Hansen}{Holtz
  et~al\mbox{.}}{2011}]%
        {holtz2011}
\bibfield{author}{\bibinfo{person}{Leif-Erik Holtz}, \bibinfo{person}{Katharina
  Nocun}, {and} \bibinfo{person}{Marit Hansen}.}
  \bibinfo{year}{2011}\natexlab{}.
\newblock \showarticletitle{{Towards Displaying Privacy Information with
  Icons}}.
\newblock \bibinfo{pages}{338--348}.
\newblock
\urldef\tempurl%
\url{https://doi.org/10.1007/978-3-642-20769-3_27}
\showDOI{\tempurl}


\bibitem[\protect\citeauthoryear{Hong, Fourney, DeBellis, and Amershi}{Hong
  et~al\mbox{.}}{2021}]%
        {hong2021}
\bibfield{author}{\bibinfo{person}{Matthew~K. Hong}, \bibinfo{person}{Adam
  Fourney}, \bibinfo{person}{Derek DeBellis}, {and} \bibinfo{person}{Saleema
  Amershi}.} \bibinfo{year}{2021}\natexlab{}.
\newblock \showarticletitle{{Planning for Natural Language Failures with the AI
  Playbook}}. In \bibinfo{booktitle}{\emph{Proceedings of the 2021 CHI
  Conference on Human Factors in Computing Systems}}. \bibinfo{publisher}{ACM},
  \bibinfo{address}{New York, NY, USA}, \bibinfo{pages}{1--11}.
\newblock
\showISBNx{9781450380966}
\urldef\tempurl%
\url{https://doi.org/10.1145/3411764.3445735}
\showDOI{\tempurl}


\bibitem[\protect\citeauthoryear{Hube, Fetahu, and Gadiraju}{Hube
  et~al\mbox{.}}{2019}]%
        {hube2019}
\bibfield{author}{\bibinfo{person}{Christoph Hube}, \bibinfo{person}{Besnik
  Fetahu}, {and} \bibinfo{person}{Ujwal Gadiraju}.}
  \bibinfo{year}{2019}\natexlab{}.
\newblock \showarticletitle{{Understanding and Mitigating Worker Biases in the
  Crowdsourced Collection of Subjective Judgments}}. In
  \bibinfo{booktitle}{\emph{Proceedings of the 2019 CHI Conference on Human
  Factors in Computing Systems}}. \bibinfo{publisher}{Association for Computing
  Machinery}, \bibinfo{address}{New York, NY, USA}, \bibinfo{pages}{1--12}.
\newblock
\showISBNx{9781450359702}
\urldef\tempurl%
\url{https://doi-org.tudelft.idm.oclc.org/10.1145/3290605.3300637}
\showURL{%
\tempurl}


\bibitem[\protect\citeauthoryear{Hutchinson, Smart, Hanna, Denton, Greer,
  Kjartansson, Barnes, and Mitchell}{Hutchinson et~al\mbox{.}}{2021}]%
        {hutchinson2021}
\bibfield{author}{\bibinfo{person}{Ben Hutchinson}, \bibinfo{person}{Andrew
  Smart}, \bibinfo{person}{Alex Hanna}, \bibinfo{person}{Emily Denton},
  \bibinfo{person}{Christina Greer}, \bibinfo{person}{Oddur Kjartansson},
  \bibinfo{person}{Parker Barnes}, {and} \bibinfo{person}{Margaret Mitchell}.}
  \bibinfo{year}{2021}\natexlab{}.
\newblock \showarticletitle{{Towards Accountability for Machine Learning
  Datasets: Practices from Software Engineering and Infrastructure}}. In
  \bibinfo{booktitle}{\emph{Proceedings of the 2021 ACM Conference on Fairness,
  Accountability, and Transparency}} \emph{(\bibinfo{series}{FAccT '21})}.
  \bibinfo{publisher}{Association for Computing Machinery},
  \bibinfo{address}{New York, NY, USA}, \bibinfo{pages}{560--575}.
\newblock
\showISBNx{9781450383097}
\urldef\tempurl%
\url{https://doi.org/10.1145/3442188.3445918}
\showDOI{\tempurl}


\bibitem[\protect\citeauthoryear{{IBM}}{{IBM}}{2019}]%
        {IBMai2019}
\bibfield{author}{\bibinfo{person}{{IBM}}.} \bibinfo{year}{2019}\natexlab{}.
\newblock \bibinfo{title}{{IBM Everyday Ethics for AI}}.
\newblock
\newblock
\urldef\tempurl%
\url{https://www.ibm.com/watson/assets/duo/pdf/everydayethics.pdf}
\showURL{%
\tempurl}


\bibitem[\protect\citeauthoryear{{IEEE}}{{IEEE}}{2008}]%
        {IEEEaudit}
\bibfield{author}{\bibinfo{person}{{IEEE}}.} \bibinfo{year}{2008}\natexlab{}.
\newblock \showarticletitle{{IEEE Standard for Software Reviews and Audits}}.
\newblock \bibinfo{journal}{\emph{IEEE Std 1028-2008}} (\bibinfo{year}{2008}),
  \bibinfo{pages}{1--53}.
\newblock
\urldef\tempurl%
\url{https://doi.org/10.1109/IEEESTD.2008.4601584}
\showDOI{\tempurl}


\bibitem[\protect\citeauthoryear{Ionescu, Hann{\'{a}}k, and Joseph}{Ionescu
  et~al\mbox{.}}{2021}]%
        {ionescu2021}
\bibfield{author}{\bibinfo{person}{Stefania Ionescu}, \bibinfo{person}{Anikó
  Hann{\'{a}}k}, {and} \bibinfo{person}{Kenneth Joseph}.}
  \bibinfo{year}{2021}\natexlab{}.
\newblock \showarticletitle{{An Agent-Based Model to Evaluate Interventions on
  Online Dating Platforms to Decrease Racial Homogamy}}. In
  \bibinfo{booktitle}{\emph{Proceedings of the 2021 ACM Conference on Fairness,
  Accountability, and Transparency}} \emph{(\bibinfo{series}{FAccT '21})}.
  \bibinfo{publisher}{Association for Computing Machinery},
  \bibinfo{address}{New York, NY, USA}, \bibinfo{pages}{412--423}.
\newblock
\showISBNx{9781450383097}
\urldef\tempurl%
\url{https://doi.org/10.1145/3442188.3445904}
\showDOI{\tempurl}


\bibitem[\protect\citeauthoryear{Japanese Cabinet~Office and
  Innovation}{Japanese Cabinet~Office and Innovation}{2019}]%
        {japanai2019}
\bibfield{author}{\bibinfo{person}{Technology Japanese Cabinet~Office, Council
  for~Science} {and} \bibinfo{person}{Innovation}.}
  \bibinfo{year}{2019}\natexlab{}.
\newblock \bibinfo{title}{{Social Principles of Human-Centric Artificial
  Intelligence}}.
\newblock
\newblock
\urldef\tempurl%
\url{https://www8.cao.go.jp/cstp/english/humancentricai.pdf}
\showURL{%
\tempurl}


\bibitem[\protect\citeauthoryear{Jia, Salem, Backes, Zhang, and Gong}{Jia
  et~al\mbox{.}}{2019}]%
        {jia2019}
\bibfield{author}{\bibinfo{person}{Jinyuan Jia}, \bibinfo{person}{Ahmed Salem},
  \bibinfo{person}{Michael Backes}, \bibinfo{person}{Yang Zhang}, {and}
  \bibinfo{person}{Neil~Zhenqiang Gong}.} \bibinfo{year}{2019}\natexlab{}.
\newblock \showarticletitle{{MemGuard: Defending against Black-Box Membership
  Inference Attacks via Adversarial Examples}}.
\newblock  (\bibinfo{date}{9} \bibinfo{year}{2019}).
\newblock


\bibitem[\protect\citeauthoryear{Jin, Fan, Gromala, Pasquier, and Hamarneh}{Jin
  et~al\mbox{.}}{2021}]%
        {jin2021}
\bibfield{author}{\bibinfo{person}{Weina Jin}, \bibinfo{person}{Jianyu Fan},
  \bibinfo{person}{Diane Gromala}, \bibinfo{person}{Philippe Pasquier}, {and}
  \bibinfo{person}{Ghassan Hamarneh}.} \bibinfo{year}{2021}\natexlab{}.
\newblock \showarticletitle{{EUCA: A Practical Prototyping Framework towards
  End-User-Centered Explainable Artificial Intelligence}}.
\newblock  (\bibinfo{date}{2} \bibinfo{year}{2021}).
\newblock
\urldef\tempurl%
\url{https://arxiv.org/abs/2102.02437}
\showURL{%
\tempurl}


\bibitem[\protect\citeauthoryear{Kaiser and Rauchfleisch}{Kaiser and
  Rauchfleisch}{2020}]%
        {kaiser2020}
\bibfield{author}{\bibinfo{person}{Jonas Kaiser} {and} \bibinfo{person}{Adrian
  Rauchfleisch}.} \bibinfo{year}{2020}\natexlab{}.
\newblock \showarticletitle{{Birds of a Feather Get Recommended Together:
  Algorithmic Homophily in YouTube’s Channel Recommendations in the United
  States and Germany}}.
\newblock \bibinfo{journal}{\emph{Social Media + Society}} \bibinfo{volume}{6},
  \bibinfo{number}{4} (\bibinfo{date}{10} \bibinfo{year}{2020}),
  \bibinfo{pages}{2056305120969914}.
\newblock
\showISSN{2056-3051}
\urldef\tempurl%
\url{https://doi.org/10.1177/2056305120969914}
\showDOI{\tempurl}


\bibitem[\protect\citeauthoryear{Kalluri}{Kalluri}{2020}]%
        {kalluri2020}
\bibfield{author}{\bibinfo{person}{Pratyusha Kalluri}.}
  \bibinfo{year}{2020}\natexlab{}.
\newblock \showarticletitle{{Don't ask if artificial intelligence is good or
  fair, ask how it shifts power.}}
\newblock \bibinfo{journal}{\emph{Nature}} \bibinfo{volume}{583},
  \bibinfo{number}{7815} (\bibinfo{year}{2020}).
\newblock
\showISSN{1476-4687}
\urldef\tempurl%
\url{https://doi.org/10.1038/d41586-020-02003-2}
\showDOI{\tempurl}


\bibitem[\protect\citeauthoryear{Kaya, Hong, and Dumitras}{Kaya
  et~al\mbox{.}}{2020}]%
        {kaya2020}
\bibfield{author}{\bibinfo{person}{Yigitcan Kaya}, \bibinfo{person}{Sanghyun
  Hong}, {and} \bibinfo{person}{Tudor Dumitras}.}
  \bibinfo{year}{2020}\natexlab{}.
\newblock \showarticletitle{{On the Effectiveness of Regularization Against
  Membership Inference Attacks}}.
\newblock  (\bibinfo{date}{6} \bibinfo{year}{2020}).
\newblock


\bibitem[\protect\citeauthoryear{Kearns, Neel, Roth, and Wu}{Kearns
  et~al\mbox{.}}{2018}]%
        {kearns2018}
\bibfield{author}{\bibinfo{person}{Michael Kearns}, \bibinfo{person}{Seth
  Neel}, \bibinfo{person}{Aaron Roth}, {and} \bibinfo{person}{Zhiwei~Steven
  Wu}.} \bibinfo{year}{2018}\natexlab{}.
\newblock \showarticletitle{{Preventing Fairness Gerrymandering: Auditing and
  Learning for Subgroup Fairness}}. In \bibinfo{booktitle}{\emph{Proceedings of
  the 35th International Conference on Machine Learning}}
  \emph{(\bibinfo{series}{Proceedings of Machine Learning Research},
  Vol.~\bibinfo{volume}{80})}, \bibfield{editor}{\bibinfo{person}{Jennifer Dy}
  {and} \bibinfo{person}{Andreas Krause}} (Eds.). \bibinfo{publisher}{PMLR},
  \bibinfo{pages}{2564--2572}.
\newblock
\urldef\tempurl%
\url{https://proceedings.mlr.press/v80/kearns18a.html}
\showURL{%
\tempurl}


\bibitem[\protect\citeauthoryear{Kearns and Roth}{Kearns and Roth}{2019}]%
        {kearns2019}
\bibfield{author}{\bibinfo{person}{Michael Kearns} {and} \bibinfo{person}{Aaron
  Roth}.} \bibinfo{year}{2019}\natexlab{}.
\newblock \bibinfo{booktitle}{\emph{{The Ethical Algorithm: The Science of
  Socially Aware Algorithm Design}}}.
\newblock \bibinfo{publisher}{Oxford University Press, Inc.},
  \bibinfo{address}{USA}.
\newblock
\showISBNx{0190948205}


\bibitem[\protect\citeauthoryear{Keyes, Hutson, and Durbin}{Keyes
  et~al\mbox{.}}{2019}]%
        {keyes2019}
\bibfield{author}{\bibinfo{person}{Os Keyes}, \bibinfo{person}{Jevan Hutson},
  {and} \bibinfo{person}{Meredith Durbin}.} \bibinfo{year}{2019}\natexlab{}.
\newblock \showarticletitle{{A Mulching Proposal: Analysing and Improving an
  Algorithmic System for Turning the Elderly into High-Nutrient Slurry}}. In
  \bibinfo{booktitle}{\emph{Extended Abstracts of the 2019 CHI Conference on
  Human Factors in Computing Systems}} \emph{(\bibinfo{series}{CHI EA '19})}.
  \bibinfo{publisher}{Association for Computing Machinery},
  \bibinfo{address}{New York, NY, USA}, \bibinfo{pages}{1--11}.
\newblock
\showISBNx{9781450359719}
\urldef\tempurl%
\url{https://doi.org/10.1145/3290607.3310433}
\showDOI{\tempurl}


\bibitem[\protect\citeauthoryear{Kleinberg, Mullainathan, and
  Raghavan}{Kleinberg et~al\mbox{.}}{2016}]%
        {kleinberg2016}
\bibfield{author}{\bibinfo{person}{Jon Kleinberg}, \bibinfo{person}{Sendhil
  Mullainathan}, {and} \bibinfo{person}{Manish Raghavan}.}
  \bibinfo{year}{2016}\natexlab{}.
\newblock \showarticletitle{{Inherent Trade-Offs in the Fair Determination of
  Risk Scores}}.
\newblock  (\bibinfo{date}{9} \bibinfo{year}{2016}).
\newblock


\bibitem[\protect\citeauthoryear{Kluttz, Kohli, and Mulligan}{Kluttz
  et~al\mbox{.}}{2018}]%
        {kluttz2018}
\bibfield{author}{\bibinfo{person}{Daniel Kluttz}, \bibinfo{person}{Nitin
  Kohli}, {and} \bibinfo{person}{Deirdre~K. Mulligan}.}
  \bibinfo{year}{2018}\natexlab{}.
\newblock \showarticletitle{{Contestability and Professionals: From
  Explanations to Engagement with Algorithmic Systems}}.
\newblock \bibinfo{journal}{\emph{SSRN Electronic Journal}}
  (\bibinfo{year}{2018}).
\newblock
\showISSN{1556-5068}
\urldef\tempurl%
\url{https://doi.org/10.2139/ssrn.3311894}
\showDOI{\tempurl}


\bibitem[\protect\citeauthoryear{Krafft and Zweig}{Krafft and Zweig}{2019}]%
        {krafft2019}
\bibfield{author}{\bibinfo{person}{TD Krafft} {and} \bibinfo{person}{K Zweig}.}
  \bibinfo{year}{2019}\natexlab{}.
\newblock \showarticletitle{{Transparenz und Nachvollziehbarkeit
  algorithmenbasierter Entscheidungsprozesse}}.
\newblock \bibinfo{journal}{\emph{Ein Regulierungsvorschlag}}
  (\bibinfo{year}{2019}).
\newblock


\bibitem[\protect\citeauthoryear{Kuijer and Giaccardi}{Kuijer and
  Giaccardi}{2018}]%
        {kuijer2018}
\bibfield{author}{\bibinfo{person}{Lenneke Kuijer} {and} \bibinfo{person}{Elisa
  Giaccardi}.} \bibinfo{year}{2018}\natexlab{}.
\newblock \showarticletitle{{Co-Performance: Conceptualizing the Role of
  Artificial Agency in the Design of Everyday Life}}. In
  \bibinfo{booktitle}{\emph{Proceedings of the 2018 CHI Conference on Human
  Factors in Computing Systems}}. \bibinfo{publisher}{Association for Computing
  Machinery}, \bibinfo{address}{New York, NY, USA}, \bibinfo{pages}{1--13}.
\newblock
\showISBNx{9781450356206}
\urldef\tempurl%
\url{https://doi-org.tudelft.idm.oclc.org/10.1145/3173574.3173699}
\showURL{%
\tempurl}


\bibitem[\protect\citeauthoryear{Kulynych, Overdorf, Troncoso, and
  G{\"{u}}rses}{Kulynych et~al\mbox{.}}{2018}]%
        {Kulynych2020}
\bibfield{author}{\bibinfo{person}{Bogdan Kulynych}, \bibinfo{person}{Rebekah
  Overdorf}, \bibinfo{person}{Carmela Troncoso}, {and} \bibinfo{person}{Seda
  G{\"{u}}rses}.} \bibinfo{year}{2018}\natexlab{}.
\newblock \showarticletitle{{POTs: Protective Optimization Technologies}}.
\newblock  (\bibinfo{date}{6} \bibinfo{year}{2018}).
\newblock
\urldef\tempurl%
\url{https://doi.org/10.1145/3351095.3372853}
\showDOI{\tempurl}


\bibitem[\protect\citeauthoryear{Kusner, Loftus, Russell, and Silva}{Kusner
  et~al\mbox{.}}{2017}]%
        {kusner2017}
\bibfield{author}{\bibinfo{person}{Matt~J Kusner}, \bibinfo{person}{Joshua
  Loftus}, \bibinfo{person}{Chris Russell}, {and} \bibinfo{person}{Ricardo
  Silva}.} \bibinfo{year}{2017}\natexlab{}.
\newblock \showarticletitle{{Counterfactual Fairness}}. In
  \bibinfo{booktitle}{\emph{Advances in Neural Information Processing
  Systems}}, \bibfield{editor}{\bibinfo{person}{I~Guyon}, \bibinfo{person}{U~V
  Luxburg}, \bibinfo{person}{S~Bengio}, \bibinfo{person}{H~Wallach},
  \bibinfo{person}{R~Fergus}, \bibinfo{person}{S~Vishwanathan}, {and}
  \bibinfo{person}{R~Garnett}} (Eds.), Vol.~\bibinfo{volume}{30}.
  \bibinfo{publisher}{Curran Associates, Inc.}
\newblock
\urldef\tempurl%
\url{https://proceedings.neurips.cc/paper/2017/file/a486cd07e4ac3d270571622f4f316ec5-Paper.pdf}
\showURL{%
\tempurl}


\bibitem[\protect\citeauthoryear{Larsonneur}{Larsonneur}{2021}]%
        {larsonnneur2021}
\bibfield{author}{\bibinfo{person}{Claire Larsonneur}.}
  \bibinfo{year}{2021}\natexlab{}.
\newblock \bibinfo{title}{{Intelligence artificielle ET/OU diversit{\'{e}}
  linguistique : les paradoxes du traitement automatique des langues}}.
\newblock
\newblock
\urldef\tempurl%
\url{http://www.hybrid.univ-paris8.fr/lodel/index.php?id=1542}
\showURL{%
\tempurl}


\bibitem[\protect\citeauthoryear{Lee}{Lee}{1973}]%
        {lee1973}
\bibfield{author}{\bibinfo{person}{Douglass~B. Lee}.}
  \bibinfo{year}{1973}\natexlab{}.
\newblock \showarticletitle{{Requiem for Large-Scale Models}}.
\newblock \bibinfo{journal}{\emph{Journal of the American Institute of
  Planners}} \bibinfo{volume}{39}, \bibinfo{number}{3} (\bibinfo{date}{5}
  \bibinfo{year}{1973}), \bibinfo{pages}{163--178}.
\newblock
\showISSN{0002-8991}
\urldef\tempurl%
\url{https://doi.org/10.1080/01944367308977851}
\showDOI{\tempurl}


\bibitem[\protect\citeauthoryear{Lee and Baykal}{Lee and Baykal}{2017}]%
        {kyunglee2017}
\bibfield{author}{\bibinfo{person}{Min~Kyung Lee} {and} \bibinfo{person}{Su
  Baykal}.} \bibinfo{year}{2017}\natexlab{}.
\newblock \showarticletitle{{Algorithmic Mediation in Group Decisions: Fairness
  Perceptions of Algorithmically Mediated vs. Discussion-Based Social
  Division}}. In \bibinfo{booktitle}{\emph{Proceedings of the 2017 ACM
  Conference on Computer Supported Cooperative Work and Social Computing}}
  \emph{(\bibinfo{series}{CSCW '17})}. \bibinfo{publisher}{Association for
  Computing Machinery}, \bibinfo{address}{New York, NY, USA},
  \bibinfo{pages}{1035--1048}.
\newblock
\showISBNx{9781450343350}
\urldef\tempurl%
\url{https://doi.org/10.1145/2998181.2998230}
\showDOI{\tempurl}


\bibitem[\protect\citeauthoryear{Lee, Kusbit, Kahng, Kim, Yuan, Chan, See,
  Noothigattu, Lee, Psomas, and Procaccia}{Lee et~al\mbox{.}}{2019}]%
        {lee2019}
\bibfield{author}{\bibinfo{person}{Min~Kyung Lee}, \bibinfo{person}{Daniel
  Kusbit}, \bibinfo{person}{Anson Kahng}, \bibinfo{person}{Ji~Tae Kim},
  \bibinfo{person}{Xinran Yuan}, \bibinfo{person}{Allissa Chan},
  \bibinfo{person}{Daniel See}, \bibinfo{person}{Ritesh Noothigattu},
  \bibinfo{person}{Siheon Lee}, \bibinfo{person}{Alexandros Psomas}, {and}
  \bibinfo{person}{Ariel~D Procaccia}.} \bibinfo{year}{2019}\natexlab{}.
\newblock \showarticletitle{{WeBuildAI: Participatory Framework for Algorithmic
  Governance}}.
\newblock \bibinfo{journal}{\emph{Proc. ACM Hum.-Comput. Interact.}}
  \bibinfo{volume}{3}, \bibinfo{number}{CSCW} (\bibinfo{date}{11}
  \bibinfo{year}{2019}).
\newblock
\urldef\tempurl%
\url{https://doi.org/10.1145/3359283}
\showDOI{\tempurl}


\bibitem[\protect\citeauthoryear{Lee and Singh}{Lee and Singh}{2021}]%
        {lee2021}
\bibfield{author}{\bibinfo{person}{Michelle Seng~Ah Lee} {and}
  \bibinfo{person}{Jatinder Singh}.} \bibinfo{year}{2021}\natexlab{}.
\newblock \showarticletitle{{Risk Identification Questionnaire for Unintended
  Bias in Machine Learning Development Lifecycle}}.
\newblock \bibinfo{journal}{\emph{SSRN Electronic Journal}}
  (\bibinfo{year}{2021}).
\newblock
\showISSN{1556-5068}
\urldef\tempurl%
\url{https://doi.org/10.2139/ssrn.3777093}
\showDOI{\tempurl}


\bibitem[\protect\citeauthoryear{leins, Lau, and Baldwin}{leins
  et~al\mbox{.}}{2020}]%
        {leins2020}
\bibfield{author}{\bibinfo{person}{kobi leins}, \bibinfo{person}{Jey~Han Lau},
  {and} \bibinfo{person}{Timothy Baldwin}.} \bibinfo{year}{2020}\natexlab{}.
\newblock \showarticletitle{{Give Me Convenience and Give Her Death: Who Should
  Decide What Uses of NLP are Appropriate, and on What Basis?}}. In
  \bibinfo{booktitle}{\emph{Proceedings of the 58th Annual Meeting of the
  Association for Computational Linguistics}}. \bibinfo{publisher}{Association
  for Computational Linguistics}, \bibinfo{address}{Stroudsburg, PA, USA}.
\newblock
\urldef\tempurl%
\url{https://doi.org/10.18653/v1/2020.acl-main.261}
\showDOI{\tempurl}


\bibitem[\protect\citeauthoryear{Liao, Gruen, and Miller}{Liao
  et~al\mbox{.}}{2020}]%
        {veraliao2020}
\bibfield{author}{\bibinfo{person}{Q.~Vera Liao}, \bibinfo{person}{Daniel
  Gruen}, {and} \bibinfo{person}{Sarah Miller}.}
  \bibinfo{year}{2020}\natexlab{}.
\newblock \showarticletitle{{Questioning the AI: Informing Design Practices for
  Explainable AI User Experiences}}.
\newblock  (\bibinfo{date}{1} \bibinfo{year}{2020}).
\newblock
\urldef\tempurl%
\url{https://doi.org/10.1145/3313831.3376590}
\showDOI{\tempurl}


\bibitem[\protect\citeauthoryear{Liao, Pribi{\'{c}}, Han, Miller, and Sow}{Liao
  et~al\mbox{.}}{2021}]%
        {veraliao2021}
\bibfield{author}{\bibinfo{person}{Q.~Vera Liao}, \bibinfo{person}{Milena
  Pribi{\'{c}}}, \bibinfo{person}{Jaesik Han}, \bibinfo{person}{Sarah Miller},
  {and} \bibinfo{person}{Daby Sow}.} \bibinfo{year}{2021}\natexlab{}.
\newblock \showarticletitle{{Question-Driven Design Process for Explainable AI
  User Experiences}}.
\newblock  (\bibinfo{date}{4} \bibinfo{year}{2021}).
\newblock


\bibitem[\protect\citeauthoryear{Liscio, van~der Meer, Siebert, Jonker, Mouter,
  and Murukannaiah}{Liscio et~al\mbox{.}}{2021}]%
        {liscio2021}
\bibfield{author}{\bibinfo{person}{Enrico Liscio}, \bibinfo{person}{Michiel
  van~der Meer}, \bibinfo{person}{Luciano~C Siebert},
  \bibinfo{person}{Catholijn~M Jonker}, \bibinfo{person}{Niek Mouter}, {and}
  \bibinfo{person}{Pradeep~K Murukannaiah}.} \bibinfo{year}{2021}\natexlab{}.
\newblock \showarticletitle{{Axies: Identifying and Evaluating Context-Specific
  Values}}. In \bibinfo{booktitle}{\emph{Proceedings of the 20th International
  Conference on Autonomous Agents and MultiAgent Systems}}.
  \bibinfo{publisher}{International Foundation for Autonomous Agents and
  Multiagent Systems}, \bibinfo{address}{Richland, SC},
  \bibinfo{pages}{799--808}.
\newblock
\showISBNx{9781450383073}


\bibitem[\protect\citeauthoryear{Long and Magerko}{Long and Magerko}{2020}]%
        {long2020}
\bibfield{author}{\bibinfo{person}{Duri Long} {and} \bibinfo{person}{Brian
  Magerko}.} \bibinfo{year}{2020}\natexlab{}.
\newblock \showarticletitle{{What is AI Literacy? Competencies and Design
  Considerations}}. In \bibinfo{booktitle}{\emph{Proceedings of the 2020 CHI
  Conference on Human Factors in Computing Systems}}.
  \bibinfo{publisher}{Association for Computing Machinery},
  \bibinfo{address}{New York, NY, USA}, \bibinfo{pages}{1--16}.
\newblock
\showISBNx{9781450367080}
\urldef\tempurl%
\url{https://doi-org.tudelft.idm.oclc.org/10.1145/3313831.3376727}
\showURL{%
\tempurl}


\bibitem[\protect\citeauthoryear{Lyons, Velloso, and Miller}{Lyons
  et~al\mbox{.}}{2021}]%
        {lyons2021}
\bibfield{author}{\bibinfo{person}{Henrietta Lyons}, \bibinfo{person}{Eduardo
  Velloso}, {and} \bibinfo{person}{Tim Miller}.}
  \bibinfo{year}{2021}\natexlab{}.
\newblock \showarticletitle{{Conceptualising Contestability: Perspectives on
  Contesting Algorithmic Decisions}}.
\newblock  (\bibinfo{date}{2} \bibinfo{year}{2021}).
\newblock
\urldef\tempurl%
\url{https://doi.org/10.1145/3449180}
\showDOI{\tempurl}


\bibitem[\protect\citeauthoryear{Lyu, Huang, and Liang}{Lyu
  et~al\mbox{.}}{2015}]%
        {lyu2015}
\bibfield{author}{\bibinfo{person}{Chunchuan Lyu}, \bibinfo{person}{Kaizhu
  Huang}, {and} \bibinfo{person}{Hai-Ning Liang}.}
  \bibinfo{year}{2015}\natexlab{}.
\newblock \showarticletitle{{A Unified Gradient Regularization Family for
  Adversarial Examples}}. In \bibinfo{booktitle}{\emph{2015 IEEE International
  Conference on Data Mining}}. \bibinfo{publisher}{IEEE},
  \bibinfo{pages}{301--309}.
\newblock
\showISBNx{978-1-4673-9504-5}
\urldef\tempurl%
\url{https://doi.org/10.1109/ICDM.2015.84}
\showDOI{\tempurl}


\bibitem[\protect\citeauthoryear{Mahendran}{Mahendran}{2021}]%
        {mahendran2021}
\bibfield{author}{\bibinfo{person}{N. Mahendran}.}
  \bibinfo{year}{2021}\natexlab{}.
\newblock \showarticletitle{{Analysis of memory consumption by neural networks
  based on hyperparameters}}.
\newblock  (\bibinfo{date}{10} \bibinfo{year}{2021}).
\newblock


\bibitem[\protect\citeauthoryear{Martin, Google Vinodkumar Prabhakaran Google
  Jill~Kuhlberg, and Smart Google William Isaac~DeepMind}{Martin
  et~al\mbox{.}}{2020}]%
        {martinJr2020}
\bibfield{author}{\bibinfo{person}{Donald Martin}, \bibinfo{person}{Jr Google
  Vinodkumar Prabhakaran Google Jill~Kuhlberg}, {and} \bibinfo{person}{Andrew~S
  Smart Google William Isaac~DeepMind}.} \bibinfo{year}{2020}\natexlab{}.
\newblock \showarticletitle{{Extending the Machine Learning Abstraction
  Boundary: A Complex Systems Approach to Incorporate Societal Context}}.
\newblock  (\bibinfo{year}{2020}).
\newblock


\bibitem[\protect\citeauthoryear{Mehldau}{Mehldau}{2007}]%
        {mehldau2007}
\bibfield{author}{\bibinfo{person}{M. Mehldau}.}
  \bibinfo{year}{2007}\natexlab{}.
\newblock \bibinfo{title}{{Iconset for data-privacy declarations v 0.1}}.
\newblock
\newblock
\urldef\tempurl%
\url{https://netzpolitik.org/wp-upload/data-privacy-icons-v01.pdf}
\showURL{%
\tempurl}


\bibitem[\protect\citeauthoryear{Mehrabi, Morstatter, Saxena, Lerman, and
  Galstyan}{Mehrabi et~al\mbox{.}}{2021}]%
        {mehrabi2021}
\bibfield{author}{\bibinfo{person}{Ninareh Mehrabi}, \bibinfo{person}{Fred
  Morstatter}, \bibinfo{person}{Nripsuta Saxena}, \bibinfo{person}{Kristina
  Lerman}, {and} \bibinfo{person}{Aram Galstyan}.}
  \bibinfo{year}{2021}\natexlab{}.
\newblock \showarticletitle{{A Survey on Bias and Fairness in Machine
  Learning}}.
\newblock \bibinfo{journal}{\emph{ACM Comput. Surv.}} \bibinfo{volume}{54},
  \bibinfo{number}{6} (\bibinfo{date}{7} \bibinfo{year}{2021}).
\newblock
\showISSN{0360-0300}
\urldef\tempurl%
\url{https://doi.org/10.1145/3457607}
\showDOI{\tempurl}


\bibitem[\protect\citeauthoryear{{Microsoft}}{{Microsoft}}{2018}]%
        {microsoftai2018}
\bibfield{author}{\bibinfo{person}{{Microsoft}}.}
  \bibinfo{year}{2018}\natexlab{}.
\newblock \bibinfo{title}{{AI Principles}}.
\newblock
\newblock
\urldef\tempurl%
\url{https://www.microsoft.com/en-us/ai/responsible-ai?activetab=pivot1%3aprimaryr6}
\showURL{%
\tempurl}


\bibitem[\protect\citeauthoryear{Mishra and Rzeszotarski}{Mishra and
  Rzeszotarski}{2021}]%
        {mishra2021}
\bibfield{author}{\bibinfo{person}{Swati Mishra} {and}
  \bibinfo{person}{Jeffrey~M Rzeszotarski}.} \bibinfo{year}{2021}\natexlab{}.
\newblock \showarticletitle{{Designing Interactive Transfer Learning Tools for
  ML Non-Experts}}. In \bibinfo{booktitle}{\emph{Proceedings of the 2021 CHI
  Conference on Human Factors in Computing Systems}}. \bibinfo{publisher}{ACM},
  \bibinfo{address}{New York, NY, USA}.
\newblock
\showISBNx{9781450380966}
\urldef\tempurl%
\url{https://doi.org/10.1145/3411764.3445096}
\showDOI{\tempurl}


\bibitem[\protect\citeauthoryear{{Mission assigned by the French Prime
  Minister}}{{Mission assigned by the French Prime Minister}}{2019}]%
        {frenchminister2019}
\bibfield{author}{\bibinfo{person}{{Mission assigned by the French Prime
  Minister}}.} \bibinfo{year}{2019}\natexlab{}.
\newblock \bibinfo{title}{{For a Meaningful Artificial Intelligence: Toward a
  French and European Strategy}}.
\newblock
\newblock
\urldef\tempurl%
\url{https://www.aiforhumanity.fr/pdfs/MissionVillani_Report_ENG-VF.pdf}
\showURL{%
\tempurl}


\bibitem[\protect\citeauthoryear{Mitchell, Wu, Zaldivar, Barnes, Vasserman,
  Hutchinson, Spitzer, Raji, and Gebru}{Mitchell et~al\mbox{.}}{2018}]%
        {mitchell2019}
\bibfield{author}{\bibinfo{person}{Margaret Mitchell}, \bibinfo{person}{Simone
  Wu}, \bibinfo{person}{Andrew Zaldivar}, \bibinfo{person}{Parker Barnes},
  \bibinfo{person}{Lucy Vasserman}, \bibinfo{person}{Ben Hutchinson},
  \bibinfo{person}{Elena Spitzer}, \bibinfo{person}{Inioluwa~Deborah Raji},
  {and} \bibinfo{person}{Timnit Gebru}.} \bibinfo{year}{2018}\natexlab{}.
\newblock \showarticletitle{{Model Cards for Model Reporting}}.
\newblock  (\bibinfo{date}{10} \bibinfo{year}{2018}).
\newblock
\urldef\tempurl%
\url{https://doi.org/10.1145/3287560.3287596}
\showDOI{\tempurl}


\bibitem[\protect\citeauthoryear{Mitra}{Mitra}{2021}]%
        {mitra2021}
\bibfield{author}{\bibinfo{person}{Tanushree Mitra}.}
  \bibinfo{year}{2021}\natexlab{}.
\newblock \showarticletitle{{Provocation: Contestability in Large-Scale
  Interactive {\{}NLP{\}} Systems}}. In \bibinfo{booktitle}{\emph{Proceedings
  of the First Workshop on Bridging Human{\{}--{\}}Computer Interaction and
  Natural Language Processing}}. \bibinfo{publisher}{Association for
  Computational Linguistics}, \bibinfo{pages}{96--100}.
\newblock


\bibitem[\protect\citeauthoryear{Mittelstadt}{Mittelstadt}{2019}]%
        {mittelstadt2019}
\bibfield{author}{\bibinfo{person}{Brent Mittelstadt}.}
  \bibinfo{year}{2019}\natexlab{}.
\newblock \showarticletitle{{Principles alone cannot guarantee ethical AI}}.
\newblock \bibinfo{journal}{\emph{Nature Machine Intelligence}}
  \bibinfo{volume}{1}, \bibinfo{number}{11} (\bibinfo{year}{2019}),
  \bibinfo{pages}{501--507}.
\newblock
\showISSN{2522-5839}
\urldef\tempurl%
\url{https://doi.org/10.1038/s42256-019-0114-4}
\showDOI{\tempurl}


\bibitem[\protect\citeauthoryear{Morley, Floridi, Kinsey, and Elhalal}{Morley
  et~al\mbox{.}}{2020}]%
        {morley2020}
\bibfield{author}{\bibinfo{person}{Jessica Morley}, \bibinfo{person}{Luciano
  Floridi}, \bibinfo{person}{Libby Kinsey}, {and} \bibinfo{person}{Anat
  Elhalal}.} \bibinfo{year}{2020}\natexlab{}.
\newblock \showarticletitle{{From What to How: An Initial Review of Publicly
  Available AI Ethics Tools, Methods and Research to Translate Principles into
  Practices}}.
\newblock \bibinfo{journal}{\emph{Science and Engineering Ethics}}
  \bibinfo{volume}{26} (\bibinfo{year}{2020}), \bibinfo{pages}{2141--2168}.
\newblock
\showISBNx{0123456789}
\urldef\tempurl%
\url{https://doi.org/10.1007/s11948-019-00165-5}
\showDOI{\tempurl}


\bibitem[\protect\citeauthoryear{Mothilal, Sharma, and Tan}{Mothilal
  et~al\mbox{.}}{2019}]%
        {mothilal2019}
\bibfield{author}{\bibinfo{person}{Ramaravind~Kommiya Mothilal},
  \bibinfo{person}{Amit Sharma}, {and} \bibinfo{person}{Chenhao Tan}.}
  \bibinfo{year}{2019}\natexlab{}.
\newblock \showarticletitle{{Explaining Machine Learning Classifiers through
  Diverse Counterfactual Explanations}}.
\newblock  (\bibinfo{date}{5} \bibinfo{year}{2019}).
\newblock
\urldef\tempurl%
\url{https://doi.org/10.1145/3351095.3372850}
\showDOI{\tempurl}


\bibitem[\protect\citeauthoryear{Murukannaiah and Singh}{Murukannaiah and
  Singh}{2014}]%
        {Murukannaiah2014}
\bibfield{author}{\bibinfo{person}{Pradeep~K Murukannaiah} {and}
  \bibinfo{person}{Munindar~P Singh}.} \bibinfo{year}{2014}\natexlab{}.
\newblock \showarticletitle{{Xipho: Extending Tropos to Engineer Context-Aware
  Personal Agents}}. In \bibinfo{booktitle}{\emph{Proceedings of the 2014
  International Conference on Autonomous Agents and Multi-Agent Systems}}
  \emph{(\bibinfo{series}{AAMAS '14})}. \bibinfo{publisher}{International
  Foundation for Autonomous Agents and Multiagent Systems},
  \bibinfo{address}{Richland, SC}, \bibinfo{pages}{309--316}.
\newblock
\showISBNx{9781450327381}


\bibitem[\protect\citeauthoryear{Nasr, Shokri, and Houmansadr}{Nasr
  et~al\mbox{.}}{2018}]%
        {nasr2018}
\bibfield{author}{\bibinfo{person}{Milad Nasr}, \bibinfo{person}{Reza Shokri},
  {and} \bibinfo{person}{Amir Houmansadr}.} \bibinfo{year}{2018}\natexlab{}.
\newblock \showarticletitle{{Machine Learning with Membership Privacy using
  Adversarial Regularization}}.
\newblock  (\bibinfo{date}{7} \bibinfo{year}{2018}).
\newblock


\bibitem[\protect\citeauthoryear{Nori, Jenkins, Koch, and Caruana}{Nori
  et~al\mbox{.}}{2019}]%
        {nori2019}
\bibfield{author}{\bibinfo{person}{Harsha Nori}, \bibinfo{person}{Samuel
  Jenkins}, \bibinfo{person}{Paul Koch}, {and} \bibinfo{person}{Rich Caruana}.}
  \bibinfo{year}{2019}\natexlab{}.
\newblock \showarticletitle{{InterpretML: A Unified Framework for Machine
  Learning Interpretability}}.
\newblock  (\bibinfo{date}{9} \bibinfo{year}{2019}).
\newblock


\bibitem[\protect\citeauthoryear{Nurwidyantoro, Shahin, Chaudron, Hussain,
  Perera, Shams, and Whittle}{Nurwidyantoro et~al\mbox{.}}{2021}]%
        {nurwidyantoro2021}
\bibfield{author}{\bibinfo{person}{Arif Nurwidyantoro},
  \bibinfo{person}{Mojtaba Shahin}, \bibinfo{person}{Michel Chaudron},
  \bibinfo{person}{Waqar Hussain}, \bibinfo{person}{Harsha Perera},
  \bibinfo{person}{Rifat~Ara Shams}, {and} \bibinfo{person}{Jon Whittle}.}
  \bibinfo{year}{2021}\natexlab{}.
\newblock \showarticletitle{{Towards a Human Values Dashboard for Software
  Development: An Exploratory Study}}.
\newblock  (\bibinfo{date}{7} \bibinfo{year}{2021}).
\newblock


\bibitem[\protect\citeauthoryear{{OECD}}{{OECD}}{2019}]%
        {OECD2019}
\bibfield{author}{\bibinfo{person}{{OECD}}.} \bibinfo{year}{2019}\natexlab{}.
\newblock \bibinfo{title}{{Recommendation of the Council on Artificial
  Intelligence}}.
\newblock
\newblock
\urldef\tempurl%
\url{https://legalinstruments.oecd.org/en/instruments/OECD-LEGAL-0406}
\showURL{%
\tempurl}


\bibitem[\protect\citeauthoryear{O'Hara}{O'Hara}{2020}]%
        {ohara2020}
\bibfield{author}{\bibinfo{person}{Kieron O'Hara}.}
  \bibinfo{year}{2020}\natexlab{}.
\newblock \showarticletitle{{Explainable AI and the philosophy and practice of
  explanation}}.
\newblock \bibinfo{journal}{\emph{Computer Law {\&} Security Review}}
  \bibinfo{volume}{39} (\bibinfo{date}{11} \bibinfo{year}{2020}),
  \bibinfo{pages}{105474}.
\newblock
\showISSN{0267-3649}
\urldef\tempurl%
\url{https://doi.org/10.1016/J.CLSR.2020.105474}
\showDOI{\tempurl}


\bibitem[\protect\citeauthoryear{Papernot, McDaniel, Wu, Jha, and
  Swami}{Papernot et~al\mbox{.}}{2016}]%
        {papernot2016}
\bibfield{author}{\bibinfo{person}{Nicolas Papernot}, \bibinfo{person}{Patrick
  McDaniel}, \bibinfo{person}{Xi Wu}, \bibinfo{person}{Somesh Jha}, {and}
  \bibinfo{person}{Ananthram Swami}.} \bibinfo{year}{2016}\natexlab{}.
\newblock \showarticletitle{{Distillation as a Defense to Adversarial
  Perturbations Against Deep Neural Networks}}. In
  \bibinfo{booktitle}{\emph{2016 IEEE Symposium on Security and Privacy (SP)}}.
  \bibinfo{publisher}{IEEE}, \bibinfo{pages}{582--597}.
\newblock
\showISBNx{978-1-5090-0824-7}
\urldef\tempurl%
\url{https://doi.org/10.1109/SP.2016.41}
\showDOI{\tempurl}


\bibitem[\protect\citeauthoryear{Parisi and Comunello}{Parisi and
  Comunello}{2020}]%
        {parisi2020}
\bibfield{author}{\bibinfo{person}{Lorenza Parisi} {and}
  \bibinfo{person}{Francesca Comunello}.} \bibinfo{year}{2020}\natexlab{}.
\newblock \showarticletitle{{Dating in the time of “relational filter
  bubbles”: exploring imaginaries, perceptions and tactics of Italian dating
  app users}}.
\newblock \bibinfo{journal}{\emph{The Communication Review}}
  \bibinfo{volume}{23}, \bibinfo{number}{1} (\bibinfo{year}{2020}),
  \bibinfo{pages}{66--89}.
\newblock
\urldef\tempurl%
\url{https://doi.org/10.1080/10714421.2019.1704111}
\showDOI{\tempurl}


\bibitem[\protect\citeauthoryear{Patel}{Patel}{2021}]%
        {patel2021}
\bibfield{author}{\bibinfo{person}{Reema Patel}.}
  \bibinfo{year}{2021}\natexlab{}.
\newblock \showarticletitle{{Reboot AI with human values}}.
\newblock \bibinfo{journal}{\emph{Nature}} \bibinfo{volume}{598},
  \bibinfo{number}{7879} (\bibinfo{date}{10} \bibinfo{year}{2021}).
\newblock
\showISSN{0028-0836}
\urldef\tempurl%
\url{https://doi.org/10.1038/d41586-021-02693-2}
\showDOI{\tempurl}


\bibitem[\protect\citeauthoryear{Paullada, Raji, Bender, Denton, and
  Hanna}{Paullada et~al\mbox{.}}{2020}]%
        {paullada2020}
\bibfield{author}{\bibinfo{person}{Amandalynne Paullada},
  \bibinfo{person}{Inioluwa~Deborah Raji}, \bibinfo{person}{Emily~M. Bender},
  \bibinfo{person}{Emily Denton}, {and} \bibinfo{person}{Alex Hanna}.}
  \bibinfo{year}{2020}\natexlab{}.
\newblock \showarticletitle{{Data and its (dis)contents: A survey of dataset
  development and use in machine learning research}}.
\newblock  (\bibinfo{date}{12} \bibinfo{year}{2020}).
\newblock
\urldef\tempurl%
\url{http://arxiv.org/abs/2012.05345}
\showURL{%
\tempurl}


\bibitem[\protect\citeauthoryear{Pommeranz, Detweiler, Wiggers, and
  Jonker}{Pommeranz et~al\mbox{.}}{2012}]%
        {pommeranz2012}
\bibfield{author}{\bibinfo{person}{Alina Pommeranz}, \bibinfo{person}{Christian
  Detweiler}, \bibinfo{person}{Pascal Wiggers}, {and}
  \bibinfo{person}{Catholijn Jonker}.} \bibinfo{year}{2012}\natexlab{}.
\newblock \showarticletitle{{Elicitation of situated values: need for tools to
  help stakeholders and designers to reflect and communicate}}.
\newblock \bibinfo{journal}{\emph{Ethics and Information Technology}}
  \bibinfo{volume}{14}, \bibinfo{number}{4} (\bibinfo{date}{12}
  \bibinfo{year}{2012}), \bibinfo{pages}{285--303}.
\newblock
\showISSN{1388-1957}
\urldef\tempurl%
\url{https://doi.org/10.1007/s10676-011-9282-6}
\showDOI{\tempurl}


\bibitem[\protect\citeauthoryear{Raghavan, Barocas, Kleinberg, and
  Levy}{Raghavan et~al\mbox{.}}{2020}]%
        {raghavan2020}
\bibfield{author}{\bibinfo{person}{Manish Raghavan}, \bibinfo{person}{Solon
  Barocas}, \bibinfo{person}{Jon Kleinberg}, {and} \bibinfo{person}{Karen
  Levy}.} \bibinfo{year}{2020}\natexlab{}.
\newblock \showarticletitle{{Mitigating Bias in Algorithmic Hiring: Evaluating
  Claims and Practices}}. In \bibinfo{booktitle}{\emph{Proceedings of the 2020
  Conference on Fairness, Accountability, and Transparency}}
  \emph{(\bibinfo{series}{FAT* '20})}. \bibinfo{publisher}{Association for
  Computing Machinery}, \bibinfo{address}{New York, NY, USA},
  \bibinfo{pages}{469--481}.
\newblock
\showISBNx{9781450369367}
\urldef\tempurl%
\url{https://doi.org/10.1145/3351095.3372828}
\showDOI{\tempurl}


\bibitem[\protect\citeauthoryear{Raji, Smart, White, Mitchell, Gebru,
  Hutchinson, Smith-Loud, Theron, and Barnes}{Raji et~al\mbox{.}}{2020}]%
        {raji2020}
\bibfield{author}{\bibinfo{person}{Inioluwa~Deborah Raji},
  \bibinfo{person}{Andrew Smart}, \bibinfo{person}{Rebecca~N. White},
  \bibinfo{person}{Margaret Mitchell}, \bibinfo{person}{Timnit Gebru},
  \bibinfo{person}{Ben Hutchinson}, \bibinfo{person}{Jamila Smith-Loud},
  \bibinfo{person}{Daniel Theron}, {and} \bibinfo{person}{Parker Barnes}.}
  \bibinfo{year}{2020}\natexlab{}.
\newblock \showarticletitle{{Closing the AI accountability gap}}. In
  \bibinfo{booktitle}{\emph{Proceedings of the 2020 Conference on Fairness,
  Accountability, and Transparency}}. \bibinfo{publisher}{ACM},
  \bibinfo{address}{New York, NY, USA}, \bibinfo{pages}{33--44}.
\newblock
\showISBNx{9781450369367}
\urldef\tempurl%
\url{https://doi.org/10.1145/3351095.3372873}
\showDOI{\tempurl}


\bibitem[\protect\citeauthoryear{Ribeiro, Ottoni, West, Almeida, and
  Meira}{Ribeiro et~al\mbox{.}}{2020}]%
        {hortaribeiro2020}
\bibfield{author}{\bibinfo{person}{Manoel~Horta Ribeiro},
  \bibinfo{person}{Raphael Ottoni}, \bibinfo{person}{Robert West},
  \bibinfo{person}{Virg\'{\i}lio A~F Almeida}, {and} \bibinfo{person}{Wagner
  Meira}.} \bibinfo{year}{2020}\natexlab{}.
\newblock \showarticletitle{{Auditing Radicalization Pathways on YouTube}}. In
  \bibinfo{booktitle}{\emph{Proceedings of the 2020 Conference on Fairness,
  Accountability, and Transparency}} \emph{(\bibinfo{series}{FAT* '20})}.
  \bibinfo{publisher}{Association for Computing Machinery},
  \bibinfo{address}{New York, NY, USA}, \bibinfo{pages}{131--141}.
\newblock
\showISBNx{9781450369367}
\urldef\tempurl%
\url{https://doi.org/10.1145/3351095.3372879}
\showDOI{\tempurl}


\bibitem[\protect\citeauthoryear{Rossi and Palmirani}{Rossi and
  Palmirani}{2017}]%
        {rossi2017}
\bibfield{author}{\bibinfo{person}{Arianna Rossi} {and} \bibinfo{person}{Monica
  Palmirani}.} \bibinfo{year}{2017}\natexlab{}.
\newblock \showarticletitle{{A Visualization Approach for Adaptive Consent in
  the European Data Protection Framework}}. In \bibinfo{booktitle}{\emph{2017
  Conference for E-Democracy and Open Government (CeDEM)}}.
  \bibinfo{publisher}{IEEE}, \bibinfo{pages}{159--170}.
\newblock
\showISBNx{978-1-5090-6718-3}
\urldef\tempurl%
\url{https://doi.org/10.1109/CeDEM.2017.23}
\showDOI{\tempurl}


\bibitem[\protect\citeauthoryear{Russell, Dewey, and Tegmark}{Russell
  et~al\mbox{.}}{2015}]%
        {russel2015}
\bibfield{author}{\bibinfo{person}{Stuart Russell}, \bibinfo{person}{Daniel
  Dewey}, {and} \bibinfo{person}{Max Tegmark}.}
  \bibinfo{year}{2015}\natexlab{}.
\newblock \showarticletitle{{Research Priorities for Robust and Beneficial
  Artificial Intelligence}}.
\newblock \bibinfo{journal}{\emph{AI Magazine}} \bibinfo{volume}{36},
  \bibinfo{number}{4} (\bibinfo{date}{12} \bibinfo{year}{2015}).
\newblock
\showISSN{2371-9621}
\urldef\tempurl%
\url{https://doi.org/10.1609/aimag.v36i4.2577}
\showDOI{\tempurl}


\bibitem[\protect\citeauthoryear{Saleiro, Kuester, Hinkson, London, Stevens,
  Anisfeld, Rodolfa, and Ghani}{Saleiro et~al\mbox{.}}{2018}]%
        {saleiro2018}
\bibfield{author}{\bibinfo{person}{Pedro Saleiro}, \bibinfo{person}{Benedict
  Kuester}, \bibinfo{person}{Loren Hinkson}, \bibinfo{person}{Jesse London},
  \bibinfo{person}{Abby Stevens}, \bibinfo{person}{Ari Anisfeld},
  \bibinfo{person}{Kit~T Rodolfa}, {and} \bibinfo{person}{Rayid Ghani}.}
  \bibinfo{year}{2018}\natexlab{}.
\newblock \showarticletitle{{Aequitas: A Bias and Fairness Audit Toolkit}}.
\newblock


\bibitem[\protect\citeauthoryear{Sandvig, Hamilton, Karahalios, and
  Langbort}{Sandvig et~al\mbox{.}}{2014}]%
        {sandvig2014}
\bibfield{author}{\bibinfo{person}{Christian Sandvig}, \bibinfo{person}{Kevin
  Hamilton}, \bibinfo{person}{Karrie Karahalios}, {and} \bibinfo{person}{Cedric
  Langbort}.} \bibinfo{year}{2014}\natexlab{}.
\newblock \showarticletitle{{Auditing algorithms: Research methods for
  detecting discrimination on internet platforms.}}. In
  \bibinfo{booktitle}{\emph{Data and discrimination: converting critical
  concerns into productive inquiry 22}}.
\newblock


\bibitem[\protect\citeauthoryear{Sap, Card, Gabriel, Choi, and Smith}{Sap
  et~al\mbox{.}}{2019}]%
        {sap2019}
\bibfield{author}{\bibinfo{person}{Maarten Sap}, \bibinfo{person}{Dallas Card},
  \bibinfo{person}{Saadia Gabriel}, \bibinfo{person}{Yejin Choi}, {and}
  \bibinfo{person}{Noah~A. Smith}.} \bibinfo{year}{2019}\natexlab{}.
\newblock \showarticletitle{{The Risk of Racial Bias in Hate Speech
  Detection}}.
\newblock \bibinfo{journal}{\emph{ACL 2019 - 57th Annual Meeting of the
  Association for Computational Linguistics, Proceedings of the Conference}}
  (\bibinfo{year}{2019}), \bibinfo{pages}{1668--1678}.
\newblock
\urldef\tempurl%
\url{https://doi.org/10.18653/V1/P19-1163}
\showDOI{\tempurl}


\bibitem[\protect\citeauthoryear{Schaerer, Kelley, and Nicolescu}{Schaerer
  et~al\mbox{.}}{2009}]%
        {schaerer2009}
\bibfield{author}{\bibinfo{person}{Enrique Schaerer}, \bibinfo{person}{Richard
  Kelley}, {and} \bibinfo{person}{Monica Nicolescu}.}
  \bibinfo{year}{2009}\natexlab{}.
\newblock \showarticletitle{{Robots as animals: A framework for liability and
  responsibility in human-robot interactions}}. In
  \bibinfo{booktitle}{\emph{RO-MAN 2009 - The 18th IEEE International Symposium
  on Robot and Human Interactive Communication}}. \bibinfo{publisher}{IEEE}.
\newblock
\showISBNx{978-1-4244-5081-7}
\urldef\tempurl%
\url{https://doi.org/10.1109/ROMAN.2009.5326244}
\showDOI{\tempurl}


\bibitem[\protect\citeauthoryear{Scheuerman, Denton, and Hanna}{Scheuerman
  et~al\mbox{.}}{2021}]%
        {scheuerman2021}
\bibfield{author}{\bibinfo{person}{Morgan~Klaus Scheuerman},
  \bibinfo{person}{Emily Denton}, {and} \bibinfo{person}{Alex Hanna}.}
  \bibinfo{year}{2021}\natexlab{}.
\newblock \showarticletitle{{Do Datasets Have Politics? Disciplinary Values in
  Computer Vision Dataset Development}}.
\newblock \bibinfo{journal}{\emph{Proc. ACM Hum.-Comput. Interact. 5, CSCW2,
  Article 317}} (\bibinfo{year}{2021}).
\newblock
\urldef\tempurl%
\url{https://doi.org/10.1145/3476058}
\showURL{%
\tempurl}


\bibitem[\protect\citeauthoryear{Schwartz}{Schwartz}{2012}]%
        {schwartz2012}
\bibfield{author}{\bibinfo{person}{Shalom~H. Schwartz}.}
  \bibinfo{year}{2012}\natexlab{}.
\newblock \showarticletitle{{An Overview of the Schwartz Theory of Basic
  Values}}.
\newblock \bibinfo{journal}{\emph{Online Readings in Psychology and Culture}}
  \bibinfo{volume}{2}, \bibinfo{number}{1} (\bibinfo{date}{12}
  \bibinfo{year}{2012}).
\newblock
\showISSN{2307-0919}
\urldef\tempurl%
\url{https://doi.org/10.9707/2307-0919.1116}
\showDOI{\tempurl}


\bibitem[\protect\citeauthoryear{Shahin, Hussain, Nurwidyantoro, Perera, Shams,
  Grundy, and Whittle}{Shahin et~al\mbox{.}}{2021}]%
        {shahin2021}
\bibfield{author}{\bibinfo{person}{Mojtaba Shahin}, \bibinfo{person}{Waqar
  Hussain}, \bibinfo{person}{Arif Nurwidyantoro}, \bibinfo{person}{Harsha
  Perera}, \bibinfo{person}{Rifat Shams}, \bibinfo{person}{John Grundy}, {and}
  \bibinfo{person}{Jon Whittle}.} \bibinfo{year}{2021}\natexlab{}.
\newblock \showarticletitle{{Operationalizing Human Values in Software
  Engineering: A Survey}}.
\newblock  (\bibinfo{date}{8} \bibinfo{year}{2021}).
\newblock


\bibitem[\protect\citeauthoryear{Shen, DeVos, Eslami, and Holstein}{Shen
  et~al\mbox{.}}{2021}]%
        {shen2021}
\bibfield{author}{\bibinfo{person}{Hong Shen}, \bibinfo{person}{Alicia DeVos},
  \bibinfo{person}{Motahhare Eslami}, {and} \bibinfo{person}{Kenneth
  Holstein}.} \bibinfo{year}{2021}\natexlab{}.
\newblock \showarticletitle{{Everyday algorithm auditing: Understanding the
  power of everyday users in surfacing harmful algorithmic behaviors}}.
\newblock  (\bibinfo{date}{5} \bibinfo{year}{2021}).
\newblock
\urldef\tempurl%
\url{https://doi.org/10.1145/3479577}
\showDOI{\tempurl}


\bibitem[\protect\citeauthoryear{Shklovski and N{\'{e}}methy}{Shklovski and
  N{\'{e}}methy}{2022}]%
        {Shklovski2022}
\bibfield{author}{\bibinfo{person}{Irina Shklovski} {and}
  \bibinfo{person}{Carolina N{\'{e}}methy}.} \bibinfo{year}{2022}\natexlab{}.
\newblock \showarticletitle{{Nodes of certainty and spaces for doubt in AI
  ethics for engineers}}.
\newblock \bibinfo{journal}{\emph{Information, Communication {\&} Society}}
  (\bibinfo{date}{1} \bibinfo{year}{2022}), \bibinfo{pages}{1--17}.
\newblock
\showISSN{1369-118X}
\urldef\tempurl%
\url{https://doi.org/10.1080/1369118X.2021.2014547}
\showDOI{\tempurl}


\bibitem[\protect\citeauthoryear{Shneiderman}{Shneiderman}{2020}]%
        {schneiderman2020}
\bibfield{author}{\bibinfo{person}{Ben Shneiderman}.}
  \bibinfo{year}{2020}\natexlab{}.
\newblock \showarticletitle{{Bridging the Gap Between Ethics and Practice:
  Guidelines for Reliable, Safe, and Trustworthy Human-Centered AI Systems}}.
\newblock \bibinfo{journal}{\emph{ACM Trans. Interact. Intell. Syst.}}
  \bibinfo{volume}{10}, \bibinfo{number}{4} (\bibinfo{date}{10}
  \bibinfo{year}{2020}).
\newblock
\showISSN{2160-6455}
\urldef\tempurl%
\url{https://doi.org/10.1145/3419764}
\showDOI{\tempurl}


\bibitem[\protect\citeauthoryear{Shokri, Stronati, Song, and Shmatikov}{Shokri
  et~al\mbox{.}}{2016}]%
        {shokri2017}
\bibfield{author}{\bibinfo{person}{Reza Shokri}, \bibinfo{person}{Marco
  Stronati}, \bibinfo{person}{Congzheng Song}, {and} \bibinfo{person}{Vitaly
  Shmatikov}.} \bibinfo{year}{2016}\natexlab{}.
\newblock \showarticletitle{{Membership Inference Attacks against Machine
  Learning Models}}.
\newblock  (\bibinfo{date}{10} \bibinfo{year}{2016}).
\newblock


\bibitem[\protect\citeauthoryear{Sokol and Flach}{Sokol and Flach}{2020}]%
        {sokol2020}
\bibfield{author}{\bibinfo{person}{Kacper Sokol} {and} \bibinfo{person}{Peter
  Flach}.} \bibinfo{year}{2020}\natexlab{}.
\newblock \showarticletitle{{Explainability fact sheets}}. In
  \bibinfo{booktitle}{\emph{Proceedings of the 2020 Conference on Fairness,
  Accountability, and Transparency}}. \bibinfo{publisher}{ACM},
  \bibinfo{address}{New York, NY, USA}.
\newblock
\showISBNx{9781450369367}
\urldef\tempurl%
\url{https://doi.org/10.1145/3351095.3372870}
\showDOI{\tempurl}


\bibitem[\protect\citeauthoryear{Srivastava, Heidari, and Krause}{Srivastava
  et~al\mbox{.}}{2019}]%
        {srivastava2019}
\bibfield{author}{\bibinfo{person}{Megha Srivastava}, \bibinfo{person}{Hoda
  Heidari}, {and} \bibinfo{person}{Andreas Krause}.}
  \bibinfo{year}{2019}\natexlab{}.
\newblock \showarticletitle{{Mathematical Notions vs. Human Perception of
  Fairness}}. In \bibinfo{booktitle}{\emph{Proceedings of the 25th ACM SIGKDD
  International Conference on Knowledge Discovery {\&} Data Mining}}.
  \bibinfo{publisher}{ACM}, \bibinfo{address}{New York, NY, USA},
  \bibinfo{pages}{2459--2468}.
\newblock
\showISBNx{9781450362016}
\urldef\tempurl%
\url{https://doi.org/10.1145/3292500.3330664}
\showDOI{\tempurl}


\bibitem[\protect\citeauthoryear{Suresh, Gomez, Nam, and Satyanarayan}{Suresh
  et~al\mbox{.}}{2021}]%
        {suresh2021}
\bibfield{author}{\bibinfo{person}{Harini Suresh}, \bibinfo{person}{Steven~R.
  Gomez}, \bibinfo{person}{Kevin~K. Nam}, {and} \bibinfo{person}{Arvind
  Satyanarayan}.} \bibinfo{year}{2021}\natexlab{}.
\newblock \showarticletitle{{Beyond Expertise and Roles: A Framework to
  Characterize the Stakeholders of Interpretable Machine Learning and their
  Needs}}. In \bibinfo{booktitle}{\emph{Proceedings of the 2021 CHI Conference
  on Human Factors in Computing Systems}}. \bibinfo{publisher}{ACM},
  \bibinfo{address}{New York, NY, USA}, \bibinfo{pages}{1--16}.
\newblock
\showISBNx{9781450380966}
\urldef\tempurl%
\url{https://doi.org/10.1145/3411764.3445088}
\showDOI{\tempurl}


\bibitem[\protect\citeauthoryear{Suresh and Guttag}{Suresh and Guttag}{2021}]%
        {suresh2021_2}
\bibfield{author}{\bibinfo{person}{Harini Suresh} {and} \bibinfo{person}{John
  Guttag}.} \bibinfo{year}{2021}\natexlab{}.
\newblock \showarticletitle{{A Framework for Understanding Sources of Harm
  throughout the Machine Learning Life Cycle}}. In
  \bibinfo{booktitle}{\emph{Equity and Access in Algorithms, Mechanisms, and
  Optimization}}. \bibinfo{publisher}{ACM}, \bibinfo{address}{New York, NY,
  USA}, \bibinfo{pages}{1--9}.
\newblock
\showISBNx{9781450385534}
\urldef\tempurl%
\url{https://doi.org/10.1145/3465416.3483305}
\showDOI{\tempurl}


\bibitem[\protect\citeauthoryear{{Telia Company}}{{Telia Company}}{2019}]%
        {teliaai2019}
\bibfield{author}{\bibinfo{person}{{Telia Company}}.}
  \bibinfo{year}{2019}\natexlab{}.
\newblock \bibinfo{title}{{Guiding Principles on Trusted AI Ethics}}.
\newblock
\newblock
\urldef\tempurl%
\url{https://www.teliacompany.com/globalassets/telia-company/documents/about-telia-company/public-policy/2018/guiding-principles-on-trusted-ai-ethics.pdf}
\showURL{%
\tempurl}


\bibitem[\protect\citeauthoryear{{The IEEE Global Initiative on Ethics of
  Autonomous and Intelligent Systems.}}{{The IEEE Global Initiative on Ethics
  of Autonomous and Intelligent Systems.}}{2019}]%
        {IEEE2019}
\bibfield{author}{\bibinfo{person}{{The IEEE Global Initiative on Ethics of
  Autonomous and Intelligent Systems.}}} \bibinfo{year}{2019}\natexlab{}.
\newblock \bibinfo{booktitle}{\emph{{Ethically Aligned Design: A Vision for
  Prioritizing Human Well-being with Autonomous and Intelligent Systems}}
  (\bibinfo{edition}{first edition} ed.)}.
\newblock \bibinfo{publisher}{IEEE}.
\newblock


\bibitem[\protect\citeauthoryear{{The Royal Society}}{{The Royal
  Society}}{2019}]%
        {royalsociety2019}
\bibfield{author}{\bibinfo{person}{{The Royal Society}}.}
  \bibinfo{year}{2019}\natexlab{}.
\newblock \bibinfo{title}{{Explainable AI: the basics }}.
\newblock
\newblock
\urldef\tempurl%
\url{https://royalsociety-org.tudelft.idm.oclc.org/-/media/policy/projects/explainable-ai/AI-and-interpretability-policy-briefing.pdf}
\showURL{%
\tempurl}


\bibitem[\protect\citeauthoryear{Thew and Sutcliffe}{Thew and
  Sutcliffe}{2018}]%
        {thew2017}
\bibfield{author}{\bibinfo{person}{Sarah Thew} {and} \bibinfo{person}{Alistair
  Sutcliffe}.} \bibinfo{year}{2018}\natexlab{}.
\newblock \showarticletitle{{Value-based requirements engineering: method and
  experience}}.
\newblock \bibinfo{journal}{\emph{Requirements Engineering}}
  \bibinfo{volume}{23}, \bibinfo{number}{4} (\bibinfo{date}{11}
  \bibinfo{year}{2018}).
\newblock
\showISSN{0947-3602}
\urldef\tempurl%
\url{https://doi.org/10.1007/s00766-017-0273-y}
\showDOI{\tempurl}


\bibitem[\protect\citeauthoryear{Tolan, Miron, G{\'{o}}mez, and Castillo}{Tolan
  et~al\mbox{.}}{2019}]%
        {tolan2019}
\bibfield{author}{\bibinfo{person}{Songül Tolan}, \bibinfo{person}{Marius
  Miron}, \bibinfo{person}{Emilia G{\'{o}}mez}, {and} \bibinfo{person}{Carlos
  Castillo}.} \bibinfo{year}{2019}\natexlab{}.
\newblock \showarticletitle{{Why Machine Learning May Lead to Unfairness:
  Evidence from Risk Assessment for Juvenile Justice in Catalonia}}. In
  \bibinfo{booktitle}{\emph{Proceedings of the Seventeenth International
  Conference on Artificial Intelligence and Law}} \emph{(\bibinfo{series}{ICAIL
  '19})}. \bibinfo{publisher}{Association for Computing Machinery},
  \bibinfo{address}{New York, NY, USA}, \bibinfo{pages}{83--92}.
\newblock
\showISBNx{9781450367547}
\urldef\tempurl%
\url{https://doi.org/10.1145/3322640.3326705}
\showDOI{\tempurl}


\bibitem[\protect\citeauthoryear{Tsarapatsanis and Aletras}{Tsarapatsanis and
  Aletras}{2021}]%
        {tsarapatsanis2021}
\bibfield{author}{\bibinfo{person}{Dimitrios Tsarapatsanis} {and}
  \bibinfo{person}{Nikolaos Aletras}.} \bibinfo{year}{2021}\natexlab{}.
\newblock \showarticletitle{{On the Ethical Limits of Natural Language
  Processing on Legal Text}}.
\newblock  (\bibinfo{date}{5} \bibinfo{year}{2021}).
\newblock


\bibitem[\protect\citeauthoryear{United States Executive Office of~the
  President and on~Technology}{United States Executive Office of~the President
  and on~Technology}{2016}]%
        {usai2016}
\bibfield{author}{\bibinfo{person}{National~Science United States Executive
  Office of~the President} {and} \bibinfo{person}{Technology Council~Committee
  on Technology}.} \bibinfo{year}{2016}\natexlab{}.
\newblock \bibinfo{title}{{Preparing for the Future of Artificial
  Intelligence}}.
\newblock
\newblock
\urldef\tempurl%
\url{https://obamawhitehouse.archives.gov/sites/default/files/whitehouse_files/microsites/ostp/NSTC/preparing_for_the_future_of_ai.pdf}
\showURL{%
\tempurl}


\bibitem[\protect\citeauthoryear{Ustek-Spilda, Powell, and
  Nemorin}{Ustek-Spilda et~al\mbox{.}}{2019}]%
        {ustek-spilda2019}
\bibfield{author}{\bibinfo{person}{Funda Ustek-Spilda}, \bibinfo{person}{Alison
  Powell}, {and} \bibinfo{person}{Selena Nemorin}.}
  \bibinfo{year}{2019}\natexlab{}.
\newblock \showarticletitle{{Engaging with ethics in Internet of Things:
  Imaginaries in the social milieu of technology developers}}.
\newblock \bibinfo{journal}{\emph{Big Data {\&} Society}} \bibinfo{volume}{6},
  \bibinfo{number}{2} (\bibinfo{date}{7} \bibinfo{year}{2019}),
  \bibinfo{pages}{205395171987946}.
\newblock
\showISSN{2053-9517}
\urldef\tempurl%
\url{https://doi.org/10.1177/2053951719879468}
\showDOI{\tempurl}


\bibitem[\protect\citeauthoryear{Vaccaro, Sandvig, and Karahalios}{Vaccaro
  et~al\mbox{.}}{2020}]%
        {vaccaro2020}
\bibfield{author}{\bibinfo{person}{Kristen Vaccaro}, \bibinfo{person}{Christian
  Sandvig}, {and} \bibinfo{person}{Karrie Karahalios}.}
  \bibinfo{year}{2020}\natexlab{}.
\newblock \showarticletitle{{"At the End of the Day Facebook Does What
  ItWants"}}.
\newblock \bibinfo{journal}{\emph{Proceedings of the ACM on Human-Computer
  Interaction}} \bibinfo{volume}{4}, \bibinfo{number}{CSCW2}
  (\bibinfo{date}{10} \bibinfo{year}{2020}), \bibinfo{pages}{1--22}.
\newblock
\showISSN{2573-0142}
\urldef\tempurl%
\url{https://doi.org/10.1145/3415238}
\showDOI{\tempurl}


\bibitem[\protect\citeauthoryear{van Berkel, Goncalves, Russo, Hosio, and
  Skov}{van Berkel et~al\mbox{.}}{2021}]%
        {vanBerkel2021}
\bibfield{author}{\bibinfo{person}{Niels van Berkel}, \bibinfo{person}{Jorge
  Goncalves}, \bibinfo{person}{Daniel Russo}, \bibinfo{person}{Simo Hosio},
  {and} \bibinfo{person}{Mikael~B. Skov}.} \bibinfo{year}{2021}\natexlab{}.
\newblock \showarticletitle{{Effect of Information Presentation on Fairness
  Perceptions of Machine Learning Predictors}}. In
  \bibinfo{booktitle}{\emph{Proceedings of the 2021 CHI Conference on Human
  Factors in Computing Systems}}. \bibinfo{publisher}{ACM},
  \bibinfo{address}{New York, NY, USA}, \bibinfo{pages}{1--13}.
\newblock
\showISBNx{9781450380966}
\urldef\tempurl%
\url{https://doi.org/10.1145/3411764.3445365}
\showDOI{\tempurl}


\bibitem[\protect\citeauthoryear{van~de Poel}{van~de Poel}{2013}]%
        {vandepoel2013}
\bibfield{author}{\bibinfo{person}{Ibo van~de Poel}.}
  \bibinfo{year}{2013}\natexlab{}.
\newblock \showarticletitle{{Translating Values into Design Requirements}}.
\newblock \bibinfo{pages}{253--266}.
\newblock
\urldef\tempurl%
\url{https://doi.org/10.1007/978-94-007-7762-0_20}
\showDOI{\tempurl}


\bibitem[\protect\citeauthoryear{Verma and Rubin}{Verma and Rubin}{2018}]%
        {verma2018}
\bibfield{author}{\bibinfo{person}{Sahil Verma} {and} \bibinfo{person}{Julia
  Rubin}.} \bibinfo{year}{2018}\natexlab{}.
\newblock \showarticletitle{{Fairness definitions explained}}. In
  \bibinfo{booktitle}{\emph{Proceedings of the International Workshop on
  Software Fairness}}. \bibinfo{publisher}{ACM}, \bibinfo{address}{New York,
  NY, USA}, \bibinfo{pages}{1--7}.
\newblock
\showISBNx{9781450357463}
\urldef\tempurl%
\url{https://doi.org/10.1145/3194770.3194776}
\showDOI{\tempurl}


\bibitem[\protect\citeauthoryear{Wachter and Mittelstadt}{Wachter and
  Mittelstadt}{2019}]%
        {wachter2019}
\bibfield{author}{\bibinfo{person}{Sandra Wachter} {and} \bibinfo{person}{Brent
  Mittelstadt}.} \bibinfo{year}{2019}\natexlab{}.
\newblock \showarticletitle{{ Right to Reasonable Inferences: Re-Thinking Data
  Protection Law in the Age of Big Data and AI.}}
\newblock \bibinfo{journal}{\emph{Columbia Business Law Review}}
  \bibinfo{volume}{2} (\bibinfo{year}{2019}), \bibinfo{pages}{494--620}.
\newblock


\bibitem[\protect\citeauthoryear{Wang, Ritchie, Zhou, Chevalier, and Bach}{Wang
  et~al\mbox{.}}{2021}]%
        {wang2020}
\bibfield{author}{\bibinfo{person}{Zezhong Wang}, \bibinfo{person}{Jacob
  Ritchie}, \bibinfo{person}{Jingtao Zhou}, \bibinfo{person}{Fanny Chevalier},
  {and} \bibinfo{person}{Benjamin Bach}.} \bibinfo{year}{2021}\natexlab{}.
\newblock \showarticletitle{{Data Comics for Reporting Controlled User Studies
  in Human-Computer Interaction}}.
\newblock \bibinfo{journal}{\emph{IEEE Transactions on Visualization and
  Computer Graphics}} \bibinfo{volume}{27}, \bibinfo{number}{2}
  (\bibinfo{date}{2} \bibinfo{year}{2021}), \bibinfo{pages}{967--977}.
\newblock
\showISSN{1077-2626}
\urldef\tempurl%
\url{https://doi.org/10.1109/TVCG.2020.3030433}
\showDOI{\tempurl}


\bibitem[\protect\citeauthoryear{Wexler, Pushkarna, Bolukbasi, Wattenberg,
  Viegas, and Wilson}{Wexler et~al\mbox{.}}{2019}]%
        {wexler2019}
\bibfield{author}{\bibinfo{person}{James Wexler}, \bibinfo{person}{Mahima
  Pushkarna}, \bibinfo{person}{Tolga Bolukbasi}, \bibinfo{person}{Martin
  Wattenberg}, \bibinfo{person}{Fernanda Viegas}, {and} \bibinfo{person}{Jimbo
  Wilson}.} \bibinfo{year}{2019}\natexlab{}.
\newblock \showarticletitle{{The What-If Tool: Interactive Probing of Machine
  Learning Models}}.
\newblock  (\bibinfo{date}{7} \bibinfo{year}{2019}).
\newblock
\urldef\tempurl%
\url{https://doi.org/10.1109/TVCG.2019.2934619}
\showDOI{\tempurl}


\bibitem[\protect\citeauthoryear{Wilson, Ghosh, Jiang, Mislove, Baker, Szary,
  Trindel, and Polli}{Wilson et~al\mbox{.}}{2021}]%
        {wilson2021}
\bibfield{author}{\bibinfo{person}{Christo Wilson}, \bibinfo{person}{Avijit
  Ghosh}, \bibinfo{person}{Shan Jiang}, \bibinfo{person}{Alan Mislove},
  \bibinfo{person}{Lewis Baker}, \bibinfo{person}{Janelle Szary},
  \bibinfo{person}{Kelly Trindel}, {and} \bibinfo{person}{Frida Polli}.}
  \bibinfo{year}{2021}\natexlab{}.
\newblock \showarticletitle{{Building and Auditing Fair Algorithms: A Case
  Study in Candidate Screening}}. In \bibinfo{booktitle}{\emph{Proceedings of
  the 2021 ACM Conference on Fairness, Accountability, and Transparency}}
  \emph{(\bibinfo{series}{FAccT '21})}. \bibinfo{publisher}{Association for
  Computing Machinery}, \bibinfo{address}{New York, NY, USA},
  \bibinfo{pages}{666--677}.
\newblock
\showISBNx{9781450383097}
\urldef\tempurl%
\url{https://doi.org/10.1145/3442188.3445928}
\showDOI{\tempurl}


\bibitem[\protect\citeauthoryear{Winner}{Winner}{1980}]%
        {winner1980}
\bibfield{author}{\bibinfo{person}{Langdon Winner}.}
  \bibinfo{year}{1980}\natexlab{}.
\newblock \showarticletitle{{Do Artifacts Have Politics?}}
\newblock \bibinfo{journal}{\emph{Daedalus}} \bibinfo{volume}{109},
  \bibinfo{number}{1} (\bibinfo{year}{1980}), \bibinfo{pages}{121--136}.
\newblock
\showISSN{00115266}
\urldef\tempurl%
\url{http://www.jstor.org/stable/20024652}
\showURL{%
\tempurl}


\bibitem[\protect\citeauthoryear{Xiong, Buffett, Iqbal, Lamontagne, Mamun, and
  Molyneaux}{Xiong et~al\mbox{.}}{2021}]%
        {xiong2021}
\bibfield{author}{\bibinfo{person}{Pulei Xiong}, \bibinfo{person}{Scott
  Buffett}, \bibinfo{person}{Shahrear Iqbal}, \bibinfo{person}{Philippe
  Lamontagne}, \bibinfo{person}{Mohammad Mamun}, {and} \bibinfo{person}{Heather
  Molyneaux}.} \bibinfo{year}{2021}\natexlab{}.
\newblock \showarticletitle{{Towards a Robust and Trustworthy Machine Learning
  System Development}}.
\newblock  (\bibinfo{date}{1} \bibinfo{year}{2021}).
\newblock


\bibitem[\protect\citeauthoryear{Xu, Yuan, Zhang, and Wu}{Xu
  et~al\mbox{.}}{2018}]%
        {xu2018}
\bibfield{author}{\bibinfo{person}{Depeng Xu}, \bibinfo{person}{Shuhan Yuan},
  \bibinfo{person}{Lu Zhang}, {and} \bibinfo{person}{Xintao Wu}.}
  \bibinfo{year}{2018}\natexlab{}.
\newblock \showarticletitle{{FairGAN: Fairness-aware Generative Adversarial
  Networks}}. In \bibinfo{booktitle}{\emph{2018 IEEE International Conference
  on Big Data (Big Data)}}. \bibinfo{pages}{570--575}.
\newblock
\urldef\tempurl%
\url{https://doi.org/10.1109/BigData.2018.8622525}
\showDOI{\tempurl}


\bibitem[\protect\citeauthoryear{Yan and Howe}{Yan and Howe}{2020}]%
        {yan2020}
\bibfield{author}{\bibinfo{person}{An Yan} {and} \bibinfo{person}{Bill Howe}.}
  \bibinfo{year}{2020}\natexlab{}.
\newblock \showarticletitle{{Fairness-Aware Demand Prediction for New
  Mobility}}.
\newblock \bibinfo{journal}{\emph{Proceedings of the AAAI Conference on
  Artificial Intelligence}} \bibinfo{volume}{34}, \bibinfo{number}{01}
  (\bibinfo{date}{4} \bibinfo{year}{2020}), \bibinfo{pages}{1079--1087}.
\newblock
\urldef\tempurl%
\url{https://doi.org/10.1609/aaai.v34i01.5458}
\showDOI{\tempurl}


\bibitem[\protect\citeauthoryear{Yazdanpanah, Gerding, Stein, Dastani, Jonker,
  and Norman}{Yazdanpanah et~al\mbox{.}}{2021}]%
        {yazdanpanah2021}
\bibfield{author}{\bibinfo{person}{Vahid Yazdanpanah}, \bibinfo{person}{Enrico
  Gerding}, \bibinfo{person}{Sebastian Stein}, \bibinfo{person}{Mehdi Dastani},
  \bibinfo{person}{Catholijn~M Jonker}, {and} \bibinfo{person}{Timothy
  Norman}.} \bibinfo{year}{2021}\natexlab{}.
\newblock \showarticletitle{{Responsibility Research for Trustworthy Autonomous
  Systems}}. In \bibinfo{booktitle}{\emph{20th International Conference on
  Autonomous Agents and Multiagent Systems (03/05/21 - 07/05/21)}}.
  \bibinfo{pages}{57--62}.
\newblock
\urldef\tempurl%
\url{https://eprints.soton.ac.uk/447511/}
\showURL{%
\tempurl}


\bibitem[\protect\citeauthoryear{Ye, Maddi, Murakonda, and Shokri}{Ye
  et~al\mbox{.}}{2021}]%
        {ye2021}
\bibfield{author}{\bibinfo{person}{Jiayuan Ye}, \bibinfo{person}{Aadyaa Maddi},
  \bibinfo{person}{Sasi~Kumar Murakonda}, {and} \bibinfo{person}{Reza Shokri}.}
  \bibinfo{year}{2021}\natexlab{}.
\newblock \showarticletitle{{Enhanced Membership Inference Attacks against
  Machine Learning Models}}.
\newblock  (\bibinfo{date}{11} \bibinfo{year}{2021}).
\newblock


\bibitem[\protect\citeauthoryear{Zhang, Lemoine, and Mitchell}{Zhang
  et~al\mbox{.}}{2018}]%
        {zhang2018}
\bibfield{author}{\bibinfo{person}{Brian~Hu Zhang}, \bibinfo{person}{Blake
  Lemoine}, {and} \bibinfo{person}{Margaret Mitchell}.}
  \bibinfo{year}{2018}\natexlab{}.
\newblock \showarticletitle{{Mitigating Unwanted Biases with Adversarial
  Learning}}. In \bibinfo{booktitle}{\emph{Proceedings of the 2018 AAAI/ACM
  Conference on AI, Ethics, and Society}} \emph{(\bibinfo{series}{AIES '18})}.
  \bibinfo{publisher}{Association for Computing Machinery},
  \bibinfo{address}{New York, NY, USA}, \bibinfo{pages}{335--340}.
\newblock
\showISBNx{9781450360128}
\urldef\tempurl%
\url{https://doi.org/10.1145/3278721.3278779}
\showDOI{\tempurl}


\bibitem[\protect\citeauthoryear{Zhou, Madras, Raji, Kulynych, Mili, and
  Zemel}{Zhou et~al\mbox{.}}{[n.\,d.]}]%
        {zhouWeb}
\bibfield{author}{\bibinfo{person}{Angela Zhou}, \bibinfo{person}{David
  Madras}, \bibinfo{person}{Inioluwa~Raji Raji}, \bibinfo{person}{Bogdan
  Kulynych}, \bibinfo{person}{Smitha Mili}, {and} \bibinfo{person}{Richard
  Zemel}.} \bibinfo{year}{[n.\,d.]}\natexlab{}.
\newblock \bibinfo{title}{{Call for participation: Participatory Approaches to
  Machine Learning}}.
\newblock
\newblock
\urldef\tempurl%
\url{https://participatoryml.github.io/}
\showURL{%
\tempurl}


\bibitem[\protect\citeauthoryear{Zhu, Xu, Lu, Governatori, and Whittle}{Zhu
  et~al\mbox{.}}{2021}]%
        {zhu2021}
\bibfield{author}{\bibinfo{person}{Liming Zhu}, \bibinfo{person}{Xiwei Xu},
  \bibinfo{person}{Qinghua Lu}, \bibinfo{person}{Guido Governatori}, {and}
  \bibinfo{person}{Jon Whittle}.} \bibinfo{year}{2021}\natexlab{}.
\newblock \showarticletitle{{AI and Ethics -- Operationalising Responsible
  AI}}.
\newblock  (\bibinfo{date}{5} \bibinfo{year}{2021}).
\newblock


\bibitem[\protect\citeauthoryear{Zimmermann, Accorsi, and Muller}{Zimmermann
  et~al\mbox{.}}{2014}]%
        {zimmerman2014}
\bibfield{author}{\bibinfo{person}{Christian Zimmermann},
  \bibinfo{person}{Rafael Accorsi}, {and} \bibinfo{person}{Gunter Muller}.}
  \bibinfo{year}{2014}\natexlab{}.
\newblock \showarticletitle{{Privacy Dashboards: Reconciling Data-Driven
  Business Models and Privacy}}. In \bibinfo{booktitle}{\emph{2014 Ninth
  International Conference on Availability, Reliability and Security}}.
  \bibinfo{publisher}{IEEE}, \bibinfo{pages}{152--157}.
\newblock
\urldef\tempurl%
\url{https://doi.org/10.1109/ARES.2014.27}
\showDOI{\tempurl}


\bibitem[\protect\citeauthoryear{Zubiaga, Wang, Liakata, and Procter}{Zubiaga
  et~al\mbox{.}}{2019}]%
        {zubiaga2019}
\bibfield{author}{\bibinfo{person}{Arkaitz Zubiaga}, \bibinfo{person}{Bo Wang},
  \bibinfo{person}{Maria Liakata}, {and} \bibinfo{person}{Rob Procter}.}
  \bibinfo{year}{2019}\natexlab{}.
\newblock \showarticletitle{{Political Homophily in Independence Movements:
  Analyzing and Classifying Social Media Users by National Identity}}.
\newblock \bibinfo{journal}{\emph{IEEE Intelligent Systems}}
  \bibinfo{volume}{34}, \bibinfo{number}{6} (\bibinfo{year}{2019}),
  \bibinfo{pages}{34--42}.
\newblock
\urldef\tempurl%
\url{https://doi.org/10.1109/MIS.2019.2958393}
\showDOI{\tempurl}


\end{thebibliography}

\newpage
\appendix

\section{Tailored communication of system-related information} \label{appC}

\setlist[itemize]{wide=0pt, noitemsep, label=\textbullet, topsep=2pt, leftmargin=*, after =\vspace*{-\baselineskip}}

\begin{longtable}{p{.15\textwidth} p{.15\textwidth}  p{.175\textwidth}  p{.175\textwidth} p{.175\textwidth} p{.175\textwidth} } \hline
\footnotesize
     & & Development team & Auditing team & Data Domain experts & Decision subjects \\ \hline
     \multirow{4}{=}{Conservation}  & Privacy & \textcolor{lust}{[K]} & \textcolor{lust}{[K]} & &  \textcolor{bleudefrance}{[A]} \textcolor{darkpastelgreen}{[B]} \\ \cline{2-6}
     & Security  & \textcolor{lust}{[K]} \textcolor{lust}{[W]} \textcolor{darkpastelgreen}{[AB]} &  \textcolor{lust}{[K]}  \textcolor{lust}{[W]} & & \\ \cline{2-6}
     & \multirow{2}{=}{Performance}  & \textcolor{darkpastelgreen}{[F]} \textcolor{lust}{[G]} \textcolor{darkpastelgreen}{[H]}  \textcolor{darkpastelgreen}{[Y]}  \textcolor{darkpastelgreen}{[Z]} \textcolor{darkpastelgreen}{[AE]} & \textcolor{lust}{[G]} \textcolor{darkpastelgreen}{[H]} \textcolor{darkpastelgreen}{[Y]} \textcolor{darkpastelgreen}{[Z]} \textcolor{darkpastelgreen}{[AE]} &  \textcolor{darkpastelgreen}{[I]} \textcolor{bleudefrance}{[J]}  & \textcolor{bleudefrance}{[J]} \\ \hline
     \multirow{6}{=}{Universalism}  & Respect for public interest & \textcolor{lust}{[E]} \textcolor{darkpastelgreen}{[AE]} & \textcolor{lust}{[E]} \textcolor{darkpastelgreen}{[AE]} &  \textcolor{lust}{[E]} &   \textcolor{bleudefrance}{[C]} \textcolor{bleudefrance}{[D]} \\ \cline{2-6}
     & Fairness  &  \textcolor{lust}{[G]}  \textcolor{darkpastelgreen}{[H]} \textcolor{lust}{[K]} \textcolor{lust}{[W]} \textcolor{darkpastelgreen}{[X]} \textcolor{darkpastelgreen}{[Y]} \textcolor{darkpastelgreen}{[Z]} \textcolor{darkpastelgreen}{[AD]}  &  \textcolor{lust}{[G]}  \textcolor{darkpastelgreen}{[H]}  \textcolor{lust}{[K]} \textcolor{lust}{[W]} \textcolor{darkpastelgreen}{[X]} \textcolor{darkpastelgreen}{[Y]} \textcolor{darkpastelgreen}{[Z]} \textcolor{darkpastelgreen}{[AD]}  & \textcolor{darkpastelgreen}{[I]} \textcolor{bleudefrance}{[J]} &  \textcolor{bleudefrance}{[J]} \\ \cline{2-6}
     & Non-discrimination &  \textcolor{darkpastelgreen}{[H]} \textcolor{lust}{[K]}  \textcolor{darkpastelgreen}{[X]} \textcolor{darkpastelgreen}{[Y]} \textcolor{darkpastelgreen}{[AD]}  &  \textcolor{darkpastelgreen}{[H]}\textcolor{lust}{[K]}  \textcolor{darkpastelgreen}{[X]} \textcolor{darkpastelgreen}{[Y]} \textcolor{darkpastelgreen}{[AD]} & \textcolor{bleudefrance}{[J]} \textcolor{bleudefrance}{[L]} & \textcolor{bleudefrance}{[J]} \textcolor{bleudefrance}{[L]} \\ \hline
     \multirow{3}{=}{Openness}  & Transparency  &  \textcolor{darkpastelgreen}{[H]} \textcolor{lust}{[K]} \textcolor{bleudefrance}{[M]} & \textcolor{darkpastelgreen}{[H]}\textcolor{lust}{[K]}  \textcolor{bleudefrance}{[M]}  &  \textcolor{darkpastelgreen}{[I]} \textcolor{bleudefrance}{[J]} \textcolor{bleudefrance}{[L]}  \textcolor{bleudefrance}{[M]}   &
     \textcolor{darkpastelgreen}{[B]}
     \textcolor{bleudefrance}{[J]} \textcolor{bleudefrance}{[L]} \textcolor{bleudefrance}{[M]} \\ \cline{2-6}
     & \multirow{2}{=}{Explainability}  & \textcolor{bleudefrance}{[M]} \textcolor{bleudefrance}{[N]} \textcolor{bleudefrance}{[O]} \textcolor{bleudefrance}{[Q]} \textcolor{darkpastelgreen}{[AC]} \textcolor{darkpastelgreen}{[AD]} \textcolor{bleudefrance}{[P]} &  \multirow{2}{=}{\textcolor{bleudefrance}{[M]}  \textcolor{bleudefrance}{[N]}  \textcolor{bleudefrance}{[O]} \textcolor{bleudefrance}{[Q]}  \textcolor{darkpastelgreen}{[AC]} \textcolor{darkpastelgreen}{[AD]} \textcolor{bleudefrance}{[P]}} &  \textcolor{bleudefrance}{[J]} \textcolor{bleudefrance}{[M]}  \textcolor{bleudefrance}{[N]}  \textcolor{bleudefrance}{[O]}  \textcolor{bleudefrance}{[Q]} \textcolor{bleudefrance}{[P]}   &  \textcolor{bleudefrance}{[J]}  \textcolor{bleudefrance}{[M]} \textcolor{bleudefrance}{[N]} \textcolor{bleudefrance}{[O]}  \textcolor{bleudefrance}{[Q]} \textcolor{bleudefrance}{[R]} \textcolor{bleudefrance}{[S]} \textcolor{bleudefrance}{[P]} \\ \hline
     \multirow{3}{=}{Individual empowerment} & Contestability  & \textcolor{bleudefrance}{[U]} & \textcolor{bleudefrance}{[U]} & \textcolor{bleudefrance}{[T]} \textcolor{bleudefrance}{[U]} & \textcolor{bleudefrance}{[T]} \textcolor{bleudefrance}{[AF]}  \\ \cline{2-6} 
     & Human Control  &  \textcolor{bleudefrance}{[V]} &  \textcolor{bleudefrance}{[V]}    & \textcolor{bleudefrance}{[T]} \textcolor{bleudefrance}{[V]}  & \textcolor{bleudefrance}{[C]}  \textcolor{bleudefrance}{[T]}  \textcolor{bleudefrance}{[V]} \\ \cline{2-6}
     & Human Agency & & & \textcolor{bleudefrance}{[T]}  & \textcolor{bleudefrance}{[T]}  \textcolor{darkpastelgreen}{[B]} \textcolor{darkpastelgreen}{[AA]}   \\ \hline

    \caption{Mapping of available means for transmitting value-specific manifestations to different stakeholders based on the purpose of their insight and the nature of their knowledge. These means have been classified into three main categories: descriptive documents specifying whether/how a value manifestation is fulfilled (red), strategies for fulfilling value manifestations (blue), and complete tools for enabling the fulfillment of value manifestations (green). This table aims at facilitating the navigation of table \ref{tab:detailedmeans}, where each means is documented.} \label{tab:tailoredmeanscodes}    
\end{longtable}

\newpage

\begin{landscape}

\setlist[itemize]{wide=0pt, noitemsep, label=\textbullet, topsep=2pt, leftmargin=*, after =\vspace*{-\baselineskip}}
\begin{singlespace}
\begin{longtable}{p{.02\textwidth} p{.09\textwidth} p{.07\textwidth}  p{.22\textwidth}  p{.02\textwidth} p{.02\textwidth} p{.02\textwidth} p{.02\textwidth} p{.1\textwidth} p{.13\textwidth} p{.16\textwidth} p{.1\textwidth}} \hline

&\multirow{2}{=}{Means} & \multirow{2}{=}{Value} & \multirow{2}{=}{Manifestation(s)} & \multicolumn{4}{c}{Stakeholder} & \multirow{2}{=}{Application (model) }& \multirow{2}{=}{Approach} & \multirow{2}{=}{Visual elements} & \multirow{2}{=}{Additional details}\\ \cline{5-8}
& & & & DT & AT & DE & DS & & & & \\ \hline \endhead
\multirow{8}{=}{\textcolor{bleudefrance}{[A]}} & \multirow{8}{=}{Iconsets for data privacy declarations \cite{GDPR2018, rossi2017, holtz2011, mehldau2007}} & \multirow{8}{=}{Privacy} & \mbox{}\par\vspace{-\baselineskip} \begin{itemize}  \item Description of what data is collected \item Description of how data is handled \item Purpose statement of data collection \item Statement of how long the data is kept \end{itemize} & & & & \multirow{8}{=}{\checkmark} & \multirow{8}{=}{Agnostic} & & \multirow{8}{=}{Iconsets} & \\ \hline
\multirow{10}{=}{\textcolor{darkpastelgreen}{[B]}} & \multirow{10}{=}{Privacy dashboards \cite{zimmerman2014, earp2016, fischerhubner2016, herder2020, farke2021}} & \multirow{6}{=}{Privacy} & \mbox{}\par\vspace{-\baselineskip} \begin{itemize}  \item Description of what data is collected \item Description of how data is handled \item Purpose statement of data collection \end{itemize} & & & & \multirow{10}{=}{\checkmark} & \multirow{10}{=}{Agnostic} & & \mbox{}\par\vspace{-\baselineskip} \begin{itemize} \item[\textcolor{white}{\textbullet}]  \item[\textcolor{white}{\textbullet}] \item[\textcolor{white}{\textbullet}] \item[\textcolor{white}{\textbullet}]   \item Timelines \item Bar charts  \end{itemize}   &  \\ \cline{3-4}
& &  Human agency & \mbox{}\par\vspace{-\baselineskip} \begin{itemize}  \item Opportunity to self-assess the system \end{itemize} & & & & & & &\mbox{}\par\vspace{-\baselineskip}\begin{itemize}\item Maps \item Network graphs \end{itemize}&  \\ \cline{3-4}
& &  Trans-parency & \mbox{}\par\vspace{-\baselineskip} \begin{itemize}  \item Disclosure of origin and properties of data \end{itemize} & & & & & & & &  \\ \hline
\multirow{7}{=}{\textcolor{bleudefrance}{[C]}} & \multirow{7}{=}{Risk matrix \cite{krafft2019, krafft2020}} & Respect for public interest & \mbox{}\par\vspace{-\baselineskip} \begin{itemize}   \item[\textcolor{white}{\textbullet}] \item Measure of social impact \end{itemize}& & & & \multirow{7}{=}{\checkmark} & \multirow{7}{=}{Agnostic} & & \multirow{7}{=}{  Two dimensional space (vulnerability vs dependence of the decision)}& \\ \cline{3-4}
& & \multirow{3}{=}{Human Control} & \mbox{}\par\vspace{-\baselineskip} \begin{itemize}  \item Ability to override the decision made by a system \end{itemize} & & & & & & & &  \\ \hline
\multirow{5}{=}{\textcolor{bleudefrance}{[D]}}&  \multirow{5}{=}{Moral space \cite{hidalgo2021}} & \multirow{5}{=}{Respect for public interest} & \mbox{}\par\vspace{-\baselineskip} \begin{itemize}  \item[\textcolor{white}{\textbullet}]  \item[\textcolor{white}{\textbullet}]\item Measure of social impact \end{itemize} & & & & \multirow{5}{=}{\checkmark} & \multirow{5}{=}{Agnostic} & \multirow{5}{=}{Based on human judgement} & \mbox{}\par\vspace{-\baselineskip} Three dimensional moral space. Wrongness as a function of intention and harm & \\ \hline
\multirow{4}{=}{\textcolor{lust}{[E]}} & \multirow{4}{=}{Social impact assessment \cite{raji2020}} & Respect for public interest & \mbox{}\par\vspace{-\baselineskip} \begin{itemize} \item[\textcolor{white}{\textbullet}]  \item Measure of social impact \end{itemize} & \multirow{4}{=}{\checkmark} & \multirow{4}{=}{\checkmark} & \multirow{4}{=}{\checkmark} & & \multirow{4}{=}{Agnostic} & \multirow{4}{=}{Anticipate scenarios} & & \\ \hline
\multirow{16}{=}{\textcolor{darkpastelgreen}{[F]}}& \multirow{16}{=}{Model Tracker interactive visualization \cite{amershi2015}} & \begin{itemize}\item[\textcolor{white}{\textbullet}]\item[\textcolor{white}{\textbullet}]\item[\textcolor{white}{\textbullet}]\item[\textcolor{white}{\textbullet}]\item[\textcolor{white}{\textbullet}]\item[\textcolor{white}{\textbullet}]\item[\textcolor{white}{\textbullet}]\end{itemize}Perfor- mance & \mbox{}\par\vspace{-\baselineskip} \begin{itemize} \item[\textcolor{white}{\textbullet}]\item[\textcolor{white}{\textbullet}]\item[\textcolor{white}{\textbullet}]\item[\textcolor{white}{\textbullet}]\item[\textcolor{white}{\textbullet}]\item[\textcolor{white}{\textbullet}] \item Accuracy \item False Positive and Negative rates \end{itemize} & \multirow{16}{=}{\checkmark} & & & & \multirow{16}{=}{Classification tasks} & & \mbox{}\par\vspace{-\baselineskip} \begin{itemize}  \item Summary statistics \item Confusion matrices \item Labels chart \item Precision-recall curves \item Connector lines to identify similar examples in feature space \item Highlighted boxes for correlations between features and target classes \end{itemize} \\ \hline
\multirow{11}{=}{\textcolor{lust}{[G]}} &  \multirow{11}{=}{Model cards for models \cite{mitchell2019}} & \multirow{5}{=}{Perfor-mance} & \mbox{}\par\vspace{-\baselineskip} \begin{itemize}  \item Accuracy \item False Positive and Negative rates \item False Discovery and omission rates \end{itemize} & \multirow{11}{=}{\checkmark} & \multirow{11}{=}{\checkmark} & & & \multirow{11}{=}{Agnostic} & & \multirow{11}{=}{\begin{itemize}  \item Confidence bars \item Bar charts \end{itemize}} & \\ \cline{3-4} 
& & \multirow{6}{=}{Fairness} & \mbox{}\par\vspace{-\baselineskip} \begin{itemize}  \item Accuracy across groups \item False Positive and Negative rates across groups \item False Discovery and omission rates across groups \end{itemize} & & & & & & & &  \\ \hline \pagebreak
\multirow{16}{=}{\textcolor{darkpastelgreen}{[H]}} & \multirow{16}{=}{What-if tool \footnote{\url{https://github.com/pair-code/what-if-tool}} \cite{wexler2019}} & \multirow{5}{=}{Perfor-mance} & \mbox{}\par\vspace{-\baselineskip} \begin{itemize}  \item Accuracy \item False Positive and Negative Rates \item False Discovery and omission rates \end{itemize} & \multirow{16}{=}{\checkmark} & \multirow{16}{=}{\checkmark} & & & \multirow{16}{=}{Classification tasks, Regression tasks} & & \multirow{16}{=}{ \begin{itemize}  \item Confusion matrices \item (Two-dimensional) Histograms \item Scatterplots \item Summary statistics of datasets \item Partial dependence plots \end{itemize}} & \multirow{16}{=}{Interactive modules include: list of feature values, inference values, and counterfactual controls} \\ \cline{3-4}
& & \multirow{6}{=}{Fairness} & \mbox{}\par\vspace{-\baselineskip} \begin{itemize}  \item Accuracy across groups \item False Positive and Negative Rates across groups \item False Discovery and omission rates across groups \end{itemize} & & & & & & & &  \\ \cline{3-4}
& & \multirow{2}{=}{Trans-parency} & \mbox{}\par\vspace{-\baselineskip} \begin{itemize} \item Disclosure of origin and properties of data \end{itemize} & & & & & & & &  \\ \cline{3-4}
& & \multirow{3}{=}{Non-discrimi-nation} & \mbox{}\par\vspace{-\baselineskip} \begin{itemize} \item Analysis of data for potential biases, data quality assessment \end{itemize} & & & & & & & &  \\ \hline
\multirow{9}{=}{\textcolor{darkpastelgreen}{[I]}} & \multirow{9}{=}{Interactive transfer learning tools \cite{mishra2021}} & \multirow{3}{=}{Perfor-mance} & \mbox{}\par\vspace{-\baselineskip} \begin{itemize} \item Accuracy \item False Positive and Negative Rates \end{itemize} & & & \multirow{9}{=}{\checkmark} & & \multirow{9}{=}{Convolutional Neural Networks} & & \multirow{6}{=}{ \begin{itemize} \item Confusion matrices \item Z-scored of each filter \item Bar charts \item Activation heatmaps \item t-SNE clusters \end{itemize}}   & \\ \cline{3-4}
& & \multirow{3}{=}{Fairness} & \mbox{}\par\vspace{-\baselineskip} \begin{itemize} \item Accuracy across groups \item False Positive and Negative Rates across groups \end{itemize} & & & & & & & &  \\ \cline{3-4}
& & \multirow{2}{=}{Trans-parency} & \mbox{}\par\vspace{-\baselineskip} \begin{itemize} \item Disclosure of properties of data \item[\textcolor{white}{\textbullet}] \end{itemize} & & & & & & & &  \\ \hline \pagebreak
\multirow{12}{=}{\textcolor{bleudefrance}{[J]}} & \multirow{12}{=}{Question-Driven XAI Design \cite{veraliao2021}} & Perfor-mance & \mbox{}\par\vspace{-\baselineskip} \begin{itemize} \item Accuracy \end{itemize} & & & \multirow{12}{=}{\checkmark} & \multirow{12}{=}{\checkmark} & \multirow{12}{=}{Agnostic} & & \multirow{12}{=}{ \begin{itemize} \item Summary statistics (percentage scores) for data explanations and performance metrics \item Feature importance \item Contrastive explanations \end{itemize}} &  \multirow{12}{=}{End users were more interested in the limitation of the model: uncertainty}\\ \cline{3-4}
& & Fairness & \mbox{}\par\vspace{-\baselineskip} \begin{itemize} \item Accuracy across groups \end{itemize} & & & & & & & &  \\ \cline{3-4}
& & \multirow{2}{=}{Trans-parency} & \mbox{}\par\vspace{-\baselineskip} \begin{itemize} \item Disclosure of origin and properties of data \end{itemize} & & & & & & & &  \\ \cline{3-4}
& & \multirow{3}{=}{Non-discrimi-nation} & \mbox{}\par\vspace{-\baselineskip} \begin{itemize} \item Analysis of data for potential biases, data quality assessment \item[\textcolor{white}{\textbullet}] \end{itemize} & & & & & & & &  \\ \cline{3-4}
& & Explain-ability & \mbox{}\par\vspace{-\baselineskip} \begin{itemize} \item Post-hoc explanations \end{itemize} & & & & & & & &  \\ \hline
\multirow{21}{=}{\textcolor{lust}{[K]}} & \multirow{21}{=}{Datasheets for datasets \cite{gebru2020}} & \multirow{5}{=}{Trans-parency} & \mbox{}\par\vspace{-\baselineskip} \begin{itemize} \item Description of data generation process \item Disclosure of origin properties of models and data \end{itemize} &  \multirow{21}{=}{\checkmark} & \multirow{21}{=}{\checkmark} & & & \multirow{21}{=}{Agnostic} & & \multirow{21}{=}{ \begin{itemize} \item Summary statistics \item Visual examples of datasets (if images, for instance) \end{itemize}}  &  \\ \cline{3-4}
& & \multirow{3}{=}{Non-discrimi-nation} & \mbox{}\par\vspace{-\baselineskip} \begin{itemize} \item Analysis of data for potential biases, data quality assessment \end{itemize} & & & & & & & &  \\ \cline{3-4}
& & \multirow{10}{=}{Privacy} & \mbox{}\par\vspace{-\baselineskip} \begin{itemize} \item Written declaration of consent \item Description of what data is collected \item Description of how data is handled \item Purpose statement of data collection \item Statement of how long the data is kept \end{itemize} & & & & & & & &  \\ \cline{3-4}
& & \multirow{2}{=}{Fairness} & \mbox{}\par\vspace{-\baselineskip} \begin{itemize} \item Election of protected classes \end{itemize} & & & & & & & &  \\ \cline{3-4}
& & Security & \mbox{}\par\vspace{-\baselineskip} \begin{itemize} \item Membership inference \end{itemize} & & & & & & & &  \\ \hline 
\multirow{8}{=}{\textcolor{bleudefrance}{[L]}} & \multirow{8}{=}{Data centric explanations \cite{anik2021}} & \multirow{5}{=}{Trans-parency} & \mbox{}\par\vspace{-\baselineskip} \begin{itemize} \item Description of data generation process \item Disclosure of origin and properties of the models and data \end{itemize} & & & \multirow{8}{=}{\checkmark} & \multirow{8}{=}{\checkmark} & \multirow{8}{=}{Agnostic} & & \multirow{8}{=}{\begin{itemize} \item Interactive list \item Q\&A format \item Pie charts \item Bar charts \item Process diagrams \item timelines \item Icons \end{itemize}} &  \\ \cline{3-4}
& & \multirow{3}{=}{Non-discrimi-nation} & \mbox{}\par\vspace{-\baselineskip} \begin{itemize} \item Analysis of data for potential biases, data quality assessment \end{itemize} & & & & & & & &  \\ \hline 
\multirow{7}{=}{\textcolor{bleudefrance}{[M]}} & \multirow{7}{=}{Example-based explanations \cite{cai2019, veraliao2020, veraliao2021, jin2021, barredoarrieta2020, dodge2019, binns2018}} & \multirow{3}{=}{Trans-parency} & \mbox{}\par\vspace{-\baselineskip} \begin{itemize} \item[\textcolor{white}{\textbullet}] \item Disclosure of properties of data \item[\textcolor{white}{\textbullet}] \end{itemize} & \multirow{7}{=}{\checkmark} & \multirow{7}{=}{\checkmark} & \multirow{7}{=}{\checkmark} & \multirow{7}{=}{\checkmark} & \multirow{7}{=}{Agnostic} & \multirow{7}{=}{ \begin{itemize} \item Similar example \item Typical example \item Counter-factual example \end{itemize}} &  \multirow{7}{=}{ \begin{itemize} \item Example images from dataset if in the visual domain \end{itemize}} & \multirow{7}{=}{Normative vs comparative explanations \cite{cai2019}}  \\ \cline{3-4}
& & \multirow{4}{=}{Explain-ability} & \mbox{}\par\vspace{-\baselineskip} \begin{itemize} \item[\textcolor{white}{\textbullet}]  \item Post-hoc explanations \item[\textcolor{white}{\textbullet}] \item[\textcolor{white}{\textbullet}]   \end{itemize} & & & & & & & &  \\ \hline
 \multirow{4}{=}{\textcolor{bleudefrance}{[N]}} & Explanation by simplification \cite{barredoarrieta2020, jin2021}& \multirow{4}{=}{Explain-ability} & \multirow{4}{=}{ \begin{itemize} \item Post-hoc explanations \end{itemize}} & \multirow{4}{=}{\checkmark} & \multirow{4}{=}{\checkmark} & \multirow{4}{=}{\checkmark} & \multirow{4}{=}{\checkmark} & \multirow{4}{=}{Agnostic} & \multirow{3}{=}{ \begin{itemize} \item Decision rule \item Decision tree \end{itemize}} & &  \\ \hline
\multirow{12}{=}{\textcolor{bleudefrance}{[O]}} & \multirow{12}{=}{Feature relevance explanation \cite{barredoarrieta2020, jin2021, veraliao2021, alqaraawi2020, dodge2019, binns2018}} & \multirow{12}{=}{Explain-ability} & \multirow{12}{=}{\begin{itemize} \item Post-hoc explanations \end{itemize}} & \multirow{12}{=}{\checkmark} & \multirow{12}{=}{\checkmark} & \multirow{12}{=}{\checkmark} & \multirow{12}{=}{\checkmark} & \multirow{12}{=}{Agnostic} & \mbox{}\par\vspace{-\baselineskip} \begin{itemize} \item Feature attribute \item Feature shape \item Feature interaction \item Sensitivity / perturbation -based \item Saliency maps (visual domain) \end{itemize} & \multirow{12}{=}{ \begin{itemize} \item Bar charts \item Visualization of element importance, saliency (visual domain) \end{itemize}} & Usability of saliency maps for non-experts \cite{alqaraawi2020}. They should be accompanied by global descriptors \\ \hline
\multirow{6}{=}{\textcolor{bleudefrance}{[P]}} &  \multirow{6}{=}{Contrastive explanations  \cite{veraliao2021, dhurandhar2018, mothilal2019}} & \multirow{6}{=}{Explain-ability} & \multirow{5}{=}{ \begin{itemize} \item Post-hoc explanations \end{itemize}} &  \multirow{6}{=}{\checkmark} & \multirow{6}{=}{\checkmark} & \multirow{6}{=}{\checkmark} & \multirow{6}{=}{\checkmark} & \multirow{6}{=}{Agnostic} & \mbox{}\par\vspace{-\baselineskip} \begin{itemize} \item Example of minimum change that leads to different outcomes \end{itemize} & &  \\ \hline
 \multirow{4}{=}{\textcolor{bleudefrance}{[Q]}} & Text-based explanation \cite{vanBerkel2021, barredoarrieta2020} & \multirow{4}{=}{Explain-ability} & \multirow{3}{=}{\begin{itemize} \item Post-hoc explanations \end{itemize}} & \multirow{4}{=}{\checkmark} & \multirow{4}{=}{\checkmark} & \multirow{4}{=}{\checkmark} & \multirow{4}{=}{\checkmark} & \multirow{4}{=}{Agnostic} &  \begin{itemize} \item With or without outcome comparison \end{itemize} & & \\\hline
\multirow{4}{=}{\textcolor{bleudefrance}{[R]}}&  Interactive demonstrations \cite{long2020} & \multirow{4}{=}{Explain-ability} & \multirow{3}{=}{\begin{itemize} \item Post-hoc explanations \end{itemize}} & & & & \multirow{4}{=}{\checkmark}& \multirow{4}{=}{Agnostic} & & & \\ \hline
\multirow{6}{=}{\textcolor{bleudefrance}{[S]}} &  \multirow{6}{=}{Experiential AI \cite{hemment2019}} & \multirow{6}{=}{Explain-ability} & \multirow{5}{=}{\begin{itemize} \item Post-hoc explanations \end{itemize}} & & & & \multirow{6}{=}{\checkmark} & \multirow{6}{=}{Agnostic} & \mbox{}\par\vspace{-\baselineskip} \begin{itemize} \item Art mediated between computer code and human comprehension \end{itemize} & & \\\hline
\multirow{9}{=}{\textcolor{bleudefrance}{[T]}} &  \multirow{9}{=}{Interactive contestations \cite{henin2021, kluttz2018}} & \multirow{4}{=}{Contest-ability} & \mbox{}\par\vspace{-\baselineskip} \begin{itemize} \item Mechanisms for users to ask questions and record disagreements with system behavior \end{itemize} & & & \multirow{9}{=}{\checkmark} & \multirow{9}{=}{\checkmark} & \multirow{9}{=}{Agnostic} & \multirow{9}{=}{\begin{itemize} \item Statements restricted to natural language \end{itemize}} & & \\ \cline{3-4}
& & \multirow{3}{=}{Human Control} & \mbox{}\par\vspace{-\baselineskip} \begin{itemize} \item Ability to override the decision made by the system \end{itemize} & & & & & & & &  \\ \cline{3-4}
& & Human agency & \mbox{}\par\vspace{-\baselineskip} \begin{itemize} \item Opportunity to self-assess the system \end{itemize} & & & & & & & &  \\ \hline
\multirow{9}{=}{\textcolor{bleudefrance}{[U]}} &  Challenge justifications provided by operator using the same means \cite{henin2021} & \multirow{9}{=}{Contest-ability} & \multirow{9}{=}{\begin{itemize} \item Mechanisms for users to ask questions and record disagreements with system behavior \end{itemize}} & \multirow{9}{=}{\checkmark} & \multirow{9}{=}{\checkmark} & \multirow{9}{=}{\checkmark} & & \multirow{9}{=}{Agnostic}  & \multirow{8}{=}{ \begin{itemize} \item Further testing \item Verification \end{itemize}} & &  \\ \hline
\multirow{7}{=}{\textcolor{bleudefrance}{[V]}} &  Mapping of actors and tasks depending on automation level \cite{calvert2019} & \multirow{7}{=}{Human Control}  & \multirow{6}{=}{ \begin{itemize} \item Establishment of levels of human discretion during the use of the system \end{itemize}} & \multirow{7}{=}{\checkmark} & \multirow{7}{=}{\checkmark} & \multirow{7}{=}{\checkmark} & \multirow{7}{=}{\checkmark} & \multirow{7}{=}{Agnostic} & &  \multirow{6}{=}{\begin{itemize} \item Relationship diagrams\end{itemize}} &  \\ \hline
 \multirow{8}{=}{\textcolor{lust}{[W]}} & \multirow{7}{=}{Failure Modes and Effects Analysis \cite{raji2020}} & \multirow{4}{=}{Security}  & \mbox{}\par\vspace{-\baselineskip} \begin{itemize} \item Threats against integrity (adversarial learning) and mitigation techniques \end{itemize} & \multirow{8}{=}{\checkmark} & \multirow{8}{=}{\checkmark} & & & \multirow{8}{=}{Agnostic} & & &  \\ \cline{3-4}
& & \multirow{3}{=}{Fairness} & \mbox{}\par\vspace{-\baselineskip} \begin{itemize} \item Accuracy across groups \item False positives and negatives across groups \end{itemize} & & & & & & & &  \\ \hline
\multirow{12}{=}{\textcolor{darkpastelgreen}{[X]}} & \multirow{12}{=}{Aequitas \footnote{\url{https://github.com/dssg/aequitas}} \cite{saleiro2018}} & \multirow{9}{=}{Fairness} & \mbox{}\par\vspace{-\baselineskip} \begin{itemize} \item Accuracy across groups \item False Positive and Negative rates across groups \item False Discovery and Omission rates across groups \item Counterfactual examples \end{itemize} & \multirow{12}{=}{\checkmark} & \multirow{12}{=}{\checkmark} & & & \multirow{11}{=}{Agnostic} & & &  \\ \cline{3-4}
& & Non-discrimi-nation & \mbox{}\par\vspace{-\baselineskip} \begin{itemize} \item Analysis of data for potential biases, data quality assessment \end{itemize} & & & & & & & &  \\ \hline
\multirow{8}{=}{\textcolor{darkpastelgreen}{[Y]}} & \multirow{8}{=}{AI Fairness 360 \footnote{\url{https://github.com/Trusted-AI/AIF360}} \cite{bellamy2018}} & \multirow{2}{=}{Perfor-mance} & \mbox{}\par\vspace{-\baselineskip} \begin{itemize} \item False Positive and Negative rates \end{itemize} & \multirow{8}{=}{\checkmark} & \multirow{8}{=}{\checkmark} & & & \multirow{8}{=}{Classifiers: logistic regression, random forest classifier and neural networks} & & \multirow{8}{=}{\begin{itemize} \item Bar charts \item Confidence bars \end{itemize}} &  \\ \cline{3-4}
& & \multirow{3}{=}{Fairness} & \mbox{}\par\vspace{-\baselineskip} \begin{itemize} \item False positive and negative rates across groups \item Debiasing algorithms \end{itemize}  & & & & & & &  \\ \cline{3-4}
& & Non-discrimi-nation & \mbox{}\par\vspace{-\baselineskip} \begin{itemize} \item Analysis of data for potential biases, data quality assessment \end{itemize} & & & & & & & &  \\ \hline 
\multirow{10}{=}{\textcolor{darkpastelgreen}{[Z]}} & \multirow{10}{=}{Fairlearn \footnote{\url{https://github.com/fairlearn/fairlearn}} \cite{bird2020}} & \multirow{5}{=}{Perfor-mance} & \mbox{}\par\vspace{-\baselineskip} \begin{itemize} \item Accuracy \item False Positive and False Negative rates \item Precision and recall rates \end{itemize} & \multirow{10}{=}{\checkmark} & \multirow{10}{=}{\checkmark} & & & \multirow{10}{=}{Agnostic} & & \multirow{9}{=}{ \begin{itemize} \item Bar charts \item Pie charts \end{itemize}} & \\ \cline{3-4}
& & \multirow{5}{=}{Fairness} & \mbox{}\par\vspace{-\baselineskip} \begin{itemize} \item Accuracy across groups \item False negative and false positive rates across groups \item Debiasing algorithms \end{itemize} & & & & & & &  \\ \hline
\multirow{5}{=}{\textcolor{darkpastelgreen}{[AA]}} & \multirow{5}{=}{Playbook AI \footnote{\url{https://github.com/microsoft/HAXPlaybook}} \cite{hong2021}} & \multirow{5}{=}{Human agency}  & \mbox{}\par\vspace{-\baselineskip} \begin{itemize} \item Give knowledge and tools to comprehend and interact with AI systems \item Opportunity to self-assess the system \end{itemize} & & & & \multirow{5}{=}{\checkmark} & \multirow{5}{=}{NLP} & \multirow{5}{=}{Early AI prototyping} & \multirow{4}{=}{ \begin{itemize} \item Interactive survey \end{itemize}} & \\ \hline
\multirow{4}{=}{\textcolor{darkpastelgreen}{[AB]}} & \multirow{4}{=}{Counterfit \footnote{\url{https://github.com/Azure/counterfit}}} & \multirow{4}{=}{Security} & \mbox{}\par\vspace{-\baselineskip} \begin{itemize} \item Defence against integrity threats \item Defence against privacy threats \end{itemize} & \multirow{4}{=}{\checkmark} & & & & \multirow{4}{=}{Agnostic} & & & \\ \hline
\multirow{4}{=}{\textcolor{darkpastelgreen}{[AC]}} & \multirow{4}{=}{InterpretML \footnote{\url{https://github.com/interpretml/interpret/}} \footnote{\url{https://github.com/interpretml/DiCE}} \cite{nori2019, mothilal2019}} & \multirow{4}{=}{Explain-ability}  & \begin{itemize} \item Interpretability by design \item Post-hoc explanations \end{itemize} & \multirow{4}{=}{\checkmark} & \multirow{4}{=}{\checkmark} & 
 &  & Both whitebox and blackbox models & & \begin{itemize} \item Bar charts \item Line charts \item Decision trees \item[\textcolor{white}{\textbullet}] \end{itemize} & \\ \hline 
\multirow{7}{=}{\textcolor{darkpastelgreen}{[AD]}} & \multirow{7}{=}{Error analysis dashboard \footnote{\url{https://github.com/microsoft/responsible-ai-toolbox/blob/main/docs/erroranalysis-dashboard-README.md}}} & Non-discrimi-nation & \mbox{}\par\vspace{-\baselineskip} \begin{itemize} \item Analysis of data for potential biases, data quality assessment \end{itemize} & \multirow{7}{=}{\checkmark}& \multirow{7}{=}{\checkmark} & & & \multirow{7}{=}{Agnostic} & & \multirow{6}{=}{\begin{itemize} \item Decision tree \item Error heatmap \end{itemize}} & \\ \cline{3-4}
& & Explain-ability & \mbox{}\par\vspace{-\baselineskip} \begin{itemize} \item  Post-hoc explanations \end{itemize} & & & & & & &  \\ \cline{3-4}
& & Fairness & \mbox{}\par\vspace{-\baselineskip} \begin{itemize} \item Accuracy across groups \end{itemize} & & & & & & &  \\ \hline
\multirow{8}{=}{\textcolor{darkpastelgreen}{[AE]}} & \multirow{8}{=}{ Breakend Impact tracker \footnote{\url{https://github.com/Breakend/experiment-impact-tracker}} \cite{henderson2020}} & \multirow{4}{=}{Perfor-mance} & \mbox{}\par\vspace{-\baselineskip} \begin{itemize} \item Estimation of energy consumption \item Estimation of GPU memory consumption \end{itemize} & \multirow{8}{=}{\checkmark} & \multirow{8}{=}{\checkmark} & &  & \multirow{8}{=}{Agnostic} & &\multirow{8}{=}{ \begin{itemize} \item Dot plots \item Bar charts \end{itemize}} &  \\ \cline{3-4}
& & Respect for public interest & \multirow{3}{=}{ \begin{itemize} \item Measure of environmental impact \end{itemize}} & & & & & & & \\ \hline
\multirow{4}{=}{\textcolor{bleudefrance}{[AF]}}  & \multirow{4}{=}{Represent-ative contestations \cite{vaccaro2020}} & \multirow{4}{=}{Contest-ability} & \mbox{}\par\vspace{-\baselineskip} \begin{itemize} \item Mechanisms for users to ask questions and record disagreement with system behaviour \end{itemize}& & & & \multirow{4}{=}{\checkmark} & \multirow{4}{=}{Agnostic} & & & \\ \hline

    \caption{Mapping of available means for transmitting value-specific manifestations to different stakeholders based on the purpose of their insight and the nature of their knowledge (DT = Development Team; AT = Auditing Team; DE = Data Domain Experts; DS = Decision Subjects). The identification and color code correspond to those on table \ref{tab:tailoredmeanscodes}. Each means is linked to the value and criteria manifestations that they communicate, the stakeholders that the original papers address, model specificity, deployed approach, visual elements and any additional details.  } \label{tab:detailedmeans}    
\end{longtable}
\end{singlespace}

\end{landscape}

\end{document}